\documentclass[10pt,twocolumn,letterpaper]{article}

\usepackage{cvpr}              
\usepackage{stfloats}
\definecolor{cvprblue}{rgb}{0.21,0.49,0.74}
\usepackage[pagebackref,breaklinks,colorlinks,allcolors=cvprblue]{hyperref}
\usepackage{array}
\usepackage{pifont} %
\usepackage{multirow}
\usepackage{makecell}
\usepackage{float}
\usepackage{enumitem}
\usepackage{xcolor}   
\usepackage[table]{xcolor}
\definecolor{lightpink}{RGB}{255, 182, 193}
\definecolor{iccvblue}{rgb}{0.21,0.49,0.74}

\def\paperID{41247} 
\def\confName{CVPR}
\def\confYear{2026}

\title{Boosting Document Parsing Efficiency and Performance \\ with Coarse-to-Fine Visual Processing }

\author{
    Cheng Cui$^{1}$\thanks{The authors contributed equally.}  \quad
    Ting Sun$^{1}$\footnotemark[1] \quad
    Suyin Liang$^{1,2}$ \quad
    Tingquan Gao$^{1}$ \quad
    Zelun Zhang$^{1}$ \quad
    Jiaxuan Liu$^{1}$ \\
    Xueqing Wang$^{1}$ \quad
    Changda Zhou$^{1}$ \quad
    Hongen Liu$^{1}$ \quad
    Manhui Lin$^{1}$ \quad
    Yue Zhang$^{1}$ \quad
    Yubo Zhang$^{1}$ \\
    Jing Zhang$^{1}$ \quad
    Jun Zhang$^{1}$ \quad
    Xing Wei$^{2}$ \quad
    Yi Liu$^{1}$\thanks{Corresponding author (liuyi22@baidu.com).} \quad
    Dianhai Yu$^{1}$ \quad
    Yanjun Ma$^{1}$ \\[3mm]
    $^{1}$PaddlePaddle Team, Baidu Inc. \qquad
    $^{2}$Xi'an Jiaotong University \\[1mm]
}

\begin{document}

\maketitle

\begin{abstract}
\vspace{-10pt}
Document parsing is a fine-grained task where image resolution significantly impacts performance. While advanced research leveraging vision-language models benefits from high-resolution input to boost model performance, this often leads to a quadratic increase in the number of vision tokens and significantly raises computational costs. We attribute this inefficiency to substantial visual regions redundancy in document images, like background.
To tackle this, we propose PaddleOCR-VL, a novel coarse-to-fine architecture that focuses on semantically relevant regions while suppressing redundant ones, thereby improving both efficiency and performance. Specifically, we introduce a lightweight Valid Region Focus Module (VRFM) which leverages localization and contextual relationship prediction capabilities to identify valid vision tokens. Subsequently, we design and train a compact yet powerful 0.9B vision-language model (PaddleOCR-VL-0.9B) to perform detailed recognition, guided by VRFM outputs to avoid direct processing of the entire large image. Extensive experiments demonstrate that PaddleOCR-VL achieves state-of-the-art performance in both page-level parsing and element-level recognition. It significantly outperforms existing solutions, exhibits strong competitiveness against top-tier VLMs, and delivers fast inference while utilizing substantially fewer vision tokens and parameters, highlighting the effectiveness of targeted coarse-to-fine parsing for accurate and efficient document understanding.
The source code and models are publicly available at \url{https://github.com/PaddlePaddle/PaddleOCR}.

\vspace{-5pt}
%

%

\end{abstract}    
\section{Introduction}
\label{sec:intro}


\begin{figure*}[htbp]
\centering
\includegraphics[width=\linewidth]{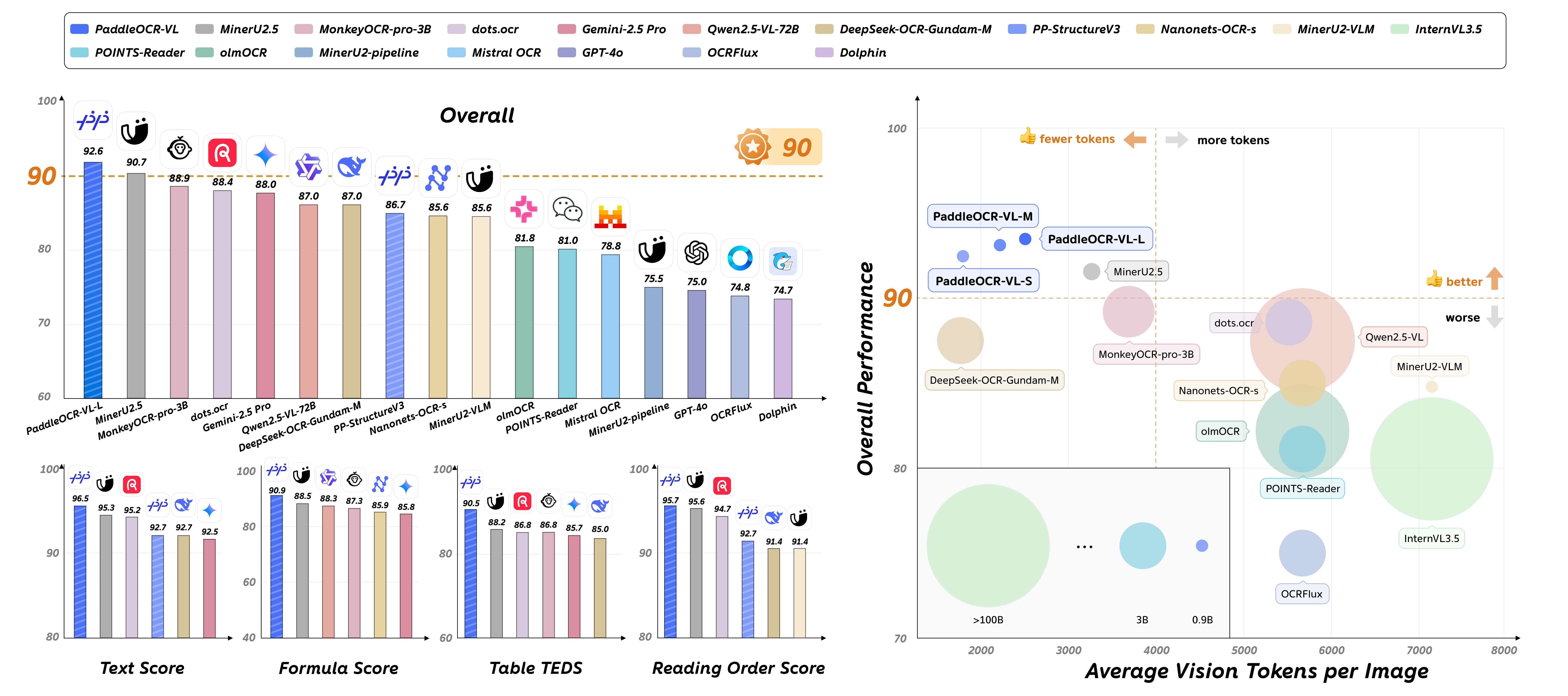} %
\caption{
    \centering PaddleOCR-VL achieves state-of-the-art performance with the fewest vision tokens and parameters on OmniDocBench v1.5.
}
\label{fig:Performance}
\end{figure*}

Documents serve as core information carriers, with their complexity and volume growing at an exponential rate, making document parsing an indispensable key technology. The primary goal of document parsing~\cite{li2025monkeyocr, niu2025mineru2, feng2025dolphin, liu2025points} is to enable deep structural and semantic understanding of documents. Specifically, it involves recognizing the document elements (text, formulas, tables, charts, images, etc.) and determining the correct reading order. Document parsing plays a crucial role in constructing training corpora for large language models (LLMs)~\cite{ernie2025technicalreport, yang2025qwen3,achiam2023gpt} and applications like RAG (Retrieval-Augmented Generation)~\cite{lewis2020retrieval}, making its accuracy and efficiency a major research focus.

Current document parsing methods can be grouped into three families: pipeline-based approaches, general vision-language models (VLMs), and specialized VLM approaches. Pipeline-based methods~\cite{wang2024mineru,cui2025paddleocr} connect expert components such as region detection, text recognition, and structure reconstruction to parse layouts; they work well on simple pages but are prone to error propagation and struggle with complex layouts, which limits applicability~\cite{li2025monkeyocr}. General VLMs, enabled by large-scale multimodal pretraining~\cite{bai2025qwen2,gemini25,liu2023visual}, have recently shown strong OCR and document-understanding capabilities without handcrafted modules; nevertheless, on handwritten or highly structured documents they frequently exhibit hallucinations, severe recognition errors, and reading order confusion, and their computational cost and latency increase with the number of visual tokens required by high-resolution pages. Specialized VLM approaches tailor the multimodal architecture to documents: end-to-end models~\cite{MinerU2,dotsocr} process entire pages to capture page-level semantics, avoiding some generalization limits of pipelines but typically relying on large parameter counts and encountering layout degradation and efficiency bottlenecks on long documents~\cite{feng2025dolphin}; VLM assembled pipelines~\cite{MinerU2,feng2025dolphin} perform layout parsing followed by element-level recognition, which can suffer from coordinate drift on densely populated pages and, at the layout stage, higher computational cost than task-specific detectors~\cite{sun2025pp}. To mitigate efficiency constraints, recent work reduces visual tokens for high-resolution inputs~\cite{wei2025deepseek}; however, uniform compression across salient and non-salient regions often degrades fine-grained layout accuracy.

While high-resolution inputs are essential for document parsing where fine-grained textual and structural details such as small and dense text, tables, and formulas significantly affect accuracy, they also lead to a quadratic increase in visual encoding and token processing costs. Notably, visual information in documents tends to be highly unevenly distributed: regions dense with semantic content coexist with large areas of redundant backgrounds or decorative elements. As illustrated in Figure~\ref{fig:architectural}, based on our analysis on OmniDocBench v1.5, valid visual regions in PowerPoint documents account for only about 39\% of the total area, whereas in information-dense documents such as newspapers, they occupy roughly 60\%. 
Motivated by the sparsity of valid visual regions in documents, we introduce PaddleOCR-VL, a coarse-to-fine document parsing framework that can eliminate visual redundancy while improving performance and efficiency. Specifically, the coarse stage employs a lightweight Valid Region Focus Module (VRFM) to identify and locate the valid visual elements efficiently. The fine stage then uses a compact vision-language model, PaddleOCR-VL-0.9B, to perform detailed recognition within the extracted valid visual regions. This decoupled hierarchical design allows independent optimization of layout and recognition modules, enabling PaddleOCR-VL to achieve impressive accuracy with a small model size while maintaining high efficiency.

In addition, we developed a high-quality data pipeline, collecting over 30M widely distributed samples from public sources and synthesis, which has become one of the most critical factors enabling our method to achieve high performance.

As illustrated in Figure~\ref{fig:Performance}, we conduct  a comprehensive evaluation on the public benchmark OmniDocBench v1.5~\cite{ouyang2025omnidocbench}. Our results demonstrate that our proposed solution surpasses existing document parsing methods, achieving state-of-the-art performance across all four critical metrics: text, formula, table, and reading order scores. Furthermore, the right panel of Figure~\ref{fig:Performance} highlights that PaddleOCR-VL utilizes the fewest vision tokens and parameters, resulting in substantially lower latency and higher throughput compared to competing approaches.


Our contributions are summarized as follows:
\begin{itemize}
\item We propose \textbf{PaddleOCR-VL}, a coarse-to-fine document parsing framework that allocates computation to semantically relevant regions while suppressing redundant ones. This paradigm effectively achieves both high-resolution accuracy and computational efficiency in multimodal document parsing.

\item We design a decoupled two-stage architecture consisting of a Valid Region Focus Module (VRFM) and PaddleOCR-VL-0.9B. The former performs lightweight layout detection and reading-order prediction to locate informative regions, while the latter conducts fine-grained recognition within these regions. This hierarchical design allows each stage to specialize, substantially improving both efficiency and recognition quality.

\item PaddleOCR-VL achieves state-of-the-art performance on multiple public and in-house benchmarks, excelling in recognizing complex elements such as text, tables, formulas, and charts. It supports 109 languages and demonstrates strong robustness on challenging content, including handwritten and historical documents.
\end{itemize}

\begin{figure*}[htbp]
\centering
\includegraphics[width=\linewidth]{images/method1.png} %
\caption{
    Architectural comparison of End-to-end VLM and our method. Among various types of document images, the valid area accounts for less than 50\% on average. Therefore, our proposed PaddleOCR-VL can achieve improvements in visual information processing by discarding redundant regions.   
}
\label{fig:architectural}
\end{figure*}

\section{Related Work}
\label{sec:related_work}


\subsection{Pipeline-based Methods}
Early works predominantly followed a pipeline paradigm~\cite{cui2025paddleocr,wang2024mineru}, in which document parsing was decomposed into several sequential stages. The process typically began with layout detection, identifying the spatial boundaries of layout elements such as text blocks, figures, and tables. Subsequently, a series of specialized expert models were applied to these detected regions to perform tasks like text recognition, table structure reconstruction, and mathematical formula recognition. Although this modular design enabled task-specific optimization, it required multiple independent models and resulted in a complex and cumbersome overall system.

\subsection{General VLM Methods}
Recent progress in general-purpose vision-language models has opened up new possibilities for document understanding~\cite{bai2025qwen2,gemini25,liu2023visual}. These models are typically pretrained on large-scale image–text datasets, enabling them to capture cross-modal correspondences between visual and linguistic signals within a unified framework. When applied to document images, they can interpret both layout and textual semantics without relying on explicit OCR or handcrafted layout modules. This allows for end-to-end processing and flexible adaptation across diverse tasks such as captioning, visual question answering, and even form understanding.

\subsection{Specialized VLM Methods}
Specialized Vision–Language Models have been proposed to address the limitations of general-purpose VLMs in document parsing~\cite{niu2025mineru2,feng2025dolphin,dotsocr,du2025unirec,wei2025deepseek,li2025monkeyocr}. Among them, MinerU2.5~\cite{niu2025mineru2} and Dolphin~\cite{feng2025dolphin} employ unified multimodal architectures that jointly perform detection and recognition, ensuring semantic consistency across stages and improving category generalization and extensibility to new element types. However, as their detection relies on generative or grounding-based multimodal outputs, the predicted coordinates may drift in densely arranged or lengthy documents, causing error propagation and degraded recognition accuracy. Other models such as MinerU2-VLM~\cite{MinerU2} and dots.ocr~\cite{dotsocr} further integrate layout understanding and element recognition into a single large-scale model through multimodal pretraining, simplifying the workflow but suffering from low visual-token efficiency and difficulties with long or complex documents, including reading-order confusion and hallucinations. Their computational cost also scales quadratically with image resolution and sequence length due to the attention mechanism. To mitigate this, DeepSeek-OCR~\cite{wei2025deepseek} introduces a visual token compression module that reduces computational overhead, but this improvement comes at the expense of fine-grained layout accuracy and does not resolve the latency from long decoding sequences.

\begin{figure*}[h]
\centering
\includegraphics[width=\linewidth]{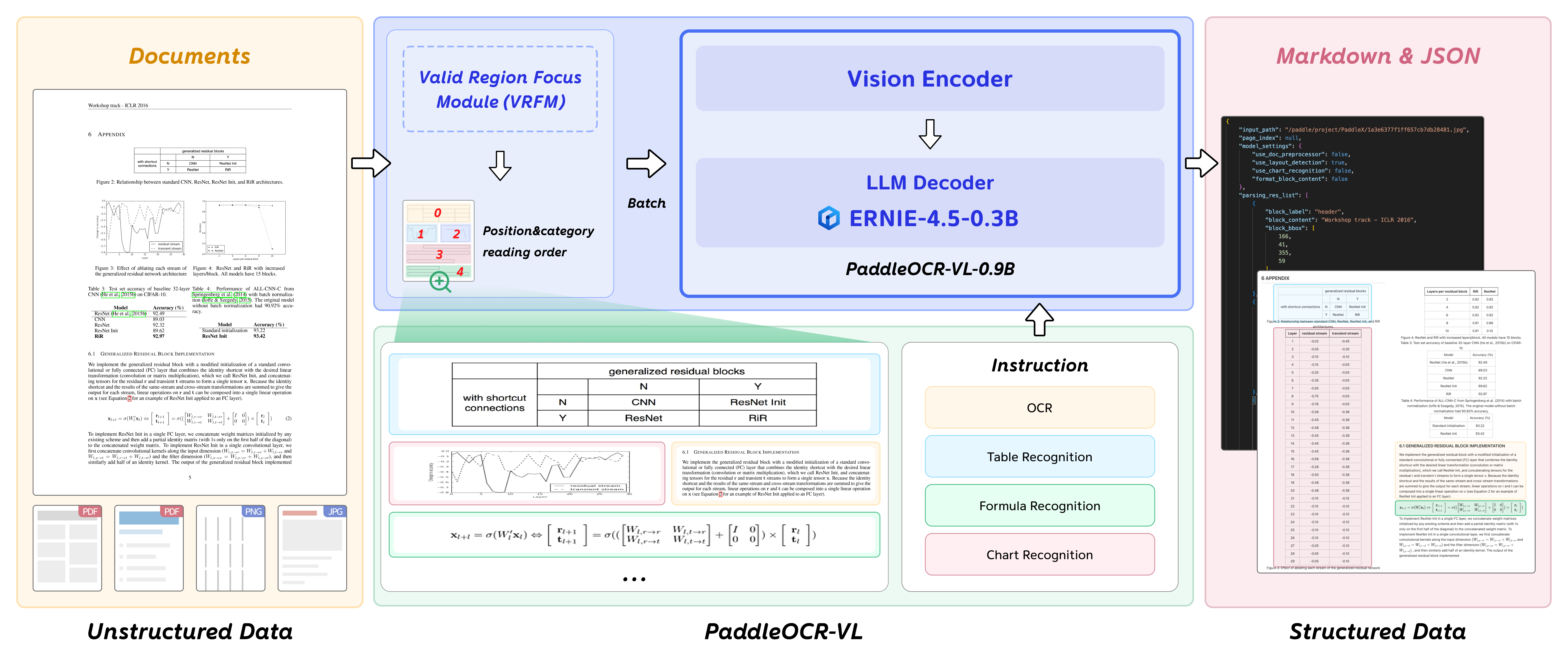} 
\caption{
    The overview of our proposed PaddleOCR-VL, which consists of two components. The first component is VRFM, which accurately extracts valid regions containing layout elements from unstructured document images and predicts the reading order. Subsequently, each valid region is fed into our designed PaddleOCR-VL-0.9B for precise recognition of individual elements. The final results are then reconstructed into a structured document according to the reading order.
}
\label{fig:model_overview}
\end{figure*}

\section{Method}

\subsection{PaddleOCR-VL}

As shown in Figure~\ref{fig:architectural}, redundant visual regions in document images cause end-to-end parsing models to generate excessive vision tokens, leading to substantially increased computational cost. To address this issue, we propose PaddleOCR-VL, a coarse-to-fine document parsing framework that decouples layout analysis from element recognition.

As illustrated in Figure~\ref{fig:model_overview}, PaddleOCR-VL consists of two stages. In the coarse stage, a lightweight Valid Region Focus Module (VRFM) identifies informative document regions, predicts their categories, and estimates their reading order. In the fine stage, the cropped valid regions are fed into PaddleOCR-VL-0.9B for element-level recognition. The recognized outputs are then merged according to the predicted reading order to reconstruct the final structured document.

This design offers two advantages. First, it avoids processing large irrelevant backgrounds and therefore reduces the number of visual tokens passed to the vision-language model. Second, by separating region localization from content recognition, each module can specialize in its own subtask, leading to better efficiency and stronger recognition performance. The details of VRFM and PaddleOCR-VL-0.9B are introduced in Sections~3.1.1 and~3.1.2, respectively.

\subsubsection{Valid Region Focus Module (VRFM)}

\begin{figure*}[h]
\centering
\includegraphics[width=\linewidth]{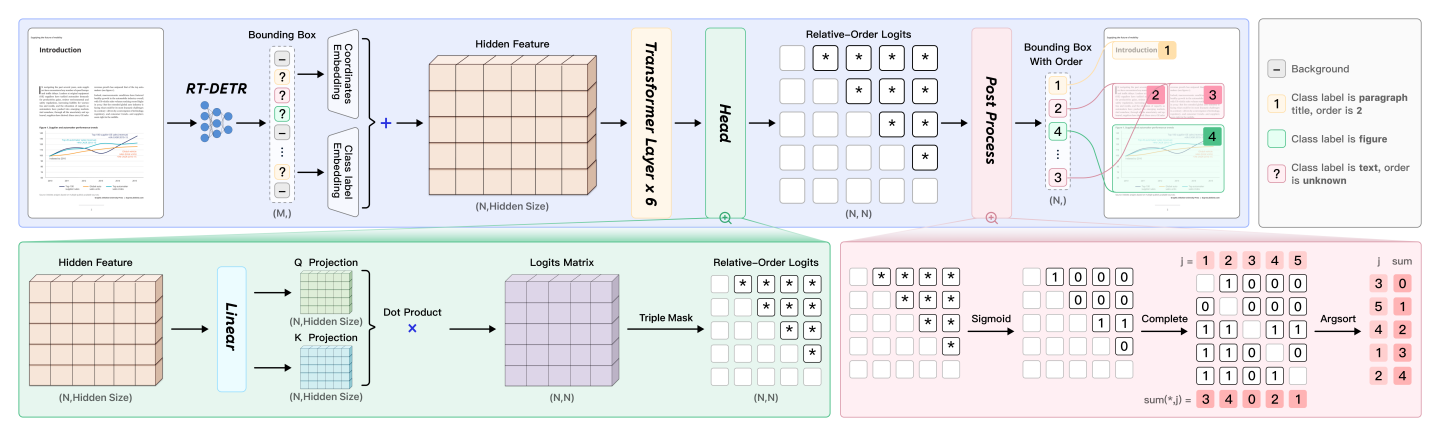} 

\caption{
    \centering
Architecture of the Valid Region Focus Module (VRFM).
}
\label{fig:pp-doclayoutv2}
\end{figure*}

The goal of VRFM is to efficiently locate visually informative regions in a document and determine their reading order before fine-grained recognition. As shown in Figure~\ref{fig:pp-doclayoutv2}, VRFM is built upon RT-DETR~\cite{zhao2024detrs} and is further extended with a pointer network for reading-order modeling.

Specifically, the RT-DETR backbone is used to detect and classify document elements, producing region-level representations for candidate layout components. On top of these representations, the pointer network models pairwise relationships among detected regions and predicts an $N \times N$ matrix that encodes their relative reading order. This design allows VRFM to jointly perform region localization, category prediction, and reading-order estimation within a unified lightweight framework.

By explicitly filtering out irrelevant background areas and retaining only task-relevant regions, VRFM provides compact and information-dense inputs for the downstream vision-language model. As a result, the overall system reduces redundant computation while preserving the structural information required for accurate document parsing.

\subsubsection{Element Recognition (PaddleOCR-VL-0.9B)}
\label{PaddleOCR-VL-0.9B arch}

We introduce PaddleOCR-VL-0.9B, an architecture designed for high accuracy with low computational overhead. Inspired by LLaVA~\cite{liu2023visual}, our model integrates a pre-trained vision encoder, an MLP projector, and a large language model, balancing their scales to optimize performance.
A key distinction from prior methods~\cite{liu2025points,MinerU2,wei2024general} that use fixed-resolution or tiling is our native dynamic-resolution processing. We employ a NaViT-style~\cite{dehghani2023patch} vision encoder, initialized from Keye-VL~\cite{team2025kwai}, which processes images at their native resolution. This approach avoids distortion, reduces hallucinations, and improves performance on text-intensive tasks. A 2-layer MLP projector with GELU~\cite{hendrycks2016gaussian} activation efficiently bridges the visual and language modalities.
To ensure low inference latency, we select the compact yet efficient ERNIE-4.5-0.3B~\cite{ernie2025technicalreport} as our language model, further enhancing it with 3D-RoPE~\cite{bai2025qwen2} for positional representation. This combination of NaViT and ERNIE-4.5-0.3B yields significant gains in document parsing, achieving faster inference with a minimal memory footprint.

\subsection{Dataset}

As a core factor in achieving SOTA model performance, we propose a systematic methodology for constructing such datasets, which gathers a diverse set of data to ensure comprehensive coverage from multiple sources as follows.



\subsubsection{Data Curation}
\label{Curation}
To ensure breadth and diversity, we construct our dataset from four primary sources: open-source, synthesized, web-crawled, and in-house data.

\textbf{Open-Source Datasets.} We build upon a foundation of established public datasets. This includes text from CASIA-HWDB~\cite{liu2011casia}, mathematical expressions from UniMER-1M~\cite{unimernet} and MathWriting~\cite{MathWriting}, and charts from datasets such as ChartQA~\cite{chartqa}, PlotQA~\cite{plotqa}, Unichart~\cite{masry2023unichart}, and others~\cite{chart2text,dvqa,beagle,chartinfo,tang2023vistext,excelchart}. All data is rigorously filtered to remove low-quality annotations.

\textbf{Synthesized Datasets.} To address the inherent imbalance in public data, we employ a cost-effective synthesis strategy to generate large volumes of underrepresented data types, promoting unbiased model performance.

\textbf{Network-Accessible Datasets.} To enhance generalization and robustness, we collect a diverse corpus of real-world documents from the internet. This collection, including academic papers, newspapers, and handwritten scans, exposes the model to varied layouts and styles, mitigating overfitting to canonical data.

\textbf{In-house Datasets.} Finally, we incorporate our extensive in-house datasets, accumulated over years of OCR research. These datasets, covering a wide range of document parsing tasks, are added in controlled proportions and are crucial for the model’s state-of-the-art performance.

\subsubsection{Automatic Data Annotation}
\label{Automatic Data Annotation}
We employ an automatic pipeline for large-scale data annotation. First, the expert model PP-StructureV3~\cite{cui2025paddleocr} generates initial, potentially noisy pseudo-labels. We then leverage these labels and the original images to prompt advanced vision-language models---ERNIE-4.5-VL~\cite{ernie2025technicalreport} and Qwen2.5-VL~\cite{bai2025qwen2} for refinement. Finally, a hallucination filtering step eliminates potential model-generated errors, yielding high-quality labels.

\subsubsection{Hard Cases Mining}
\label{Hard Cases Mining}
To address performance bottlenecks, we propose a hard case mining pipeline. We first construct a diverse, manually annotated evaluation set covering text, tables, formulas, and charts, each with fine-grained sub-categories (e.g., handwritten text, rotated tables). By evaluating our model on this set using specialized metrics—EditDist for text, TEDS~\cite{teds} for tables, RMS-F1~\cite{deplot} for charts, and BLEU~\cite{bleu_score} for formulas—we accurately identify its weaknesses. Finally, we leverage rendering tools like XeLaTeX and web browsers with rich assets (e.g., font libraries, corpora) to synthetically generate new, high-quality data targeting these identified hard cases.

\subsection{Training Recipe}
The following sections introduce the training details of these two modules: Valid Region Focus Module, which leverages text-aware region localization capabilities to identify valid visual tokens, and the PaddleOCR-VL-0.9B vision-language model for valid region element recognition. 

\subsubsection{Valid Region Focus Module (VRFM)}
\label{Layout Analysis}


We employ VRFM for layout localization, classification, and reading order prediction. The model extends RT-DETR~\cite{zhao2024detrs} with a pointer network~\cite{hou2024relation} to predict element ordering.
Our training follows a two-stage process. First, we train the RT-DETR core for layout detection. It is initialized with PP-DocLayout\_Plus-L~\cite{sun2025pp} weights and trained for 100 epochs on our 20k+ sample dataset. Next, we freeze the core and train only the pointer network for 200 epochs. This stage learns a pairwise ordering matrix using the noise-robust Generalized Cross Entropy Loss~\cite{zhang2018generalized}, the AdamW optimizer, and a constant learning rate of 2e-4.

\subsubsection{Element-level Recognition}
\label{Element-level Recognition}

PaddleOCR-VL-0.9B is a vision-language Model composed of a vision encoder, a projector, and a language model. We adopt a post-adaptation strategy, initializing the vision encoder with Keye-VL and the language model with ERNIE-4.5-0.3B weights. The training, built upon ERNIEKit~\cite{ERNIEkit}, is divided into two stages.
\begin{table}[!h]
\centering
 \fontsize{7}{7}\selectfont
\renewcommand{\arraystretch}{1.2}
\begin{tabular}{l|cc}
\toprule
\textbf{Stages} & \textbf{Stage 1} & \textbf{Stage 2} \\  \midrule
Training Samples & 29M & 2.7M \\
Max Resolution & 1280 $\times$ 28 $\times$ 28 & 2048 $\times$ 28 $\times$ 28 \\
Sequence length & 16384 & 16384 \\
Trainable components & All & All \\
Batch sizes & 128 & 128 \\
Data Augmentation & Yes & Yes \\
Maximum LR & $5 \times 10^{-5}$ & $5 \times 10^{-6}$ \\
Minimum LR & $5 \times 10^{-6}$ & $5 \times 10^{-7}$ \\
Epoch & 1 & 2 \\ \bottomrule
\end{tabular}
\caption{Training settings of PaddleOCR-VL-0.9B.}
\label{training}
\end{table}
\textbf{Stage 1: Pre-training Alignment.} We first align the vision and language modalities by pre-training for one epoch on 29 million image-text pairs. This stage uses a batch size of 128, a sequence length of 16384, and a learning rate scheduled from $5 \times 10^{-5}$ down to $5 \times 10^{-6}$.

\textbf{Stage 2: Instruction Fine-tuning.} We then fine-tune the model on a diverse 2.7-million-sample dataset for two epochs to adapt it for specific document intelligence tasks. For this stage, we increase the max number of vision tokens to 2048 and apply a finer learning rate ($5 \times 10^{-6}$ to $5 \times 10^{-7}$). 
The fine-tuning tasks include:
\begin{enumerate}[label=\arabic*.]
\item \textbf{OCR:} Recognizing text at various granularities, from characters to page-level layouts.
\item \textbf{Table Recognition:} Parsing tables into the structured OTSL~\cite{lysak2023optimized} format.
\item \textbf{Formula Recognition:} Transcribing visual formulas into \LaTeX, distinguishing between inline \textbackslash(…\textbackslash) and display \textbackslash[…\textbackslash] styles.
\item \textbf{Chart Recognition:} Extracting data from various chart types (e.g., bar, line) into Markdown tables.
\end{enumerate}

\section{Evaluation}

\begin{table*}[h]
    \centering
    \resizebox{\textwidth}{!}{%
    \renewcommand{\arraystretch}{1.2}
    \begin{tabular}{l|ll|c|c c c c c c} 
        \toprule
        \textbf{Model Type} & \textbf{Methods} & \textbf{Parameters} & \textbf{Vision Tokens} & \textbf{Overall$\uparrow$} & \textbf{Text\textsuperscript{Edit}$\downarrow$} & \textbf{Formula\textsuperscript{CDM}$\uparrow$} & \textbf{Table\textsuperscript{TEDS}$\uparrow$} & \textbf{Table\textsuperscript{TEDS-S}$\uparrow$} & \textbf{Reading Order\textsuperscript{Edit}$\downarrow$} \\    \midrule
        \multirow{3}{*}{\textbf{Pipeline Tools}} & Marker-1.8.2~\cite{vik2024marker} & - & - & 71.30 & 0.206 & 76.66 & 57.88 & 71.17 & 0.250 \\
        & Mineru2-pipeline~\cite{MinerU2} & - & - & 75.51 & 0.209 & 76.55 & 70.90 & 79.11 & 0.225 \\
        & PP-StructureV3~\cite{cui2025paddleocr} & - & - & 86.73 & 0.073 & 85.79 & 81.68 & 89.48 & 0.073 \\
        \midrule
        \multirow{5}{*}{\textbf{General VLMs}} & GPT-4o~\cite{achiam2023gpt} & - & - & 75.02 & 0.217 & 79.70 & 67.07 & 76.09 & 0.148 \\
        & InternVL3-76B~\cite{zhu2025internvl3} & 76B & 7277 & 80.33 & 0.131 & 83.42 & 70.64 & 77.74 & 0.113 \\
        & InternVL3.5-241B~\cite{wang2025internvl35} & 241B & 7277 & 82.67 & 0.142 & 87.23 & 75.00 & 81.28 & 0.125 \\
        & Qwen2.5-VL-72B~\cite{bai2025qwen2} & 72B & 5626 & 87.02 & 0.094 & 88.27 & 82.15 & 86.22 & 0.102 \\
        & Gemini-2.5 Pro~\cite{gemini25} & - & - & 88.03 & 0.075 & 85.82 & 85.71 & 90.29 & 0.097 \\
        \midrule
        \multirow{16}{*}{\textbf{Specialized VLMs}} & Dolphin~\cite{feng2025dolphin} & 322M & - & 74.67 & 0.125 & 67.85 & 68.70 & 77.77 & 0.124 \\
        & OCRFlux-3B~\cite{OCRFlux2025} & 3B & 5626 & 74.82 & 0.193 & 68.03 & 75.75 & 80.23 & 0.202 \\
        & Mistral OCR~\cite{mistral} & - & - & 78.83 & 0.164 & 82.84 & 70.03 & 78.04 & 0.144 \\
        & POINTS-Reader~\cite{liu2025points} & 3B & 5626 & 80.98 & 0.134 & 79.20 & 77.13 & 81.66 & 0.145 \\
        & olmOCR-7B~\cite{poznanski2025olmocr} & 7B & 5626 & 81.79 & 0.096 & 86.04 & 68.92 & 74.77 & 0.121 \\
        & MinerU2-VLM~\cite{MinerU2} & 0.9B & 7277 & 85.56 & 0.078 & 80.95 & 83.54 & 87.66 & 0.086 \\
        & Nanonets-OCR-s~\cite{Nanonets-OCR-S} & 3B & 5626 & 85.59 & 0.093 & 85.90 & 80.14 & 85.57 & 0.108 \\
        & DeepSeek-OCR-Gundam-M & 3B & 1854 & 86.46 & 0.081 & 89.45 & 78.02 & 81.55 & 0.093 \\
        & MonkeyOCR-pro-1.2B~\cite{li2025monkeyocr} & 1.9B & 3962 & 86.96 & 0.084 & 85.02 & 84.24 & 89.02 & 0.130 \\
        & MonkeyOCR-3B~\cite{li2025monkeyocr} & 3.7B & 3962 & 87.13 & 0.075 & 87.45 & 81.39 & 85.92 & 0.129 \\
        & dots.ocr~\cite{dotsocr} & 3B & 5513 & 88.41 & 0.048 & 83.22 & 86.78 & 90.62 & 0.053 \\
        & MonkeyOCR-pro-3B~\cite{li2025monkeyocr} & 3.7B & 3962 & 88.85 & 0.075 & 87.25 & 86.78 & 90.63 & 0.128 \\
        & MinerU2.5~\cite{niu2025mineru2} & 1.2B & 3256 & 90.67 & \cellcolor{cyan!15}\underline{0.047} & 88.46 & 88.22 & 92.38 & \cellcolor{cyan!15}\underline{0.044} \\ 
        & \textbf{PaddleOCR-VL-S} & 0.9B & 1898 & 91.55 & \cellcolor{red!15}\textbf{0.035} & \cellcolor{cyan!15}\underline{90.30} & 
        87.89 & 
        92.31 & 
        \cellcolor{cyan!15}\underline{0.044} \\
        & \textbf{PaddleOCR-VL-M} & 0.9B & 2259 & \cellcolor{cyan!15}\underline{92.17} & \cellcolor{red!15}\textbf{0.035} & 90.22 & \cellcolor{cyan!15}\underline{89.75} & \cellcolor{cyan!15}\underline{93.72} & \cellcolor{red!15}\textbf{0.043} \\
        & \textbf{PaddleOCR-VL-L} & 0.9B & 2561 & \cellcolor{red!15}\textbf{92.62} & \cellcolor{red!15}\textbf{0.035} & \cellcolor{red!15}\textbf{90.90} & \cellcolor{red!15}\textbf{90.48} & \cellcolor{red!15}\textbf{94.19} & \cellcolor{red!15}\textbf{0.043} \\
        \bottomrule
    \end{tabular}%
    }
   \caption{Comprehensive evaluation of document parsing on OmniDocBench v1.5. Results are reported by OmniDocBench~\cite{ouyang2025omnidocbench} unless Ours. In PaddleOCR-VL, S, M, and L refer to the same PaddleOCR-VL weights using different vsion tokens.}
   \label{tab:omni15_performance}
\end{table*}

\label{sec:experiments} 

To thoroughly assess the effectiveness of PaddleOCR-VL, we compared it against leading general vision language models and specialized document parsing models across multiple public benchmarks and in-house benchmarks. We conducted comprehensive performance comparisons in two aspects: page-level document parsing and element-level recognition, which are detailed in Sections \ref{Page-level Evaluation} and \ref{subsec:element_level_evaluation}. Page-level evaluation involves analyzing entire document pages to parse their overall content, structure, and layout, while element-level is dedicated exclusively to assessing the recognition of specific elements, such as text, tables, formulas, and charts, within the document.

\subsection{Page-level Evaluation}
\label{Page-level Evaluation}

This section details the evaluation of end-to-end document parsing capabilities using the following three benchmarks, aiming to measure its overall performance in real-world document scenarios. 

\noindent\textbf{OmniDocBench v1.5} To comprehensively evaluate the document parsing capabilities, we conducted extensive experiments on the OmniDocBench v1.5~\cite{ouyang2025omnidocbench} benchmark. It is an expansion of version v1.0, adding 374 new documents for a total of 1,355 document pages. It features a more balanced distribution of data in both Chinese and English, as well as a richer inclusion of all elements. The evaluation method has been updated, with formulas assessed using the CDM method. The overall metric is a weighted combination of the metrics for text, formulas, and tables. 

Table~\ref{tab:omni15_performance} shows that the PaddleOCR-VL series models achieve SOTA performance across all key metrics, surpassing existing pipeline tools, general VLMs, and other specialized document parsing models, despite having significantly fewer visual tokens. Our series of models, S, M, and L, are configured to accommodate different input visual resolution ranges. Specifically, the S model is suitable for image resolutions within the range of [3136, 235200], the M model supports resolutions from [3136, 392000], and the L model is designed for resolutions ranging from [3136, 627200]. Specifically, PaddleOCR-VL-L achieves a top-ranking overall score of 92.62 with 2561 vision tokens, surpassing the next best model, MinerU2.5 (90.67) with 3256 vision tokens. Compared to the DeepSeek-OCR-Gundam-M model with a similar number of visual tokens, our method achieves overall metrics that are more than 6 points higher. Moreover, our model establishes new SOTA results in the sub-tasks, including the lowest Text-Edit distance~\cite{lcvenshtcin1966binary} of 0.035, the highest Formula-CDM score of 90.90, the leading scores of 90.48 and 94.19 in Table-TEDS and Table-TEDS-S, and the best reading order scores of 0.043, respectively. These results underscore its superior accuracy in text recognition, formula recognition, and complex table structure analysis.

\subsection{Element-level Evaluation}
\label{subsec:element_level_evaluation}

Besides end-to-end document parsing performance, another advantage of our two-stage design is the ability to precisely focus on individual document elements. By isolating each element region via VRFM before feeding into the VLM, the model can dedicate its full capacity to recognizing each element type independently. To validate this advantage, we conduct element-level evaluations on text, tables, formulas, and charts using both public and in-house benchmarks.

\subsubsection{Table Recognition} For table recognition, we utilize OmniDocBench-Table-block to validate the effectiveness of PaddleOCR-VL-0.9B based on TEDS~\cite{teds} and Edit Distance.
\begin{table}[!htpb] 
    \centering
\fontsize{7}{7}\selectfont

    \setlength{\tabcolsep}{2pt} 
      \renewcommand{\arraystretch}{1.2}
    \begin{tabular}{l|ccc}
\toprule[.9pt]
\textbf{Methods} & \textbf{Overall TEDS$\uparrow$} & \textbf{Structural TEDS$\uparrow$} & \textbf{Overall Edit Dist$\downarrow$} \\
\midrule
MinerU2-VLM~\cite{MinerU2} &0.9002 &0.9369 &0.0734 \\


dots.ocr~\cite{dotsocr} &0.8194 &0.8442 & 0.1508 \\

MinerU2.5~\cite{niu2025mineru2} & \cellcolor{cyan!15}\underline{0.9005} &\cellcolor{red!15}\textbf{0.9539} & \cellcolor{cyan!15}\underline{0.0693} \\

\textbf{PaddleOCR-VL-L} &\cellcolor{red!15}\textbf{0.9046} &\cellcolor{cyan!15}\underline{0.9420} &\cellcolor{red!15}\textbf{0.0681} \\

\bottomrule
\end{tabular}
\caption{Comparison of OmniDocBench-Table-block Performance}
\vspace{-10pt}
\label{result_on_omnidoc_table_v1_5}
\end{table}

\noindent\textbf{OmniDocBench-Table-block:} To evaluate table recognition performance, we used the OmniDocBench-Table-block benchmark, with metrics including TEDS and Edit Distance. As shown in Table~\ref{result_on_omnidoc_table_v1_5}, PaddleOCR-VL-L achieves the best overall performance among all compared methods. Specifically, it obtains the highest Overall TEDS of 0.9046 and the lowest Overall Edit Distance of 0.0681, indicating superior performance in capturing both table structure and content while reducing recognition errors. Although MinerU2.5 achieves the best Structural TEDS of 0.9539, PaddleOCR-VL-L remains more competitive in overall table recognition accuracy. These results demonstrate the effectiveness of PaddleOCR-VL-L for table recognition on OmniDocBench-Table-block.

\subsubsection{Text Recognition} 

For text recognition, we utilize OmniDocBench-OCR-block to validate the effectiveness of models based on the edit distance metric. 

\noindent\textbf{OmniDocBench-OCR-block:} From the 1355 images of OmniDocBench v1.5, we extracted all text-related sub-images based on layout detection labels, removing any with null annotations. This process resulted in a total of 17,148 block-level images. This evaluation set is named OmniDocBench-OCR-block, with the ground truth still sourced from OmniDocBench. This evaluation set can more accurately assess the model’s text recognition performance without being affected by layout detection. We use the average normalized edit distance for evaluation.
        
In Table \ref{tab:performance_ocr_overall}, we present a comprehensive comparison of performance across various document types using different models. Our model, PaddleOCR-VL, consistently demonstrates superior performance, achieving the lowest error rates in almost all categories. Specifically, PaddleOCR-VL achieves the best results in the PPT2PDF (0.049), Academic Literature (0.021), Book (0.047), Colorful Textbook (0.082), Exam Paper (0.115), Magazine (0.020), Newspaper (0.035), Note (0.077), and Research Report (0.033) categories. These results highlight PaddleOCR-VL’s robust and versatile capability in handling diverse document types, establishing it as the leading method in the OmniDocBench-OCR-block performance evaluation.

\subsubsection{Formula Recognition} 
For formula recognition, we validate the effectiveness of our model based on the Character Detection Matching (CDM)~\cite{cdm} metric on OmniDocBench-Formula-block. 

\noindent\textbf{OmniDocBench-Formula-block:} Using the formula bounding boxes from OmniDocBench v1.5, 1050 formula sub-images were cropped. This step was taken to minimize the influence of layout detection on formula recognition. As shown in Table \ref{result_on_omnidoc_formula}, the model achieved a SOTA CDM score of 0.9404. 

\begin{table*}[!ht]
  \centering
  \setlength{\tabcolsep}{4pt}
  \renewcommand{\arraystretch}{1.2}
    \centering
    \fontsize{7}{7}\selectfont
  \begin{tabular}{l|c *{8}{c}} 
    \toprule
       \multirow{2}{*}{\textbf{Methods}} & \multicolumn{9}{c}{\textbf{Edit Distance $\downarrow$}}
      \\\cline{2-10} & \textbf{PPT} &
    \multicolumn{1}{p{1.4cm}}{\centering\textbf{Academic\newline Literature}} & 
    \textbf{Book} &
    \multicolumn{1}{p{1.4cm}}{\centering\textbf{Colorful\newline Textbook}} & 
    \multicolumn{1}{p{1.4cm}}{\centering\textbf{ Exam\newline Paper}} & 
    \textbf{Magazine} & \textbf{Newspaper} & \textbf{Note} &
    \multicolumn{1}{p{1.4cm}}{\centering\textbf{Research\newline Report}} \\ 
    \midrule
    Qwen2.5-VL-72B~\cite{bai2025qwen2}  & \cellcolor{cyan!15}\underline{0.054} & \cellcolor{cyan!15}\underline{0.023} & \cellcolor{cyan!15}\underline{0.061} & \cellcolor{cyan!15}\underline{0.084} & 0.195 & 0.032 & \cellcolor{cyan!15}\underline{0.056} & \cellcolor{cyan!15}\underline{0.118} & \cellcolor{cyan!15}\underline{0.040} \\
    MonkeyOCR-pro-3B~\cite{li2025monkeyocr} & 0.058 &  \cellcolor{red!15}\textbf {0.021} & 0.064 & 
    0.096 & \cellcolor{cyan!15}\underline{0.116} & \cellcolor{cyan!15}\underline{0.023} & 0.058 & 0.124 & 0.052 \\
    MinerU2.5~\cite{niu2025mineru2}   & 0.195 & 0.089 & 0.111 & 0.234 & 0.194 & 0.147 & \cellcolor{cyan!15}\underline{0.056} & 0.142 & 0.094 \\
    Dolphin~\cite{feng2025dolphin}  & 0.237 & 0.095 & 0.135 & 0.347 & 0.248 & 0.233 & 0.121 & 0.309 & 0.213 \\
    \textbf{PaddleOCR-VL-L}   &  \cellcolor{red!15}\textbf {0.049} & \cellcolor{red!15}\textbf {0.021} & \cellcolor{red!15}\textbf {0.047} & \cellcolor{red!15}\textbf {0.082} & \cellcolor{red!15}\textbf {0.115} & \cellcolor{red!15}\textbf {0.020} & \cellcolor{red!15}\textbf {0.035} &
     \cellcolor{red!15}\textbf {0.077} &
     \cellcolor{red!15}\textbf{0.031} \\
    \bottomrule
    \end{tabular}
  \caption{Overall Comparison of OmniDocBench-OCR-block Performance.}
  \label{tab:performance_ocr_overall}
\end{table*}

\begin{table}[H] 
    \centering
\fontsize{7}{7}\selectfont

    \setlength{\tabcolsep}{6pt} 
      \renewcommand{\arraystretch}{1.2}
    \begin{tabular}{l|ccc}
\toprule[.9pt]
\textbf{Methods} & \textbf{Overall CDM $\uparrow$} & \textbf{EN CDM $\uparrow$} & \textbf{ZH CDM $\uparrow$} \\
\midrule
dots.ocr~\cite{dotsocr} &0.4641 &0.4868 &0.4414 \\

MinerU2-VLM~\cite{MinerU2} &0.8286 &0.9616 &0.6956 \\

MonkeyOCR-pro-1.2B~\cite{li2025monkeyocr} &0.8531 &0.9642 &0.7419 \\

MonkeyOCR-3B ~\cite{li2025monkeyocr} &0.8621 &0.9718 &0.7524 \\

Qwen2.5-VL-72B~\cite{bai2025qwen2} &0.8747 &0.9574 &0.7920 \\

MinerU2.5~\cite{niu2025mineru2} &\cellcolor{cyan!15}\underline{0.9187} &\cellcolor{cyan!15}\underline{0.9751} &\cellcolor{cyan!15}\underline{0.8623} \\

\textbf{PaddleOCR-VL-L} &\cellcolor{red!15}\textbf{0.9404} & \cellcolor{red!15}\textbf{0.9773} &\cellcolor{red!15}\textbf{0.9035} \\

\bottomrule
\end{tabular}
\caption{Comparison of OmniDocBench v1.5 Formula-block Performance. Due to dots.ocr \cite{dotsocr} easily recognizing cropped formulas as images, the score is relatively low.}
\label{result_on_omnidoc_formula}
\vspace{-10pt}
\end{table}

\subsubsection{Chart Recognition} 
For chart recognition, we evaluate only on an in-house benchmark using RMS-F1~\cite{deplot}, due to the limited scale of existing datasets, imbalanced chart categories, and poor annotation quality in public test sets. The benchmark contains 1,801 manually verified samples from 11 chart categories, including 851 English and 950 Chinese charts, with both predictions and ground truth normalized into a unified markdown format. As shown in Table~\ref{tab:rms_f1_performance}, PaddleOCR-VL outperforms expert OCR VLMs and even a 72B-scale vision-language model.
\begin{table}[H]
\centering
\fontsize{7}{7}\selectfont

\setlength{\tabcolsep}{8pt}
 \renewcommand{\arraystretch}{1.2}
\begin{tabular}{@{} l| c c c @{}}
\toprule
\multirow{2}{*}{\textbf{Methods}} & \multicolumn{3}{c}{\textbf{RMS-F1 $\uparrow$}} \\
\cline{2-4}
& \textbf{Overall} & \textbf{EN} & \textbf{ZH} \\
\midrule
TinyChart~\cite{zhang2024tinychart}      & 0.2159          & 0.4726          & 0.0876 \\

GOT~\cite{wei2024general}              & 0.3160          & 0.1100          & 0.4190 \\
OneChart~\cite{onechart}      & 0.3716          & 0.1384          & 0.4882 \\

Qwen2.5-VL-72B~\cite{bai2025qwen2}    & 0.7300          & 0.6972          & 0.7464 \\

PP-StructureV3~\cite{cui2025paddleocr}   & \cellcolor{cyan!15}\underline{0.8060}          & \cellcolor{cyan!15}\underline{0.7963}         & \cellcolor{cyan!15}\underline{0.8109} \\
\textbf{PaddleOCR-VL-L}     & \cellcolor{red!15}\textbf {0.8440}          & \cellcolor{red!15}\textbf {0.8222}          & \cellcolor{red!15}\textbf {0.8549} \\
\bottomrule
\end{tabular}
\caption{Comparison of In-house-Chart Performance}
\label{tab:rms_f1_performance}
\vspace{-10pt}
\end{table}


\subsection{Inference Performance}



We evaluated end-to-end inference speed and GPU usage on the OmniDocBench dataset, processing PDF files in batches of 512 on a single NVIDIA A100 GPU. We compared high-accuracy solutions across three architectural designs: end-to-end VLM, VLM-based detection-recognition, and three-stage pipeline.

\begin{table}[!htpb]
  \centering
  \fontsize{7}{7}\selectfont
  \setlength{\tabcolsep}{3pt}
  \renewcommand{\arraystretch}{1.2}
  \begin{tabular}{l|ccccc}
    \toprule
    \textbf{Methods} & \textbf{\makecell{Total\\ Time (s)$\downarrow$}} & \textbf{Pages/s$\uparrow$} & \textbf{Tokens/s$\uparrow$} & \textbf{\makecell{Avg. VRAM \\Usage (GB)$\downarrow$}} \\
    \midrule
    MonkeyOCR-pro-1.2B~\cite{li2025monkeyocr} & 1456.4 & 0.6730 & 1120.3 & 75.5 \\
    dots.ocr~\cite{dotsocr} & 2784.6 & 0.3522 & 532.9 & 78.5 \\
    MinerU2.5~\cite{niu2025mineru2} & 927.3 & 1.0574 & \cellcolor{cyan!15}\underline{1647.9} & \cellcolor{red!15}\textbf{41.9} \\
    \textbf{PaddleOCR-VL-L} & \cellcolor{red!15}\textbf{605.2} & \cellcolor{red!15}\textbf{1.6192} & \cellcolor{red!15}\textbf{2470.7} & \cellcolor{cyan!15}\underline{42.1} \\
    \bottomrule
  \end{tabular}
  \caption{
    End-to-End Inference Performance Comparison. All models were tested using the vLLM 0.10.2 backend.
  }
  \label{tab:inference_performance}
\end{table}

As shown in Table~\ref{tab:inference_performance}, PaddleOCR-VL achieves superior speed and memory efficiency, validating our lightweight layout analysis combined with VLM architecture. With vLLM backend, it surpasses the strongest baseline MinerU2.5 by 53.1\% in page throughput and 49.9\% in token throughput, while reducing GPU memory by 46\% compared to dots.ocr. These results confirm that PaddleOCR-VL achieves SOTA inference efficiency with balanced speed-memory optimization, making it well-suited for real-world high-throughput document understanding.

\section{Conclusion}
PaddleOCR-VL is an efficient document parsing solution featuring two core components: the Valid Region Focus Module (VRFM) for layout analysis and PaddleOCR-VL-0.9B for high-speed recognition. By focusing only on valid document regions, it achieves state-of-the-art (SOTA) performance on complex elements (text, tables, formulas, charts) across 109 languages. Extensive experiments demonstrate that PaddleOCR-VL significantly outperforms existing methods in both accuracy and efficiency, providing lower latency and higher throughput for reliable AI applications.


{
    \small
    \bibliographystyle{ieeenat_fullname}
    \bibliography{main}
}

\clearpage
\setcounter{page}{1}
\maketitlesupplementary
\setcounter{section}{0}
\setcounter{table}{0}
\setcounter{figure}{0}
\renewcommand{\thesection}{\Roman{section}}
\renewcommand{\thetable}{S\arabic{table}}
\renewcommand{\thefigure}{S\arabic{figure}}

\section{Training Dataset Details}

This two-stage approach offers unique advantages in terms of data collection, as obtaining isolated element images along with their annotations is more feasible than collecting complete document pages containing different elements. In the following sections, we will elaborate on the construction of multimodal model training data for text, tables, formulas, and charts.

\subsection{Text}

We have curated a large-scale dataset comprising 20 Million High-Quality Image-Text Pairs. As shown in Figure~\ref{fig:text_dataset}, the dataset generation follows a rigorous multi-stage pipeline which primarily involves:

\begin{figure*}[h]
\centering
\includegraphics[width=\linewidth]{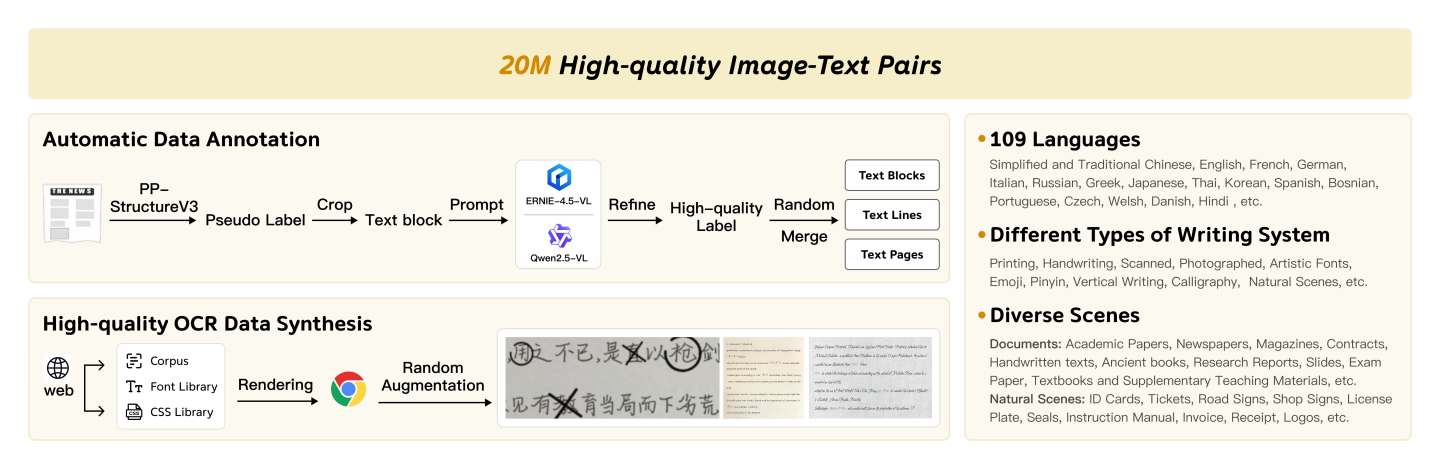} 

\caption{
    \centering
    The construction method and characteristics of the text training data for PaddleOCR-VL-0.9B. 
}
\label{fig:text_dataset}
\end{figure*}

\begin{enumerate}[label=\arabic*.]

\item \textbf{Automatic Data Annotation:} We design an automatic annotation pipeline that integrates lightweight document-structure models with large multimodal language models. Specifically, PP-StructureV3 is employed as an expert model to perform layout analysis and text recognition, generating pseudo labels that are converted into prompts for multimodal models such as ERNIE-4.5-VL and Qwen2.5-VL to refine. Finally, the refined labels are aggregated and randomly merged at multiple granularities to produce 20 million high-quality image–text training samples.

\item \textbf{High-quality OCR Data Synthesis:} During data distillation, low label quality in challenging scenarios like messy handwriting and dense blurry text was addressed by expanding the dataset through synthetic generation. Utilizing diverse CSS styles, over 200 fonts, and various corpora, we rendered a large amount of images, thereby enhancing the model’s capabilities in these difficult scenarios.

\end{enumerate}

Ultimately, the data is meticulously annotated at three distinct hierarchical levels: text lines, text blocks, and text pages. With extensive language coverage of 109 languages, including major global ones like Chinese, English, French, and Hindi. It includes diverse scenes including Academic Papers, Newspapers, Handwritten texts, Ancient books, Id cards, tickets, seals, etc. Additionally, the dataset addresses compatibility with a variety of writing systems and text styles, covering Printing, Handwriting, Scanned text, Artistic Fonts, etc.

\subsection{Table}

As shown in Figure \ref{fig:table_dataset}, we constructed a large-scale dataset of over $5$ million high-quality image-table pairs. Our dataset construction employs three key strategies: automatic data annotation, potential annotation mining, and high-quality data synthesis. For coding efficiency, we adopt OTSL~\cite{lysak2023optimized} as the model’s target format instead of conventional HTML. The main dataset construction process is as follows:
\begin{figure*}[h]
\centering
\includegraphics[width=\linewidth]{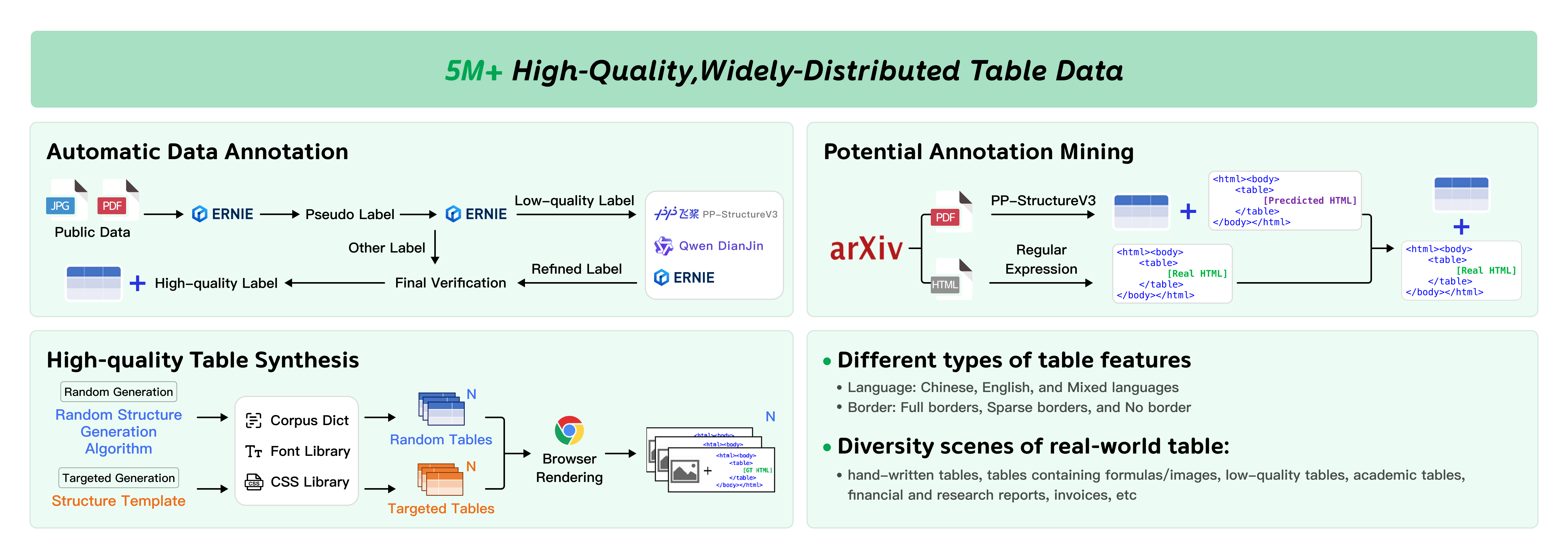} 

\caption{
    \centering
    The construction method and characteristics of the table training data for PaddleOCR-VL-0.9B.
}
\label{fig:table_dataset}
\end{figure*}
\begin{enumerate}
\item \textbf{Automatic Data Annotation:} To enhance the performance of PaddleOCR-VL in table recognition, we built a large-scale, diverse dataset covering various languages, border styles, and table types. Tables are first located using PP-StructureV3~\cite{cui2025paddleocr}. For unlabeled images, we employed a multi-stage annotation pipeline: ERNIE-4.5-VL~\cite{ernie2025technicalreport} first generates pseudo-labels, which are then validated by an ERNIE-4.5-VL-28B-A3B~\cite{ernie2025technicalreport} as discriminative model. Rejected annotations are refined using DianJin-OCR-R1~\cite{chen2025dianjin} (for tools, we use ERNIE-4.5-VL and PP-StructureV3~\cite{cui2025paddleocr}). Finally, all annotations undergo rigorous rule-based verification, including n-gram analysis and HTML validation, to ensure only high-quality samples are used for training.
\item \textbf{Potential Annotation Mining:}

For public data with potential annotations (e.g., from arXiv), we extract tables and their corresponding official-supported HTML source code. We then employ a mechanism combining regular expression matching with contextual and sequential alignment to construct accurate table-HTML pairs. The extracted HTML subsequently undergoes rule-based filtering, yielding high-quality data samples ready for model training.
\item \textbf{High-quality Table Synthesis:} 

To overcome data imbalance and high annotation costs, we introduce an innovative high-quality table synthesis tool which constitutes the cornerstone of our table data collection pipeline. This tool enables both randomized synthesis for comprehensive data supplement and targeted synthesis to enhance recognition of specific table categories. Specifically, we first leverage LLMs to gather a diverse and extensive corpus.Then, our tool generates table training pairs through randomized configurations of structures, fonts, CSS styles, and textual content, while also supporting customized synthesis by specifying particular parameters to accurately simulate specialized table types. With a synthesis speed of $10,000$ samples per hour, our tool has produced over $5,500,000$ training instances, substantially enhancing our model's generalization capability and comprehensive performance in table recognition.
\end{enumerate}

Through the aforementioned data construction strategies, we build a comprehensive table dataset encompassing diverse table categories and recognition scenarios, thereby providing robust support for training our model in the table recognition task.

\subsection{Formula}

\par As shown in Figure \ref{fig:formula_dataset}, this dataset was developed using a range of strategies, including source code rendering, automatic data annotation, targeted synthesis of long-tail data, and public data collection. It encompasses a variety of formula scenarios, such as educational supplementary materials, test papers for primary and secondary schools, mathematical papers, PowerPoint courseware, university theses, financial research reports, and handwritten mathematical notes. The dataset features four types of formulas: Simple Printed Expressions, Complex Printed Expressions, Screen-Captured Expressions, and Handwritten Expressions, available in both Chinese and English. The main process for constructing the dataset is as follows:
\begin{figure*}[h]
\centering

\includegraphics[width=\linewidth]{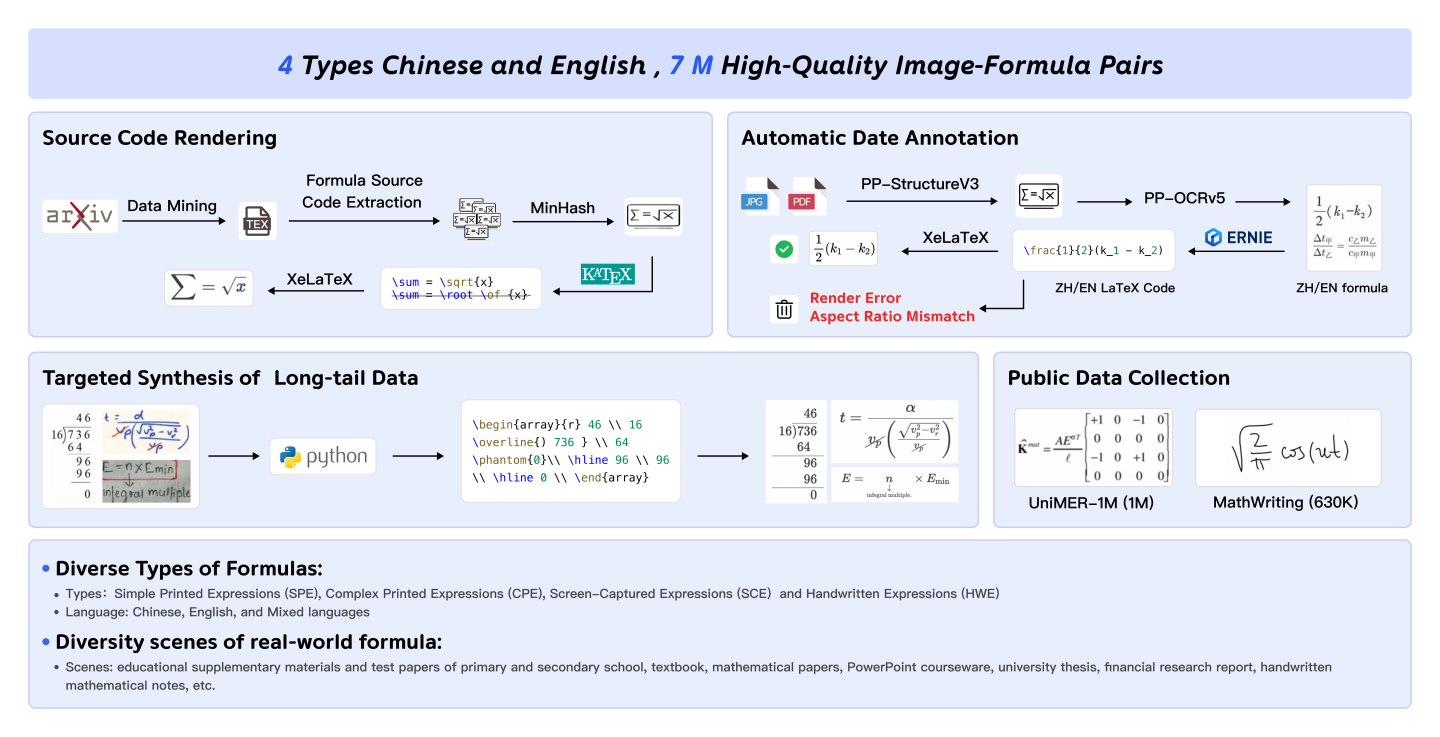} 

\caption{
    \centering
    The construction method and characteristics of the formula training data for PaddleOCR-VL-0.9B.
}
\label{fig:formula_dataset}
\end{figure*}
\begin{enumerate}[label=\arabic*.]
\item \textbf{Source Code Rendering:} To enhance the model's adaptability to a wide variety of unusual formula structures, a large amount of paper source code was scraped from arXiv, and LaTeX code for the formulas was extracted using regular expressions. Then, MinHash was used to remove duplicate and highly similar formula source codes, and KaTeX was employed to normalize the formula source codes, thereby reducing their ambiguity. Finally, the formulas were re-rendered into images using a formula rendering engine.

\item \textbf{Automatic Data Annotation:} For real-world formula data from exam papers, educational materials, and handwritten notes, the process begins with the use of the layout analysis method PP-StructureV3 \cite{cui2025paddleocr} to identify the bounding boxes for formulas. Based on these bounding boxes, formula regions are cropped from the images. Subsequently, large multimodal language models, such as ERNIE-4.5-VL-28B-A3B~\cite{ernie2025technicalreport}, are employed to generate the LaTeX source code for these formulas. Given the rarity of Chinese formulas in real-world scenarios—where approximately 1 out of 100 formulas contains Chinese characters—PP-OCRv5~\cite{cui2025paddleocr} is utilized to recognize characters within the cropped regions, enabling targeted optimization when Chinese characters are detected. Due to the complex and diverse nature of real-world formulas, recognition errors may occur with existing large models. To address this, a LaTeX rendering engine is used to filter the formulas generated by these models. Specifically, image-formula pairs that cannot be successfully rendered by xelatex are discarded. For those that render successfully, a more in-depth screening is conducted by comparing metrics such as the aspect ratio between the recognized image and the rendered image.

\item \textbf{Targeted Synthesis of Long-tail Data:}  For certain long-tail formula structures, such as elementary school vertical calculations, formulas with strikethroughs, and handwritten formulas with explanatory arrows, existing multimodal large models struggle to accurately recognize them due to data distribution issues. To address this, LaTeX code is synthetically generated based on rules and inverse rendering is performed using a LaTeX rendering engine, thereby constructing image-formula matching pairs for these long-tail scenarios.

\item \textbf{Public Data Collection:} In order to enable the model to learn high-quality formula representations, a substantial amount of data has been collected from existing public datasets, including  UniMER-1M~\cite{unimernet} and MathWriting~\cite{MathWriting}. Specifically,  UniMER-1M is oriented towards real document scenarios and has gathered 1 million formula data from arXiv, Pix2tex~\cite{pix2tex}, CROHME~\cite{chrome2014,mouchere2016icfhr2016,chrome2019}, and HME100K~\cite{yuan2022syntax}. On the other hand, MathWriting is currently the largest handwritten mathematical formula dataset, comprising 230,000 real handwritten formula samples and 400,000 synthetic handwritten formula samples.

\end{enumerate}

\subsection{Chart}

\par We constructed a large-scale, bilingual (Chinese and English) dataset of over 0.8 million high-quality image-chart pairs. Our dataset construction employs four key strategies: public data collection and cleaning, automatic data annotation, data synthesis, and targeted long-tail data augmentation. The dataset covers a wide array of chart types from diverse sources, including academic papers, financial reports, and web pages. The main dataset construction process is as follows:

\begin{figure*}[H]
\centering
\includegraphics[width=\linewidth]{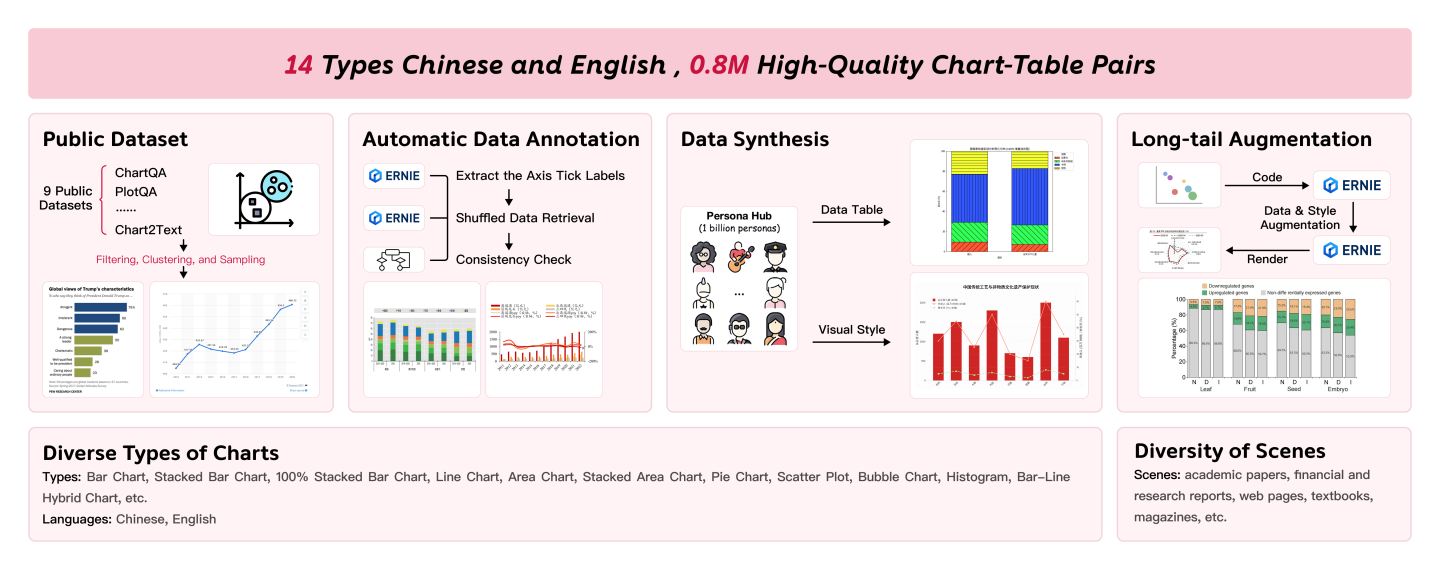} 

\caption{
    \centering
    The construction method and characteristics of the chart training data for PaddleOCR-VL-0.9B.
}
\label{fig:chart_dataset}
\end{figure*}

\begin{enumerate}[label=\arabic*.]

\item \textbf{Public Data Collection and Cleaning:} We collected a large number of samples from public datasets, including ChartQA~\cite{chartqa}, PlotQA~\cite{plotqa}, Chart2Text~\cite{chart2text}, DVQA~\cite{dvqa}, Unichart~\cite{masry2023unichart}, Beagle~\cite{beagle}, ChartINFO~\cite{chartinfo}, visText~\cite{tang2023vistext}, and ExcelChart~\cite{excelchart}. However, the raw datasets suffered from poor annotation quality and extremely imbalanced data distributions. Thus, a meticulous data cleaning and filtering pipeline was implemented to remove noisy samples and ensure balanced clustering, resulting in a high-quality dataset of 220k samples.

\item \textbf{Automatic Data Annotation:} To annotate our large collection of unlabeled public and in-house data, we developed a two-stage annotation pipeline based on the Vision Large Language Model ERNIE-4.5-VL~\cite{ernie2025technicalreport}. In the first stage, the model extracts tick labels from the x- and y-axes; in the second, random permutations of these labels are used to query corresponding data points, framing annotation as a data retrieval task. A final consistency check ensures that only verified annotations are included in the training set, guaranteeing high reliability.

\item \textbf{Data Synthesis:} To capture diverse visual styles and enhance model generalization, we designed a three-stage data synthesis pipeline. It begins with a large collection of base data tables, followed by an LLM Persona~\cite{ge2024scaling} strategy using ERNIE-X1~\cite{ernie2025technicalreport}, which diversifies table content and generates persona-specific rendering code. This enables control over chart aesthetics such as color, font, and layout. Leveraging a billion distinct personas, the pipeline produces highly varied data structures and visual styles, substantially improving PaddleOCR-VL’s generalization across real-world charts. For rendering, we employ matplotlib and seaborn.

\item \textbf{Targeted Long-tail Data Augmentation:} To improve generalization on real-world long-tail samples, we designed a data augmentation pipeline based on seed charts. It first selects long-tail samples by their distinctive visual features, then uses ERNIE-4.5-VL~\cite{ernie2025technicalreport} to replicate their rendering code. ERNIE-X1~\cite{ernie2025technicalreport}, guided by a specific persona~\cite{ge2024scaling}, further diversifies the code by altering data tables and visual styles. Executing the modified code produces new augmented charts with corresponding data tables.

\end{enumerate}

\par Through the four data construction strategies mentioned above, the final chart dataset covers a wide range of application scenarios and a rich variety of chart styles, providing strong support for the training of chart models.


\section{Supported Languages}

\begin{table}[!h]
  \fontsize{7}{7}\selectfont
  \setlength{\tabcolsep}{3pt}
  \renewcommand{\arraystretch}{1.2}
\centering
\begin{tabular}{>{\centering\arraybackslash}m{0.22\linewidth}|>{\centering\arraybackslash}m{0.75\linewidth}}
\toprule
\textbf{Language Category} & \textbf{Specific Languages} \\
\midrule
Chinese & Chinese \\
\midrule
English & English \\
\midrule
Korean & Korean \\
\midrule
Japanese & Japanese \\
\midrule
Thai & Thai\\
\midrule
Greek & Greek \\
\midrule
Tamil & Tamil \\
\midrule
Telugu & Telugu \\
\midrule
Arabic & Arabic, Persian, Uyghur, Urdu, Pashto, Kurdish, Sindhi, Balochi \\
\midrule
Latin & French, German, Afrikaans, Italian, Spanish, Bosnian, Portuguese, Czech, Welsh, Danish, Estonian, Irish, Croatian, Uzbek, Hungarian, Serbian (Latin), Indonesian, Occitan, Icelandic, Lithuanian, Maori, Malay, Dutch, Norwegian, Polish, Slovak, Slovenian, Albanian, Swedish, Swahili, Tagalog, Turkish, Latin, Azerbaijani, Kurdish, Latvian, Maltese, Pali, Romanian, Vietnamese, Finnish, Basque, Galician, Luxembourgish, Romansh, Catalan, Quechua \\
\midrule
Cyrillic & Russian, Belarusian, Ukrainian, Serbian (Cyrillic), Bulgarian, Mongolian, Abkhazian, Adyghe, Kabardian, Avar, Dargin, Ingush, Chechen, Lak, Lezgin, Tabasaran, Kazakh, Kyrgyz, Tajik, Macedonian, Tatar, Chuvash, Bashkir, Malian, Moldovan, Udmurt, Komi, Ossetian, Buryat, Kalmyk, Tuvan, Sakha, Karakalpak \\
\midrule
Devanagari & Hindi, Marathi, Nepali, Bihari, Maithili, Angika, Bhojpuri, Magahi, Santali, Newari, Konkani, Sanskrit, Haryanvi \\
\bottomrule
\end{tabular}
   \caption{Supported Languages}
       \label{tab:supported_language}
\end{table}
PaddleOCR-VL supports a total of 109 languages. Table~\ref{tab:supported_language} lists the correspondence between each language category and the specific supported languages.

\section{Additional Experiments}
\subsection{Ablation Study on VRFM}
\label{sec:supp_ablation}

To further analyze the architectural design of VRFM, we conduct controlled comparisons with several representative baselines, including an end-to-end vision--language model (Qwen3-VL-8B), a token-reduction approach based on region cropping (DocKylin~\cite{zhang2025dockylin} + Qwen3-VL-8B), and a detection-based pipeline with heuristic reading-order rules (PP-DocLayout).

As shown in Table~\ref{tab:ablation}, DocKylin reduces the number of visual tokens (5818 $\rightarrow$ 4316) but leads to a noticeable performance drop ($-1.85\%$), indicating that token reduction without explicit structural modeling may discard important layout information. PP-DocLayout achieves competitive token efficiency; however, its reliance on hand-crafted reading-order rules limits its generalization ability.

In contrast, VRFM jointly models region detection and reading order within a unified framework and explicitly preserves spatial relationships. As a result, it achieves both lower token consumption and higher accuracy. These results suggest that the performance gains stem from explicit document structure modeling rather than naive token pruning.

\vspace{-1em}
\begin{table}[htbp]
\centering
\small
\renewcommand{\arraystretch}{0.8}
\resizebox{\columnwidth}{!}{
\begin{tabular}{l|c|c}
\toprule
\textbf{Method} & \textbf{Visual Tokens} & \textbf{Acc.(\%)} \\
\midrule
Qwen3-VL-8B~\cite{bai2025qwen3} & 5818 & 89.60 \\
DocKylin~\cite{zhang2025dockylin} + Qwen3-VL-8B~\cite{bai2025qwen3} & 4316 & 87.75 \\
PP-DocLayout~\cite{sun2025pp} + Qwen3-VL-8B & 3876 & 89.30 \\
PP-DocLayout + PaddleOCR-VL-0.9B  & 2347 & 89.96 \\
VRFM + Qwen3-VL-8B & 4262 & 90.37 \\
\textbf{VRFM + PaddleOCR-VL-0.9B} & 2561 & \textbf{92.62} \\
\bottomrule
\end{tabular}
}
\vspace{-0.7em}
\caption{Comparison of VRFM with representative baselines in terms of token efficiency and accuracy.}
\label{tab:ablation}
\vspace{-1em}
\end{table}

\subsection{Multilingual Performance}
\label{sec:supp_multilingual}

To evaluate the generalization ability of our method, we further conduct experiments on multilingual document parsing across diverse scripts. The evaluation covers a wide range of languages, including Arabic, Korean, Tamil, Greek, Thai, Telugu, Devanagari, Cyrillic, Latin, and Japanese.

As shown in Table~\ref{tab:performance_multilingual}, our method consistently achieves superior performance across all evaluated languages, demonstrating strong robustness to diverse linguistic and script variations. In particular, it significantly outperforms existing methods on low-resource and structurally complex scripts such as Telugu and Tamil.

These results validate that PaddleOCR-VL-0.9B effectively captures language-agnostic structural representations, enabling robust performance across multilingual scenarios.

\begin{table}[h]
    \small
    \renewcommand{\arraystretch}{0.8}
    \resizebox{\columnwidth}{!}{
    \begin{tabular}{l|cccc}
    \toprule
    \textbf{Language} & Qwen2.5-VL-72B & Dolphin & MinerU2.5 & Ours \\
    \midrule
    Arabic & 0.405 & 0.682 & 0.978 & 0.122 \\
    Korean & 0.056 & 0.699 & 0.917 & 0.052 \\
    Tamil & 0.389 & 0.912 & 0.957 & 0.043 \\
    Greek & 0.165 & 0.691 & 0.661 & 0.135 \\
    Thai & 0.194 & 0.709 & 0.880 & 0.081 \\
    Telugu & 0.758 & 0.832 & 0.937 & 0.011 \\
    Devanagari & 0.164 & 0.818 & 0.915 & 0.097 \\
    Cyrillic & 0.220 & 0.549 & 0.832 & 0.109 \\
    Latin & 0.021 & 0.037 & 0.063 & 0.013 \\
    Japanese & 0.181 & 0.309 & 0.588 & 0.086 \\
    \bottomrule
    \end{tabular}
    }
    \caption{Multilingual OCR performance comparison (Edit Distance $\downarrow$).}
    \label{tab:performance_multilingual}
\end{table}

\section{Inference Performance on Different Hardware Configurations}

We measured the inference performance of PaddleOCR-VL on different hardware configurations, as summarized in Table~\ref{tab:inference_performance_hardware}. As observed, PaddleOCR-VL demonstrates stable and efficient inference performance across a wide range of hardware and backend configurations, showing that the system can flexibly adapt to diverse computing environments. Moreover, we are currently integrating the FastDeploy backend, which is expected to further enhance inference efficiency in future releases.

\begin{table}[!h]
  \centering
  \fontsize{7}{7}\selectfont
  \setlength{\tabcolsep}{3pt}
  \renewcommand{\arraystretch}{1.2}
  \begin{tabular}{l|c|ccccc}
    \toprule
    \textbf{Hardware} & \textbf{Backend} & \textbf{Total Time (s)$\downarrow$} & \textbf{Pages/s$\uparrow$} & \textbf{Tokens/s$\uparrow$} & \textbf{\makecell{Avg. VRAM \\Usage (GB)$\downarrow$}} \\
    \midrule
    \multirow{2}{*}{A100} & vLLM & 800.9 & 1.2241 & 1881.2 & 43.7 \\
                          & SGLang & 917.6 & 1.0684 & 1641.5 & 49.8 \\
    \midrule
    \multirow{2}{*}{A10}  & vLLM & 1238.0 & 0.7921 & 1217.2 & 14.1 \\
                          & SGLang & 1429.9 & 0.6858 & 1055.8 & 20.0 \\
    \midrule
    \multirow{2}{*}{RTX 3060} & vLLM & 2749.1 & 0.3568 & 548.2 & 11.9 \\
                              & SGLang & 2792.4 & 0.3513 & 540.8 & 11.8 \\
    \midrule
    RTX 5070 & vLLM & 1292.9 & 0.7584 & 1165.5 & 8.9 \\
    \midrule
    \multirow{2}{*}{RTX 4090D} & vLLM & 845.3 & 1.1597 & 1781.8 & 16.7 \\
                               & SGLang & 951.8 & 1.0303 & 1586.1 & 21.8 \\
    \bottomrule
  \end{tabular}%
    \caption{End-to-End Inference Performance}
      \label{tab:inference_performance_hardware}
\end{table}

\section{More Visualization Results}
\subsection{Real-world Samples}

This appendix showcases the parsing and recognition capabilities of our proposed algorithm across a variety of challenging scenarios.

\begin{figure}[H]
\centering
\includegraphics[width=0.84\linewidth]{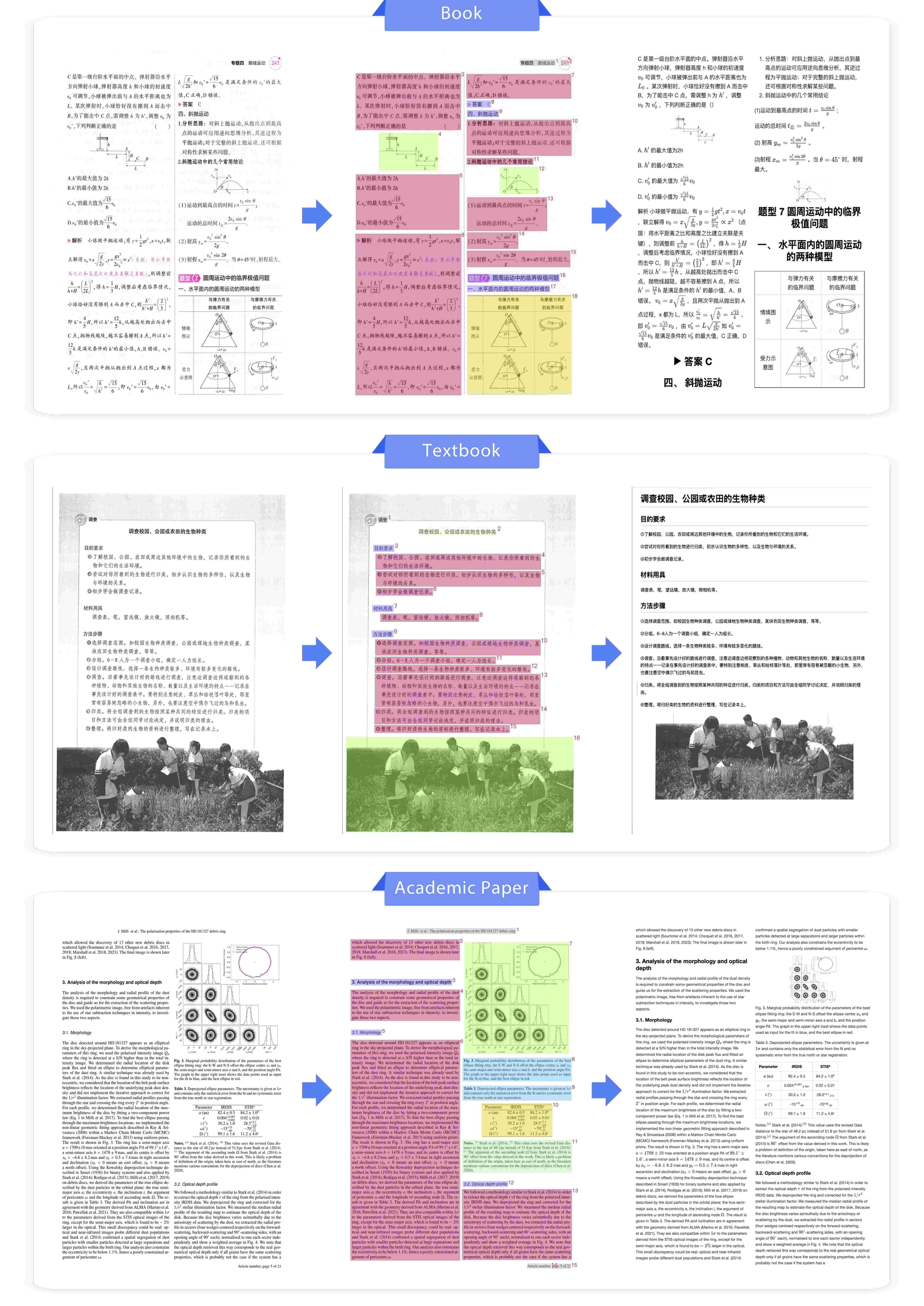} 

\caption{
    \centering
     The Layout and Markdown Output for Book, Textbook and Academic Paper.
}
\label{fig:overview1}
\end{figure}

\begin{figure}[t]
\centering
\includegraphics[width=0.86\linewidth]{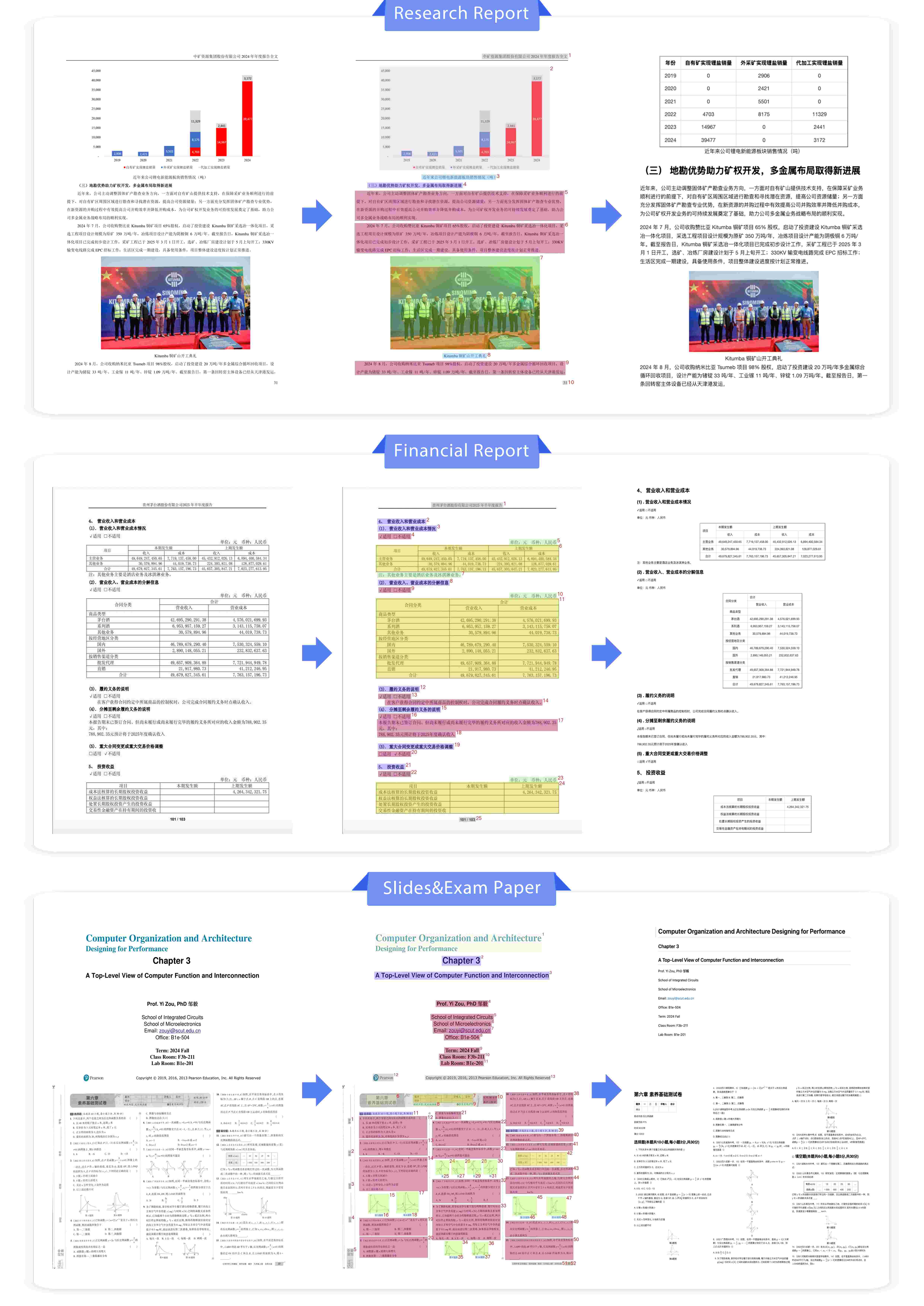} 

\caption{
    \centering
    The Layout and Markdown Output for Research Report(with chart recognition enabled), Financial Report, Slides and Exam Paper.
}
\label{fig:overview2}
\end{figure}

\begin{figure}[t]
\centering
\includegraphics[width=0.88\linewidth]{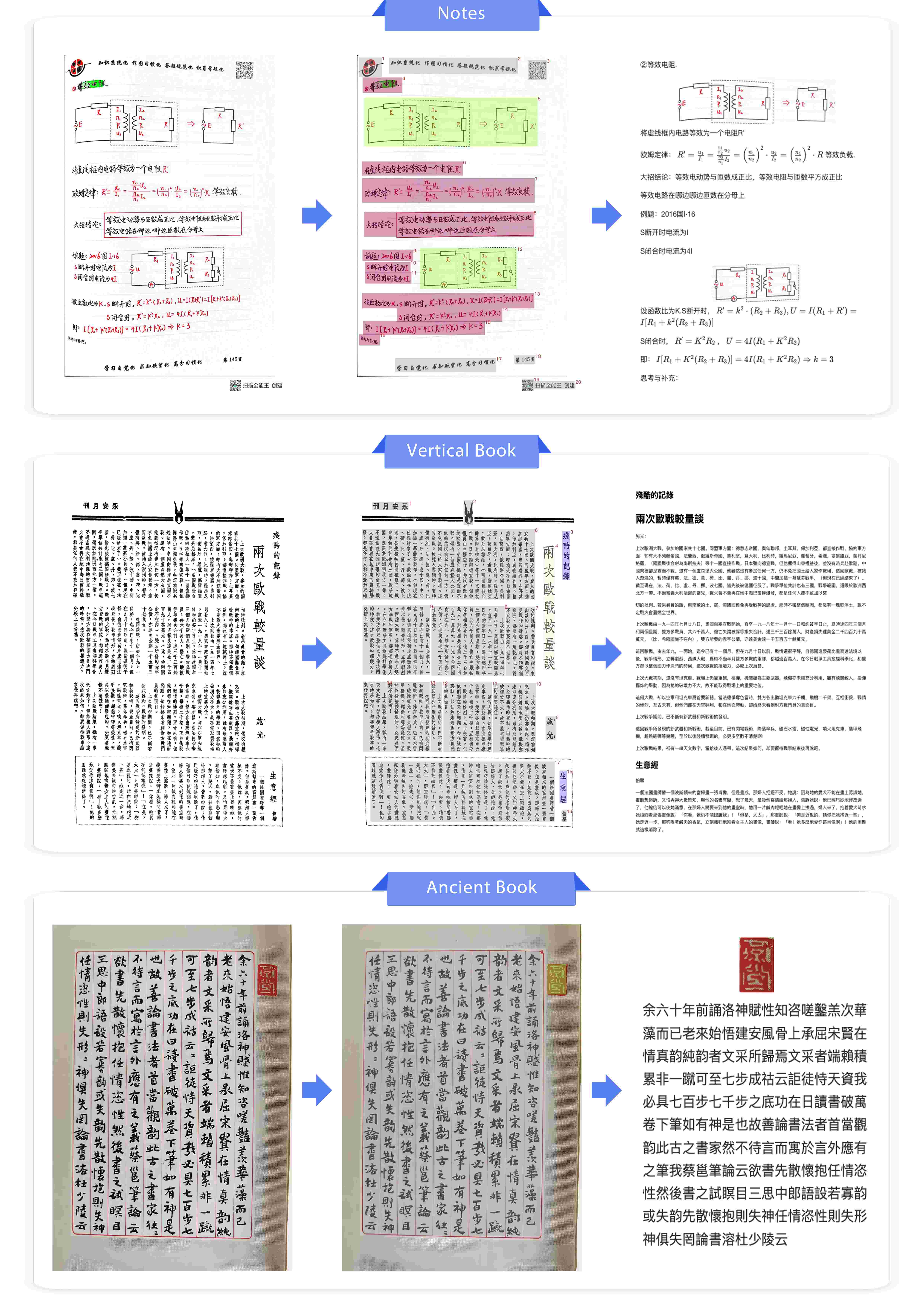} 

\caption{
    The Layout and Markdown Output for Notes, Vertical Book and Ancient Book.
}
\label{fig:overview3}
\end{figure}

\begin{figure}[t]
\centering
\includegraphics[width=0.88\linewidth]{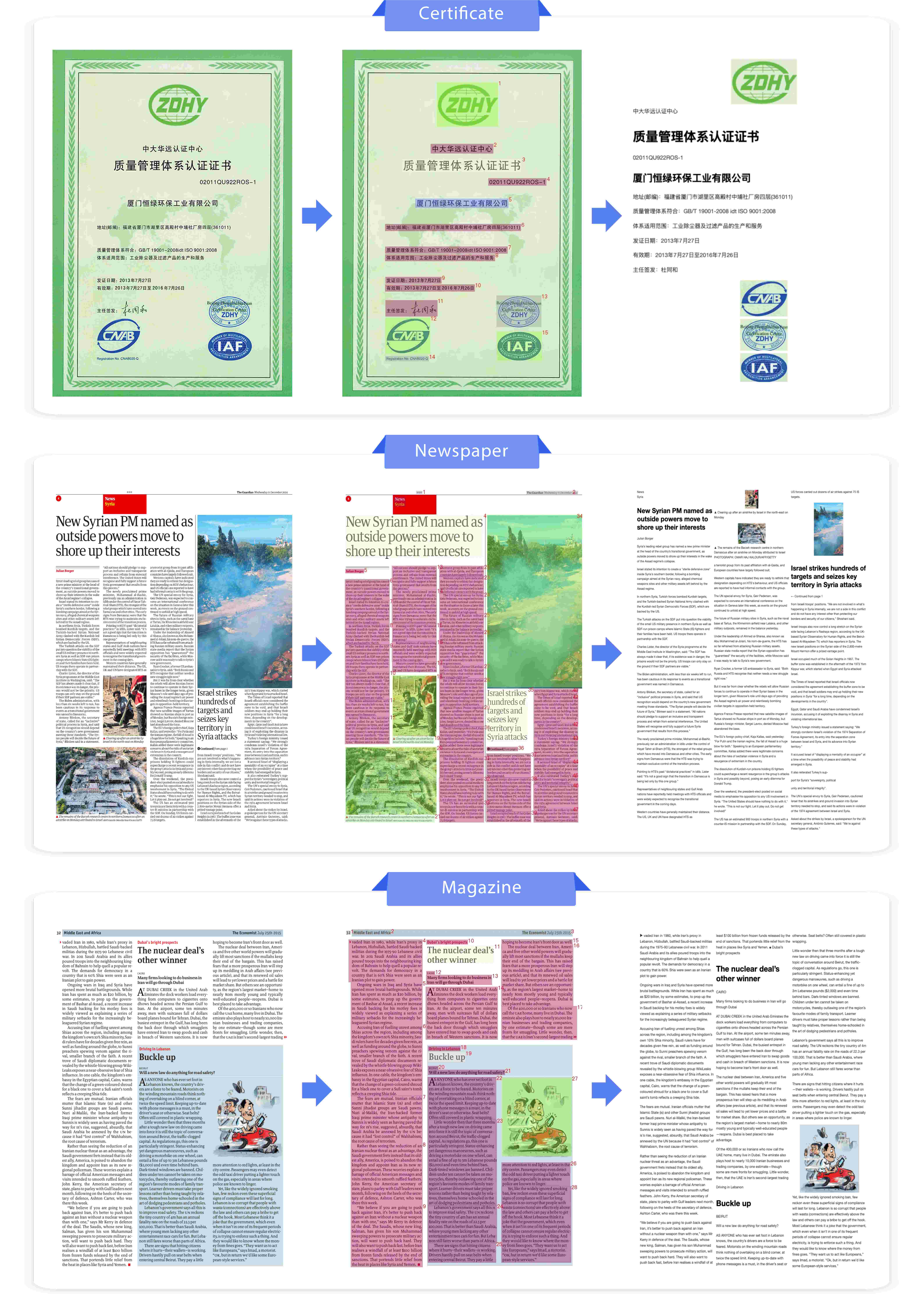} 

\caption{
    \centering
    The Layout and Markdown Output for Certificate, Newspaper and Magazine.
}
\label{fig:overview4}
\end{figure}

Section \ref{subsec:Comprehensive Document Parsing} demonstrates the overall document parsing capability of PaddleOCR-VL. Figures \ref{fig:overview1}-\ref{fig:overview4} are examples of parsing different types of documents in Markdown format.

\begin{figure}[H]
\centering
\includegraphics[width=0.80\linewidth]{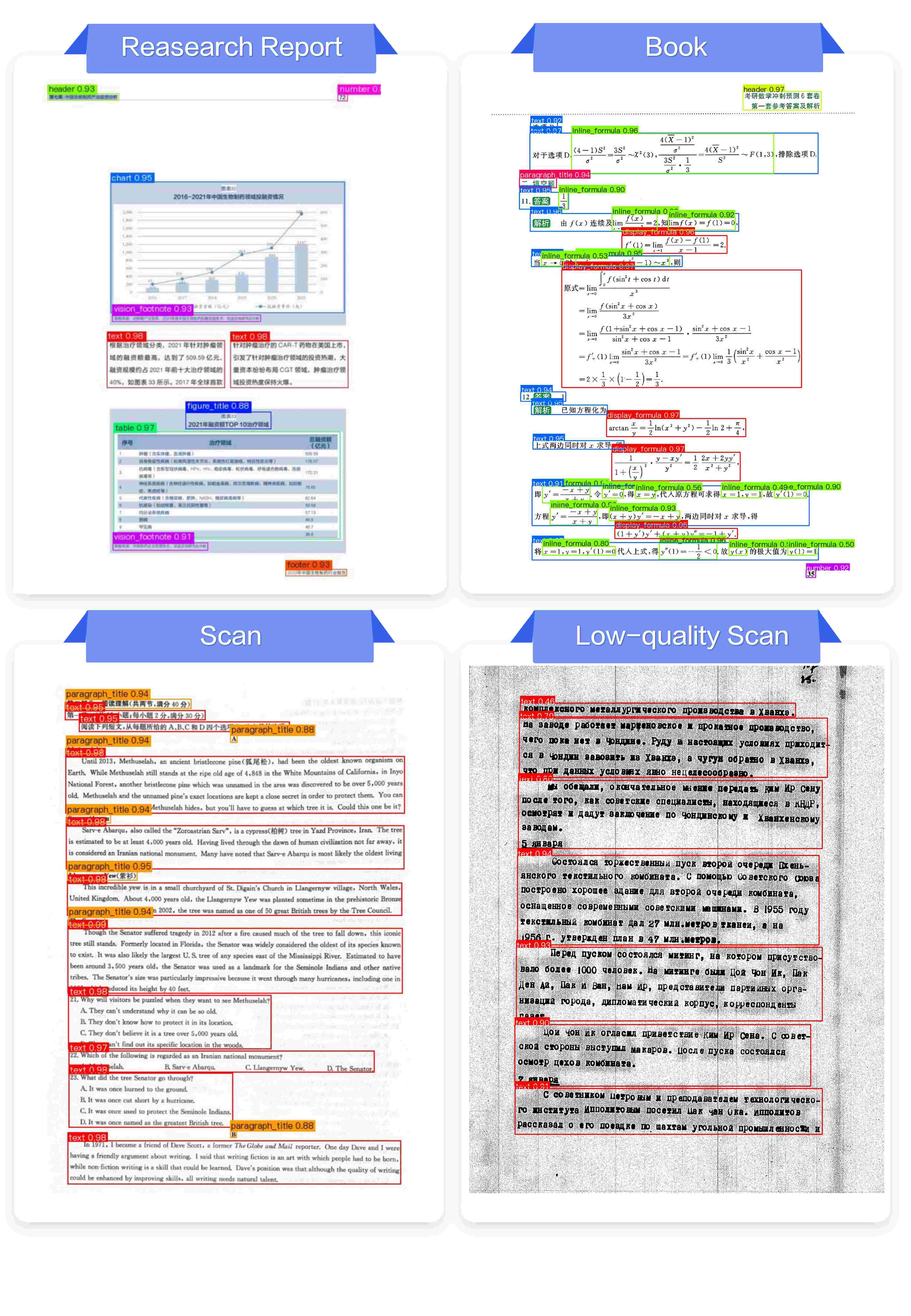} 

\caption{
    \centering
    The Layout Detection results for various types of documents.
}
\label{fig:layout01}
\end{figure}

\begin{figure}[h]
\centering
\includegraphics[width=0.86\linewidth]{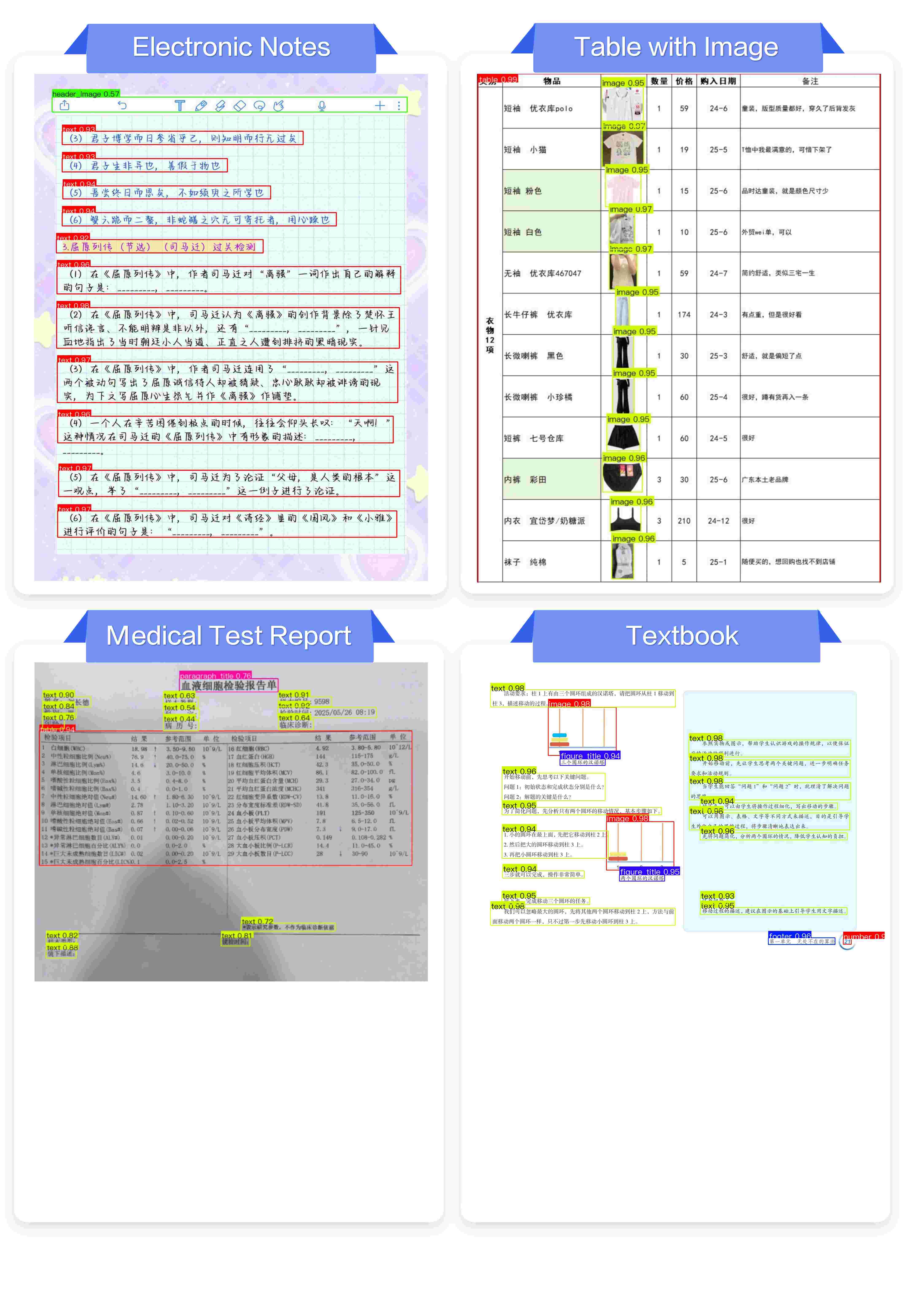} 

\caption{
    \centering
    The Layout Detection results for various types of documents.
}
\label{fig:layout02}
\end{figure}

\begin{figure}[h]
\centering
\includegraphics[width=0.86\linewidth]{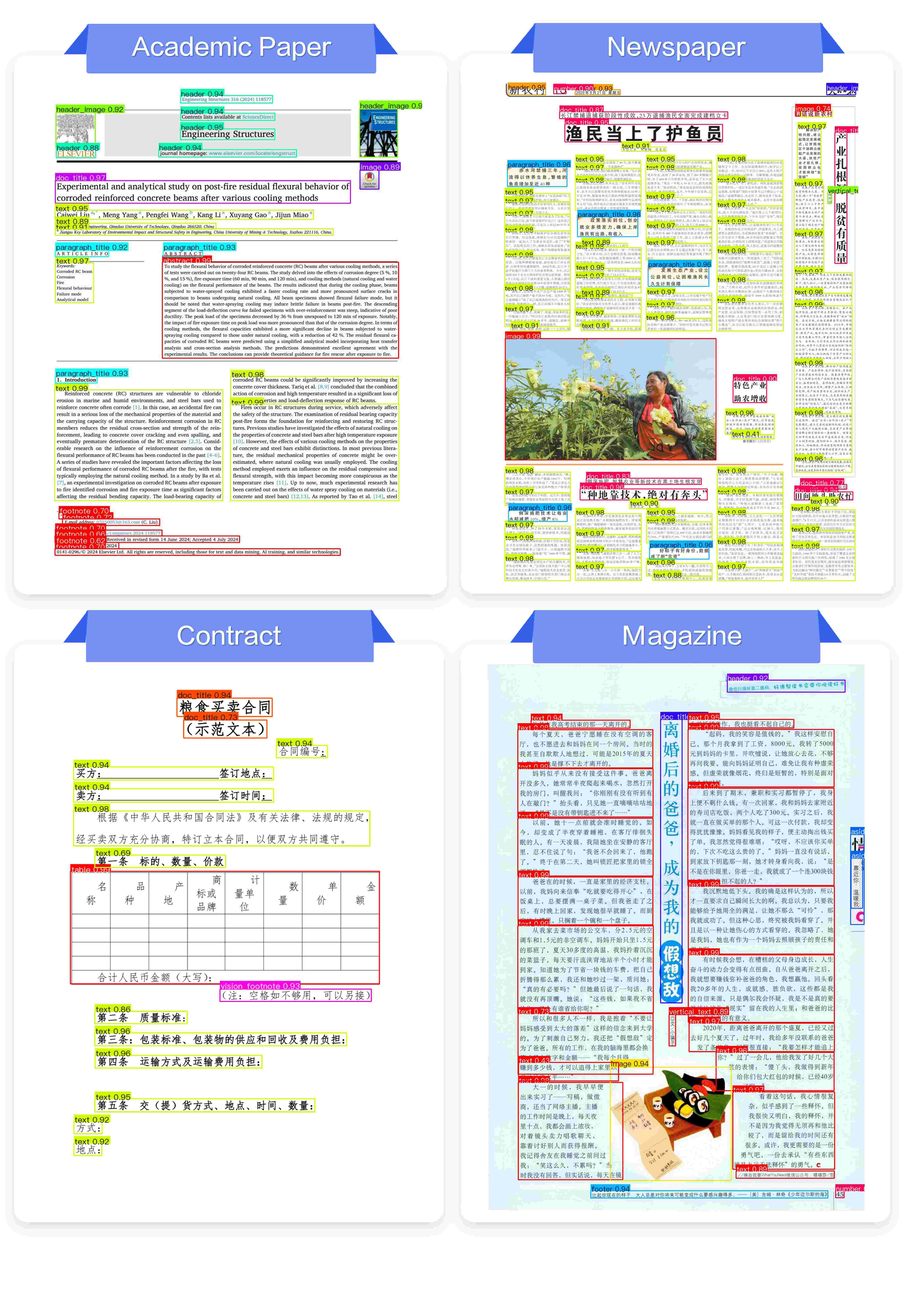} 

\caption{
    \centering
    The Layout Detection results for various types of documents.
}
\label{fig:layout03}
\end{figure}

Figures \ref{fig:layout01}-\ref{fig:layout03} in section \ref{subsec:Layout Detection} illustrate the superior ability of PaddleOCR-VL to process pages featuring intricate or challenging layouts.

\begin{figure}[h]
\centering
\includegraphics[width=0.80\linewidth]{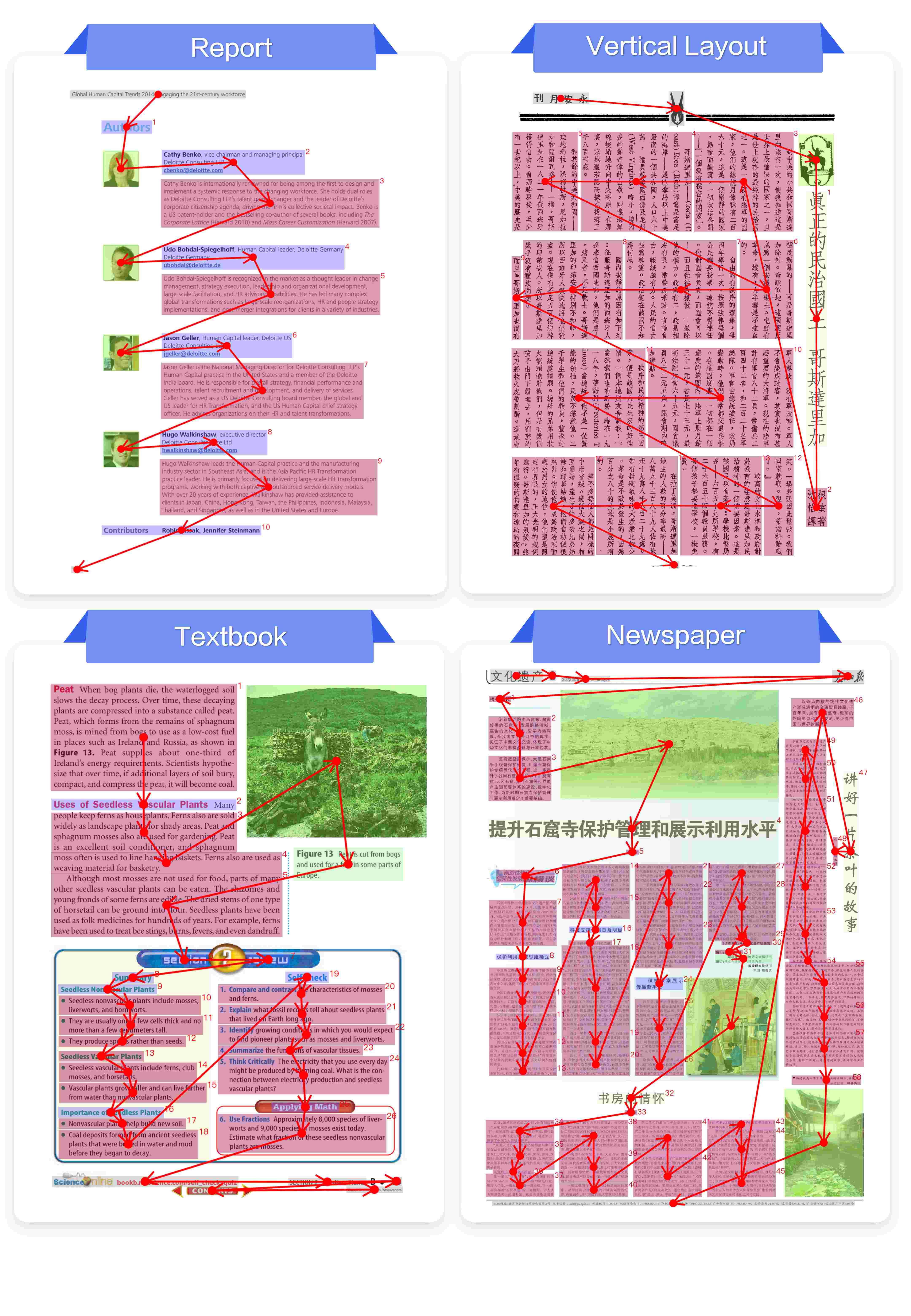} 

\caption{
    \centering
    The Reading Order results for various types of documents.
}
\label{fig:order_01}
\end{figure}

\begin{figure}[h]
\centering
\includegraphics[width=0.86\linewidth]{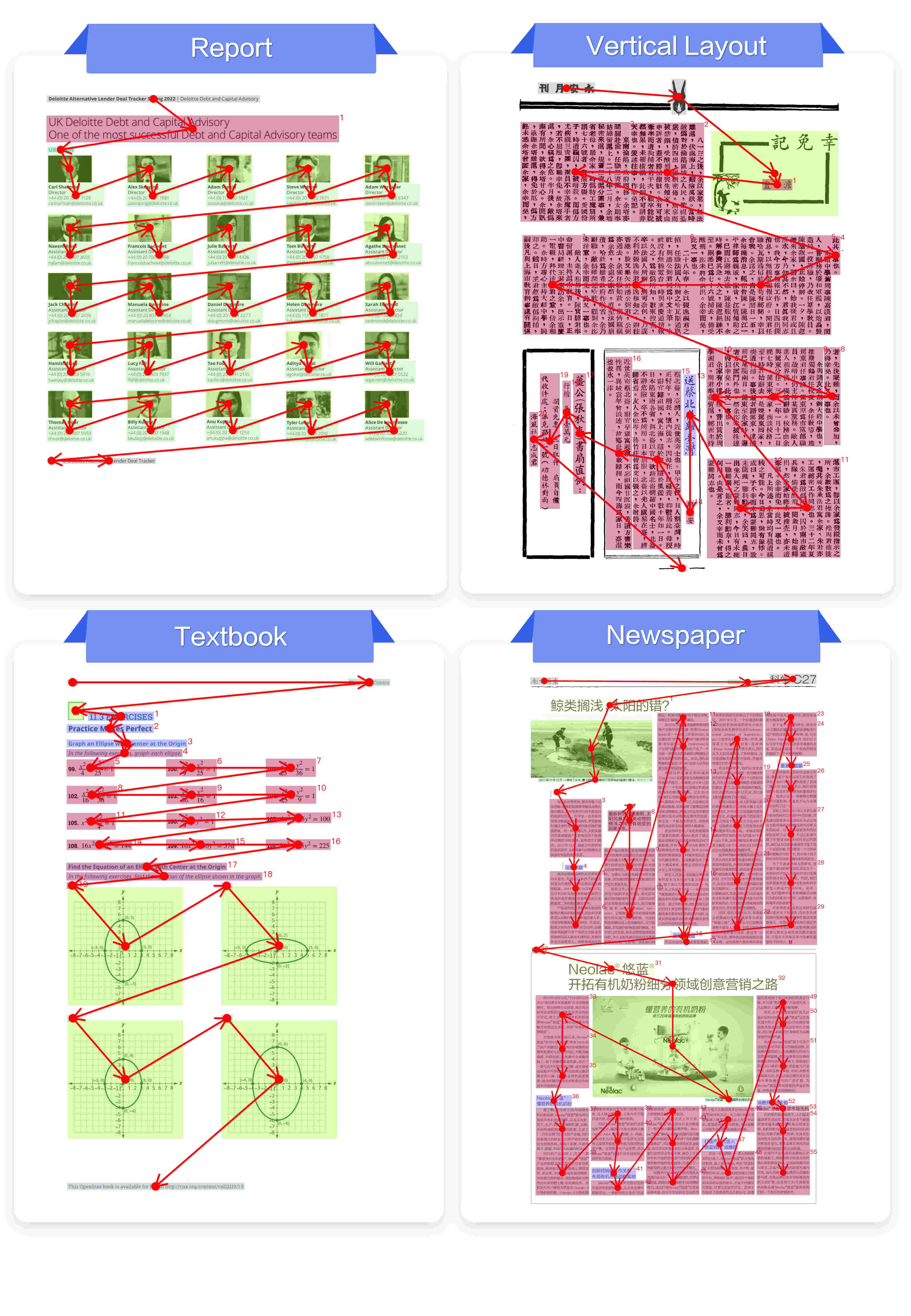} 

\caption{
    \centering
    The Reading Order results for various types of documents.
}
\label{fig:order_02}
\end{figure}

Figures \ref{fig:order_01} and \ref{fig:order_02} in section \ref{subsec:Reading Order} demonstrate that PaddleOCR-VL maintains excellent reading order when faced with complex layouts, such as those found in various reports, textbooks, newspapers, magazines, and even vertical documents.

\begin{figure}[h]
\centering
\includegraphics[width=0.80\linewidth]{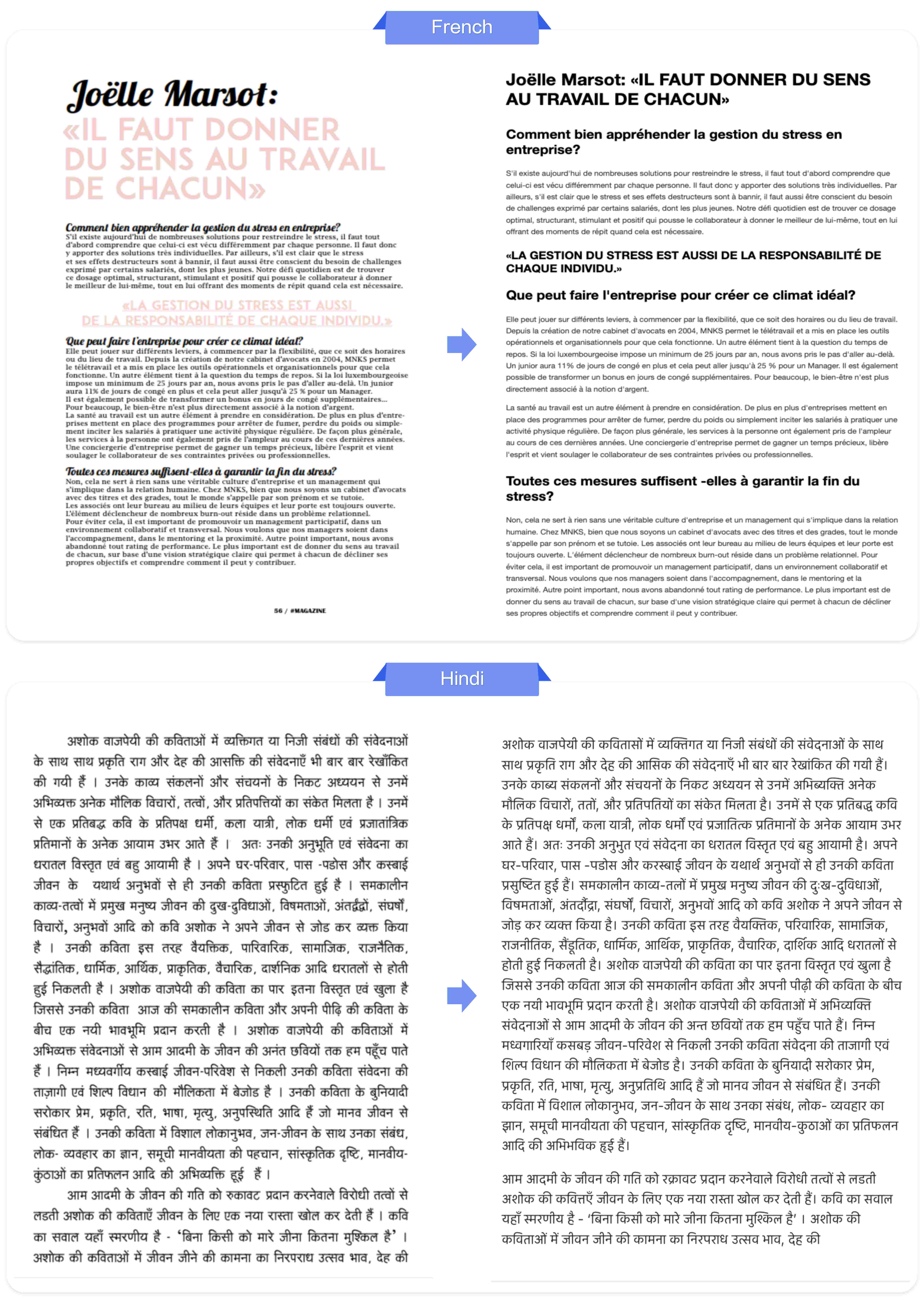} 

\caption{
    \centering
    The markdown output for French and Hindi documents.
}
\label{fig:text_french_hindi}
\end{figure}

\begin{figure}[h]
\centering
\includegraphics[width=0.88\linewidth]{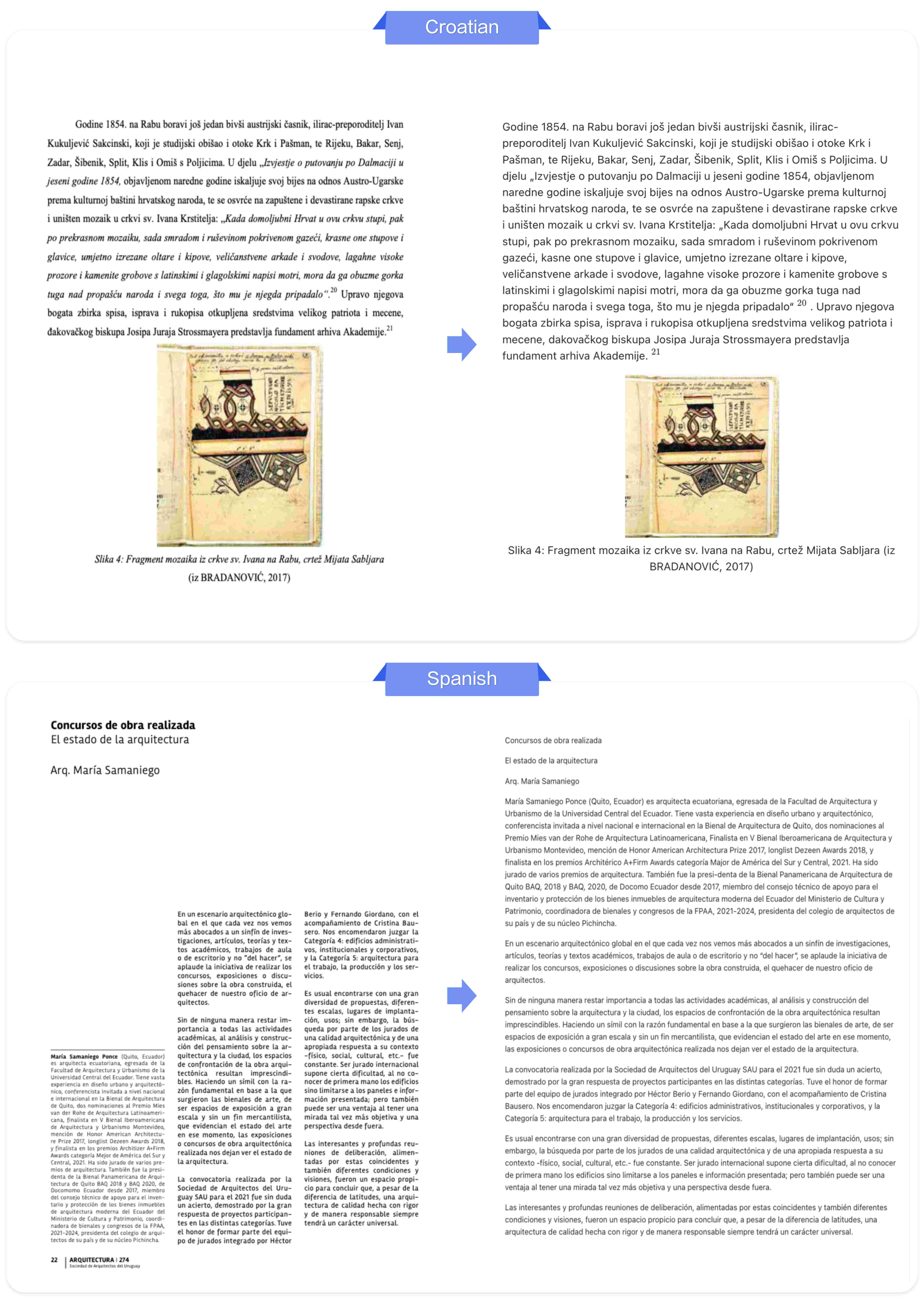} 

\caption{
    \centering
    The markdown output for Croatian and Spanish documents.
}
\label{fig:text_croatian_spanish}
\end{figure}

\begin{figure}[h]
\centering
\includegraphics[width=0.88\linewidth]{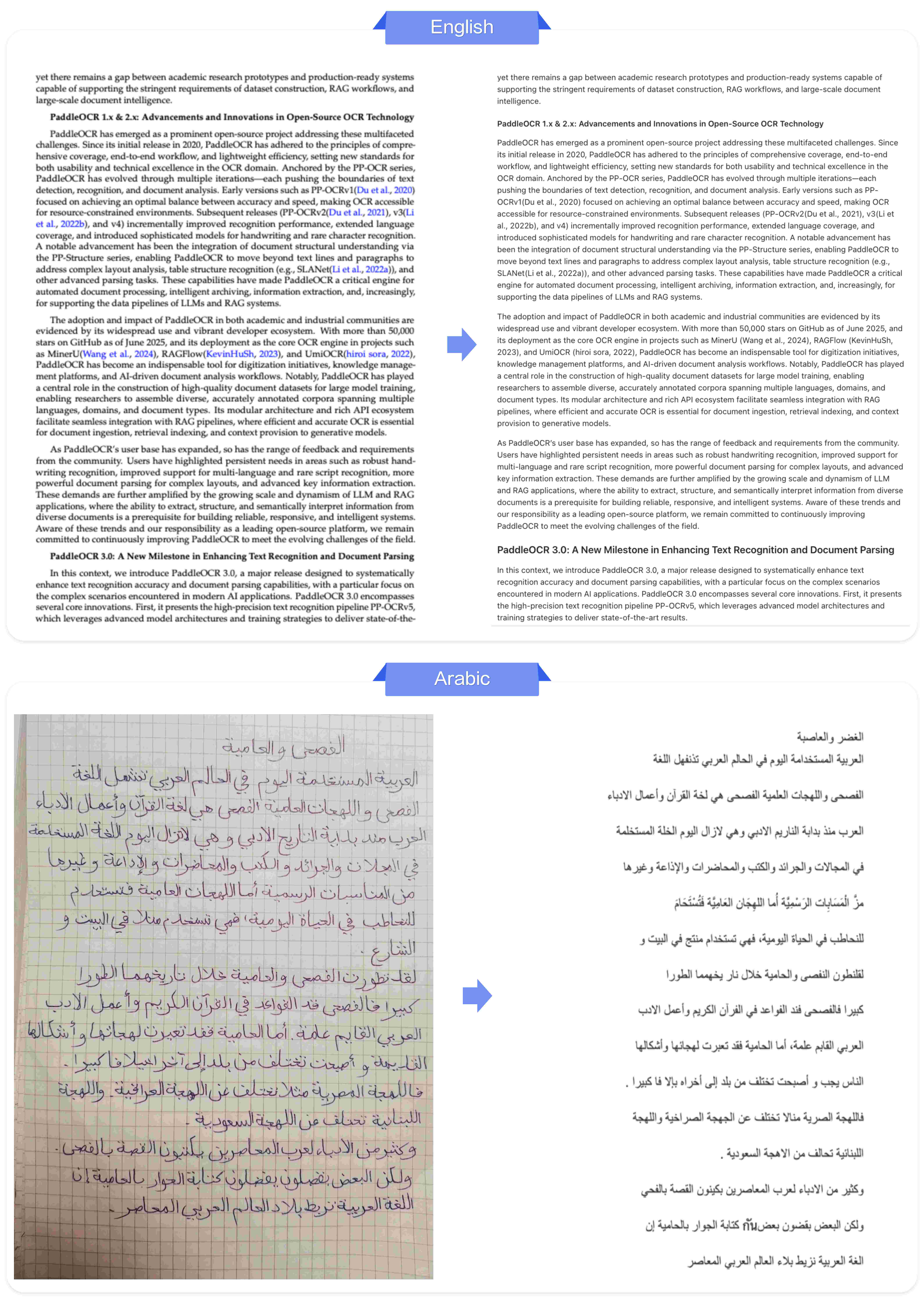} 

\caption{
    \centering
    The markdown output for English and Arabic documents.
}
\label{fig:ext_english_arabic}
\end{figure}

\begin{figure}[h]
\centering
\includegraphics[width=0.88\linewidth]{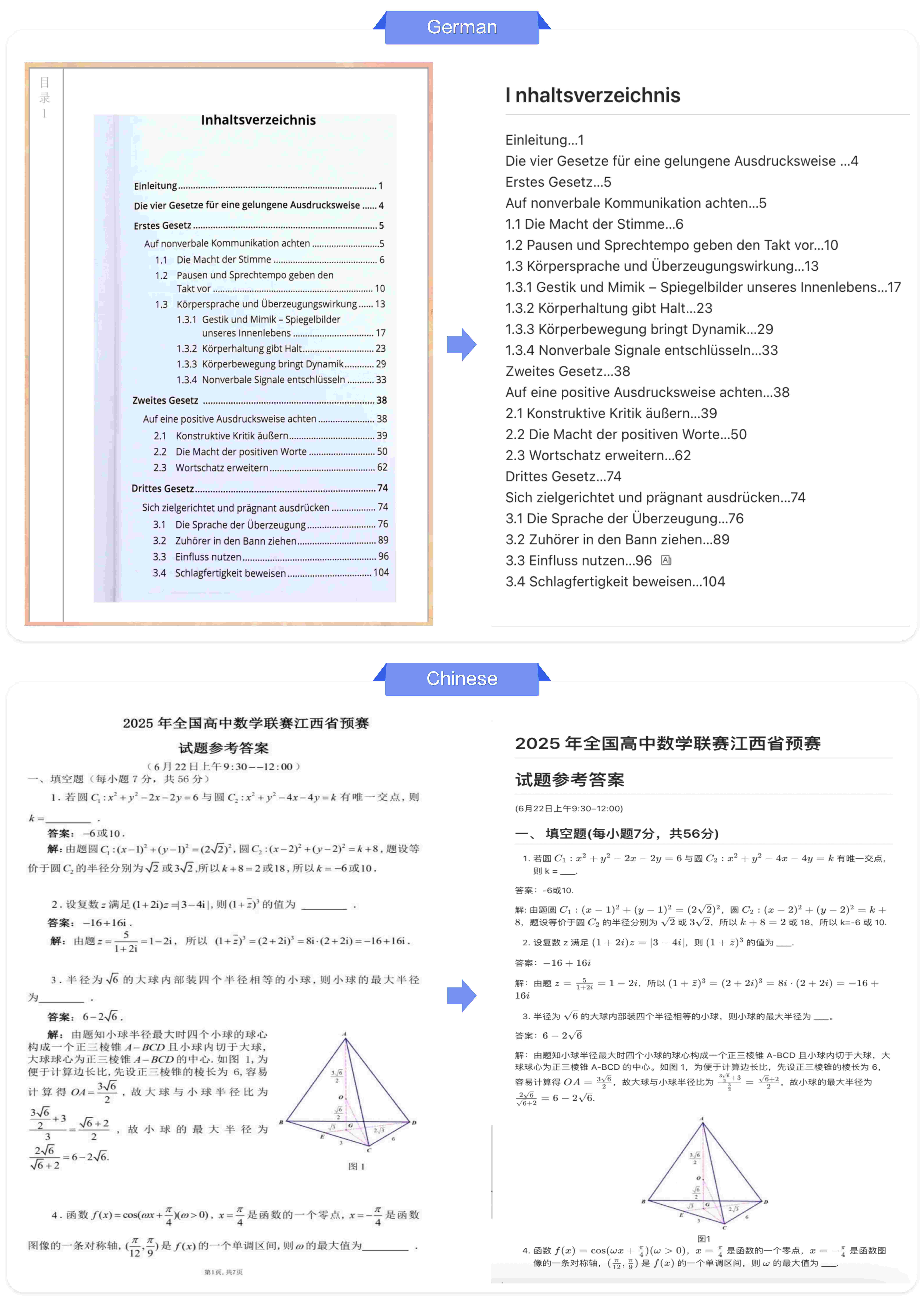} 

\caption{
    \centering
    The markdown output for German and Chinese documents.
}
\label{fig:text_german_chinese}
\end{figure}

\begin{figure}[h]
\centering
\includegraphics[width=0.88\linewidth]{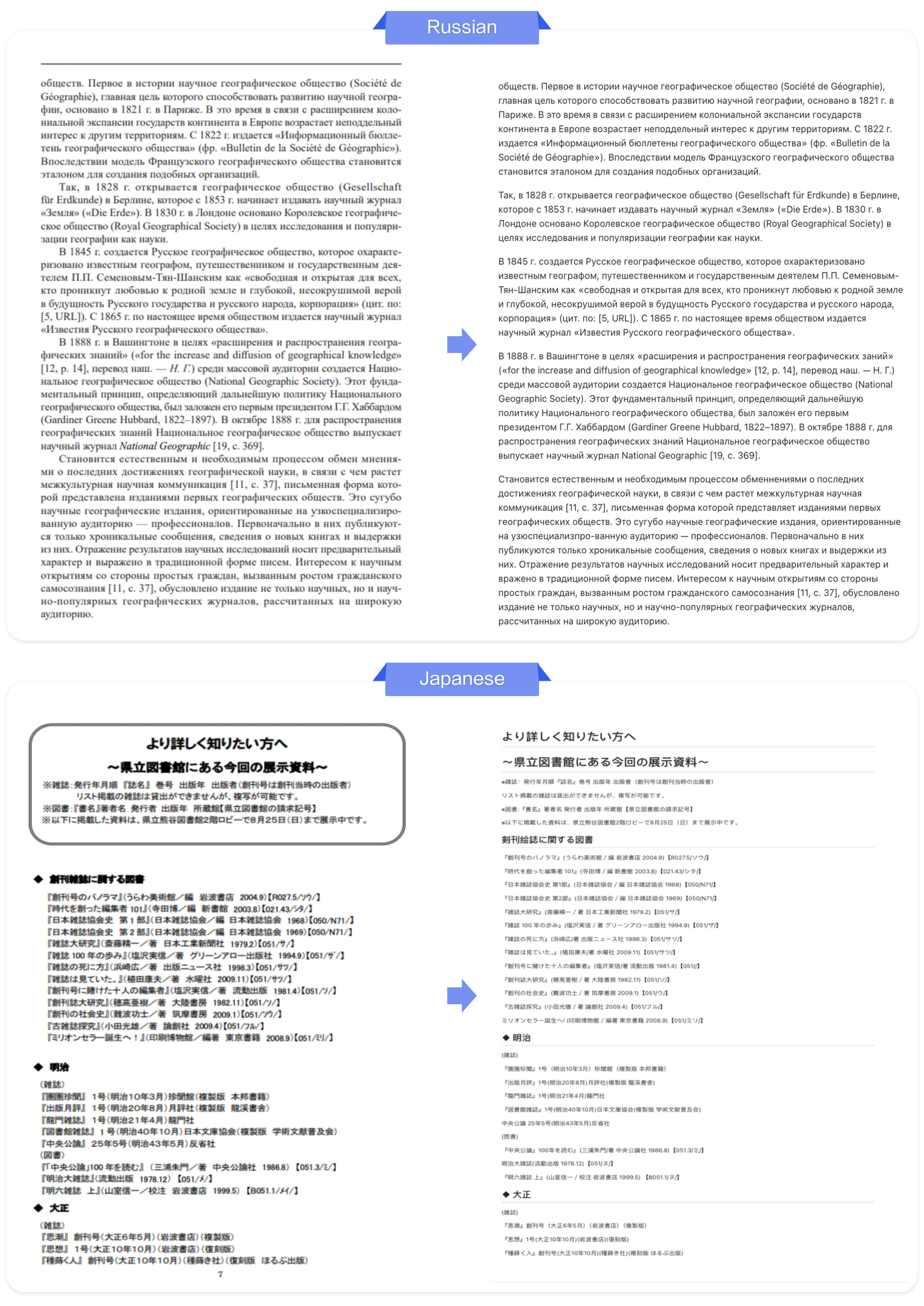} 

\caption{
    \centering
    The markdown output for Russian and Japanese documents.
}
\label{fig:text_russian_japanese}
\end{figure}

\begin{figure}[h]
\centering
\includegraphics[width=0.88\linewidth]{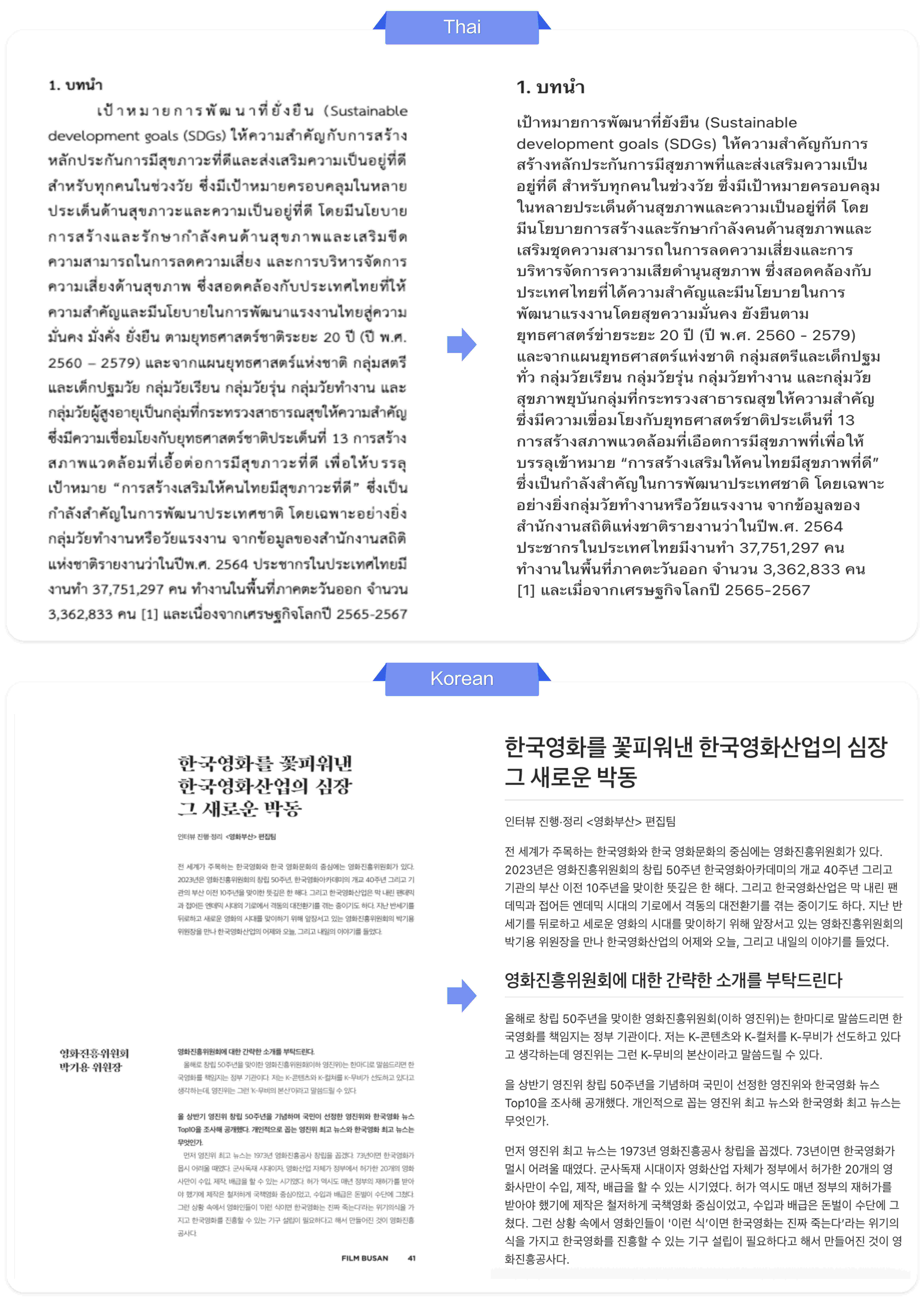} 

\caption{
    \centering
    The markdown output for Thai and Korean documents.
}
\label{fig:text_thai_korean}
\end{figure}

\begin{figure}[h]
\centering
\includegraphics[width=0.81\linewidth]{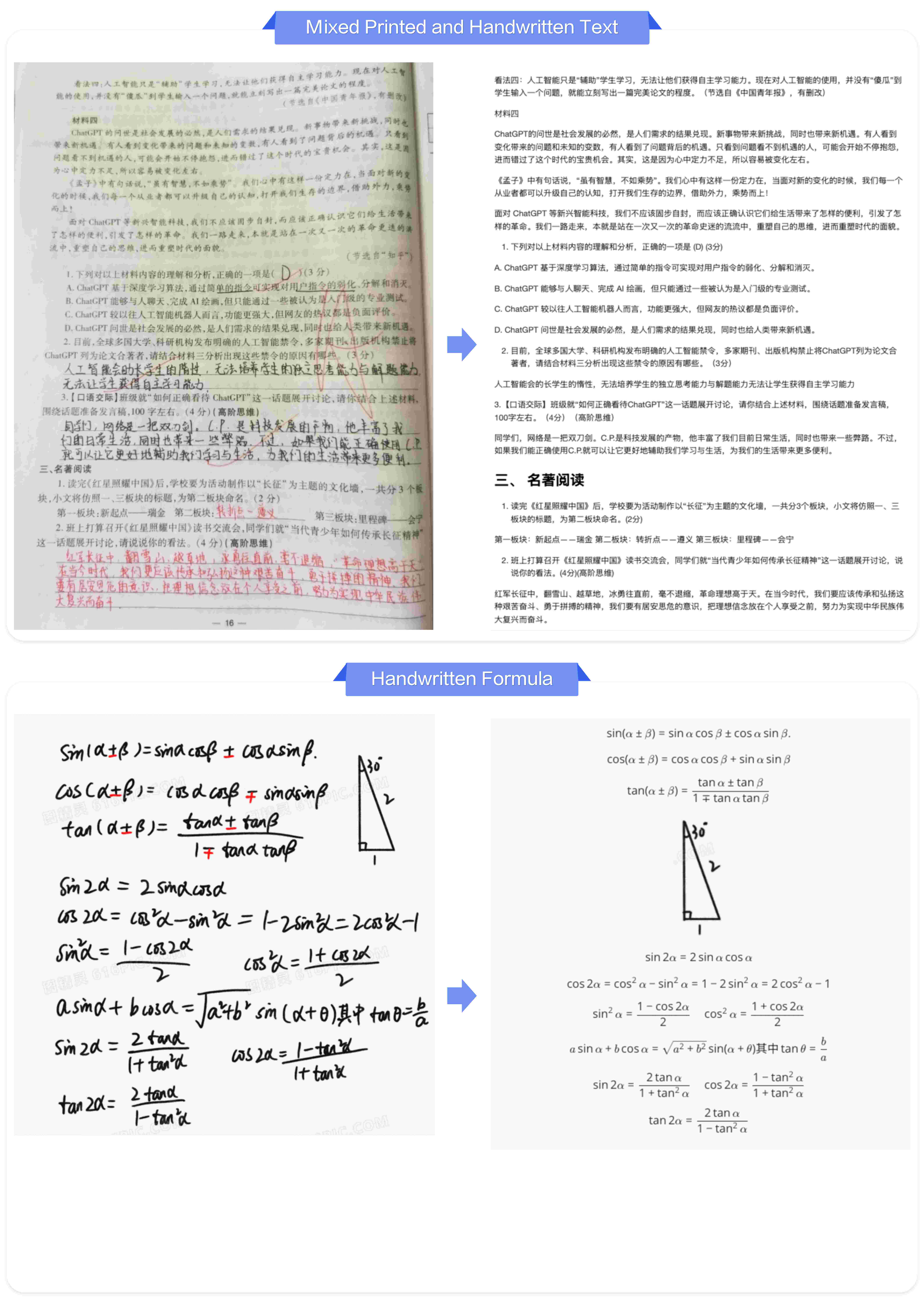} 

\caption{
    \centering
    The markdown output for Mixed Printed Handwritten Text and Handwritten Formula documents.
}
\label{fig:text_handwriting01}
\end{figure}

\begin{figure}[h]
\centering
\includegraphics[width=0.80\linewidth]{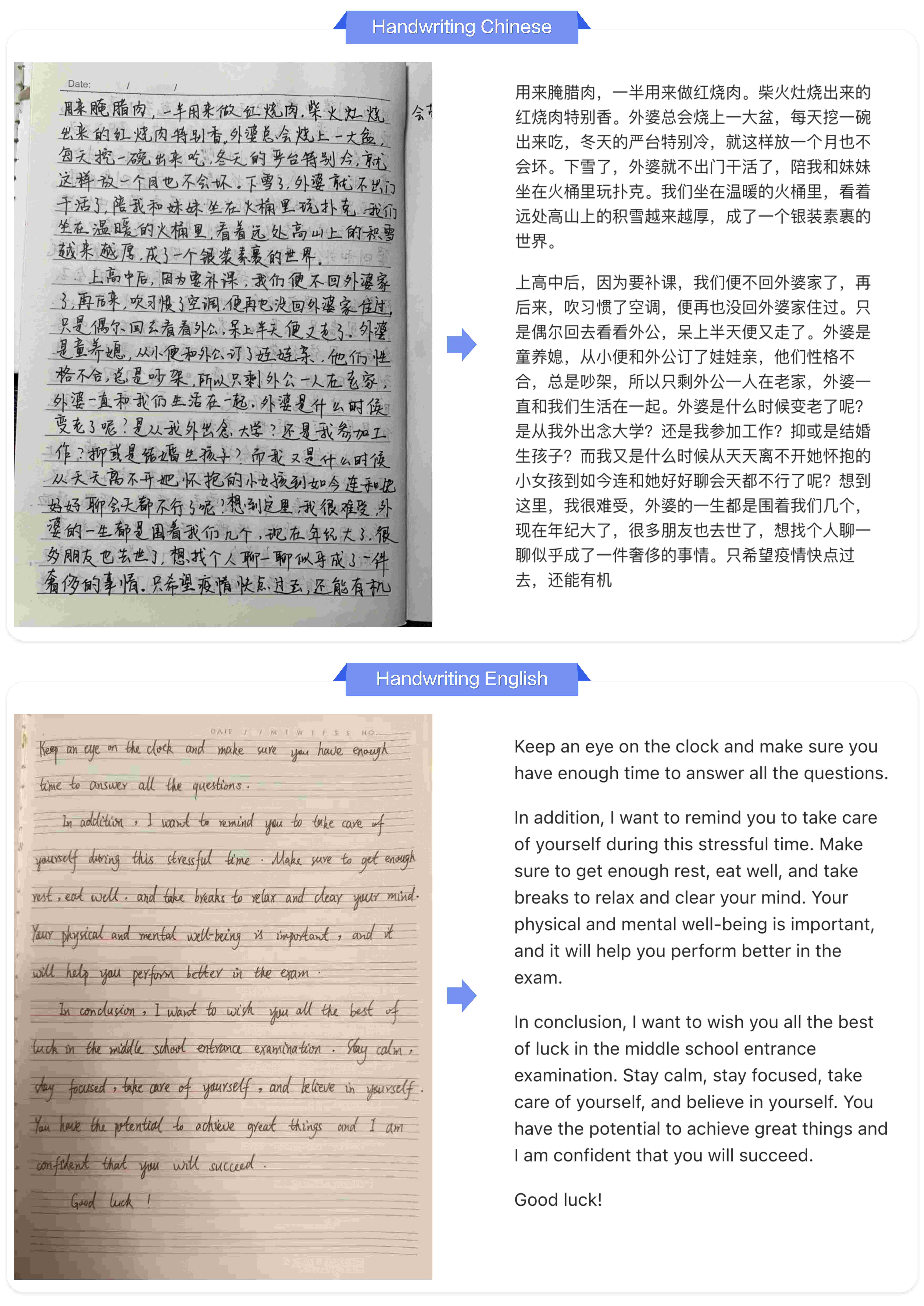} 

\caption{
    \centering
    The markdown output for Handwriting Chinese and Handwriting English documents.
}
\label{fig:text_handwriting02}
\end{figure}

\begin{figure}[h]
\centering
\includegraphics[width=0.83\linewidth]{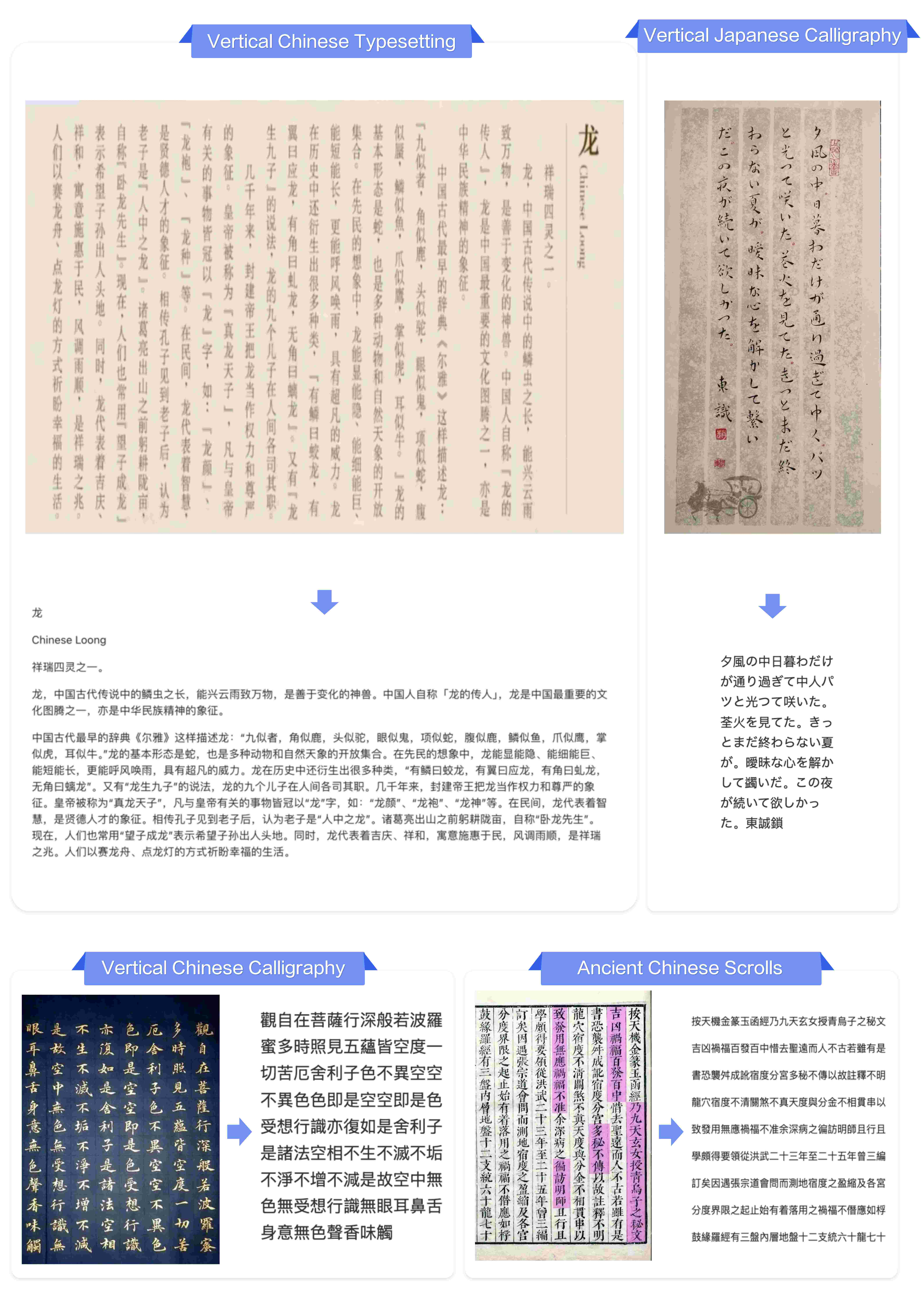} 

\caption{
    \centering
    The markdown output for various types of vertical documents.
}
\label{fig:text_vertical}
\end{figure}

Section \ref{subsec:Text Recognition} highlights the robust text recognition performance of PaddleOCR-VL in challenging cases, including multilingual text, handwriting text, and vertical text, which are presented in Figures \ref{fig:text_french_hindi}-\ref{fig:text_vertical}.

\begin{figure*}[h]
\centering
\includegraphics[width=0.88\linewidth]{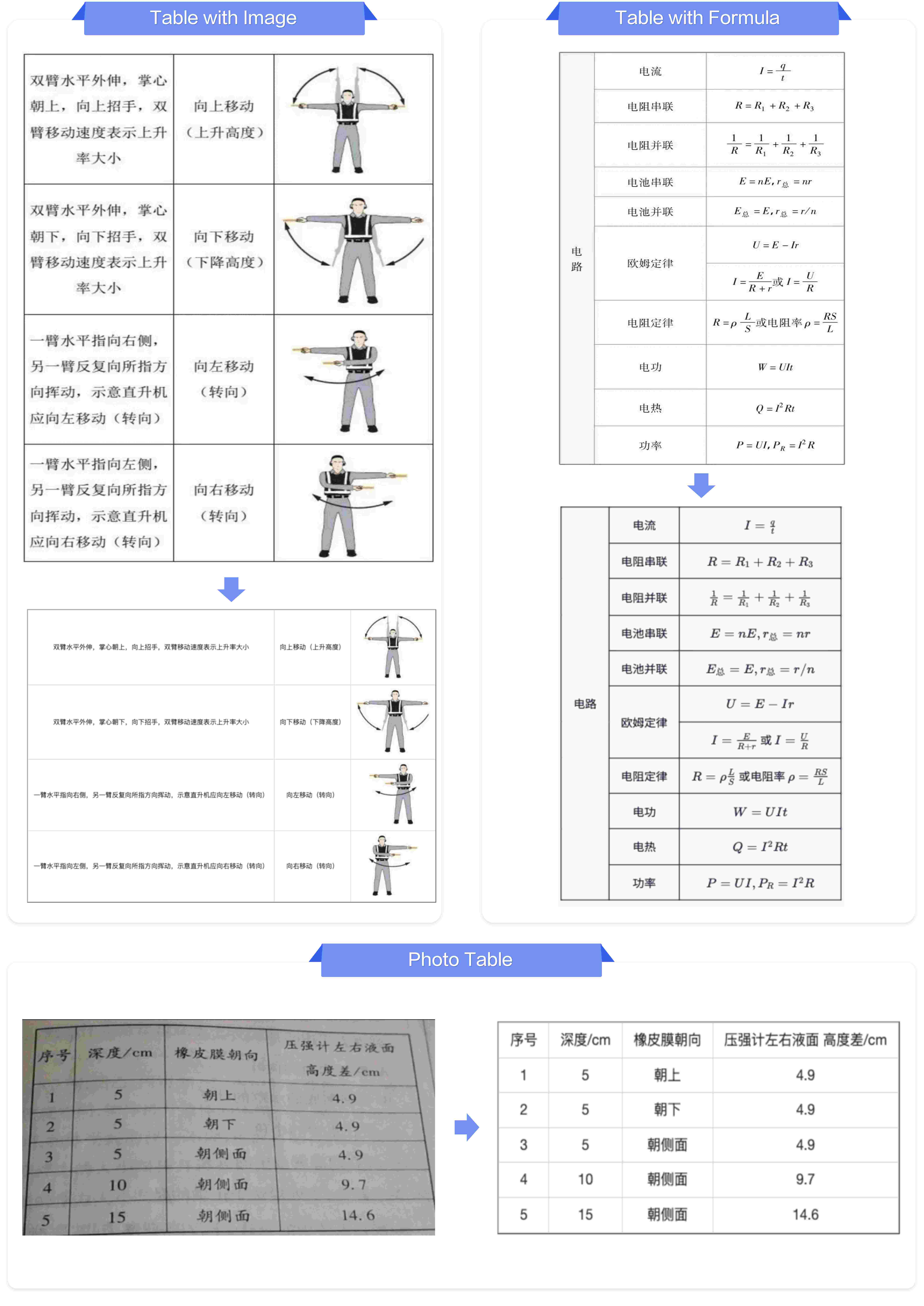} 

\caption{
    \centering
    The markdown output for various types of Tables.
}
\label{fig:table_02}
\end{figure*}

The model's table recognition abilities are demonstrated in section \ref{subsec:Table Recognition}. Figures \ref{fig:table_01} and \ref{fig:table_02} showcase its robust handling of a wide array of table formats, including tables from academic papers, tables from financial reports, tables with watermark, tables with image, tables with formulas and photograph of tables.

\begin{figure}[h]
\centering
\includegraphics[width=0.83\linewidth]{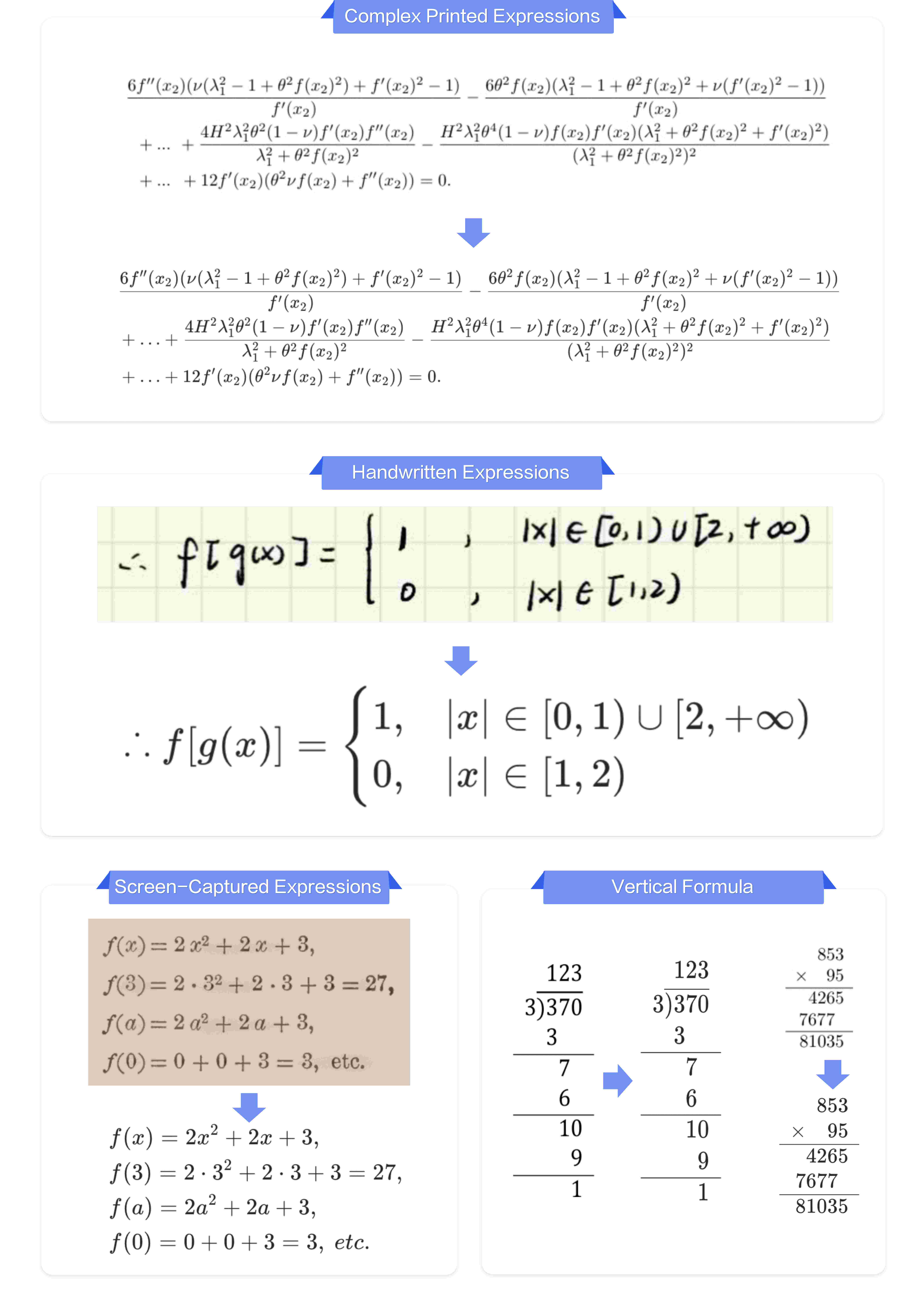} 

\caption{
    \centering
    The markdown output for various types of Formulas.
}
\label{fig:formula_EN}
\end{figure}

\begin{figure*}[h]
\centering
\includegraphics[width=0.88\linewidth]{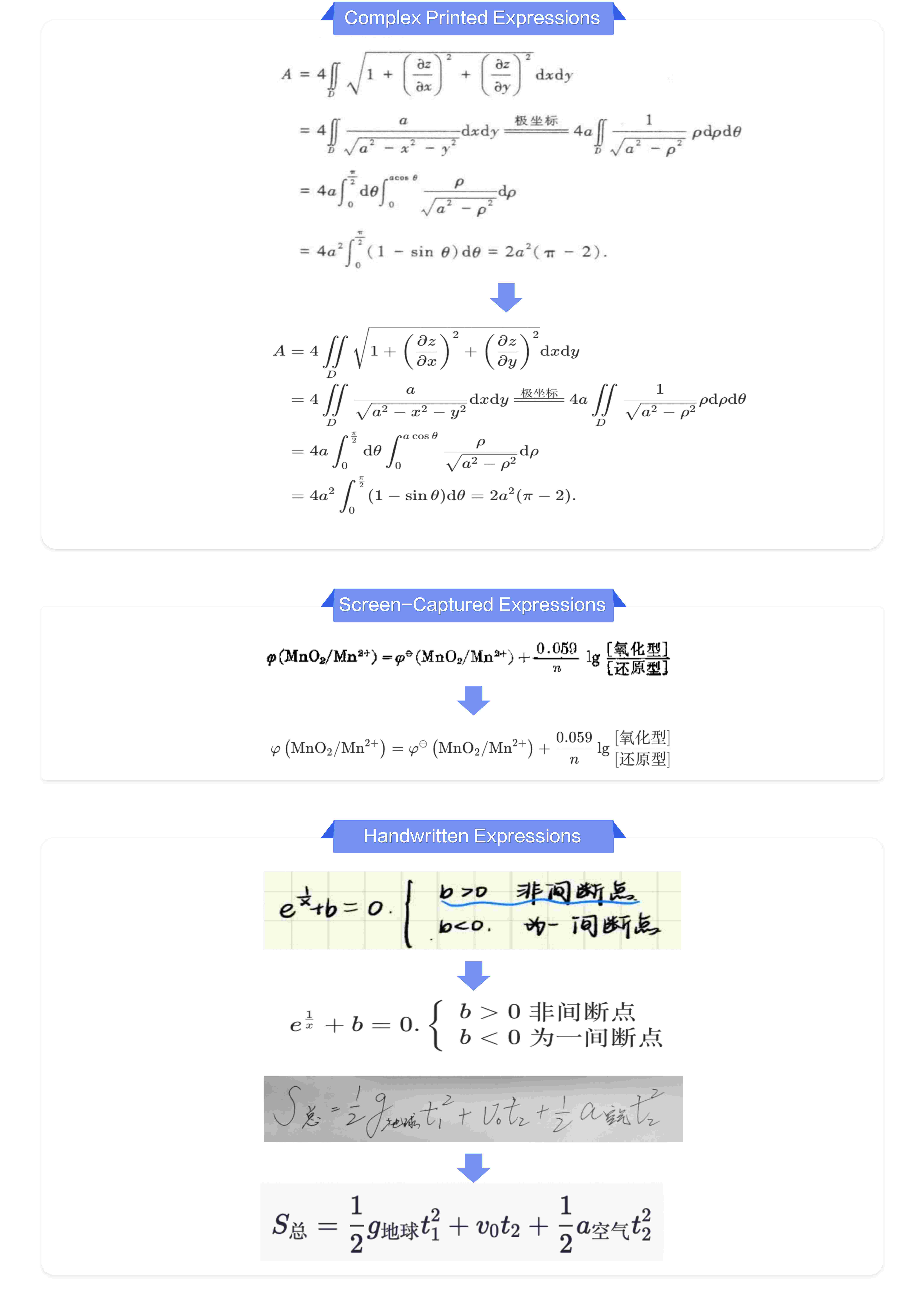} 

\caption{
    \centering
    The markdown output for various types of Formulas.
}
\label{fig:formula_ZH}
\end{figure*}

Figures in section \ref{subsec:Formula Recognition} detail the formula recognition performance. Figure \ref{fig:formula_EN} demonstrates the ability to handle various types of english formulas including complex printed expressions, handwritten expressions screen-captured expressions and vertical formula, while Figure \ref{fig:formula_ZH} focuses on the ability to handle formulas that contain Chinese characters.

\begin{figure}[h]
\centering
\includegraphics[width=0.80\linewidth]{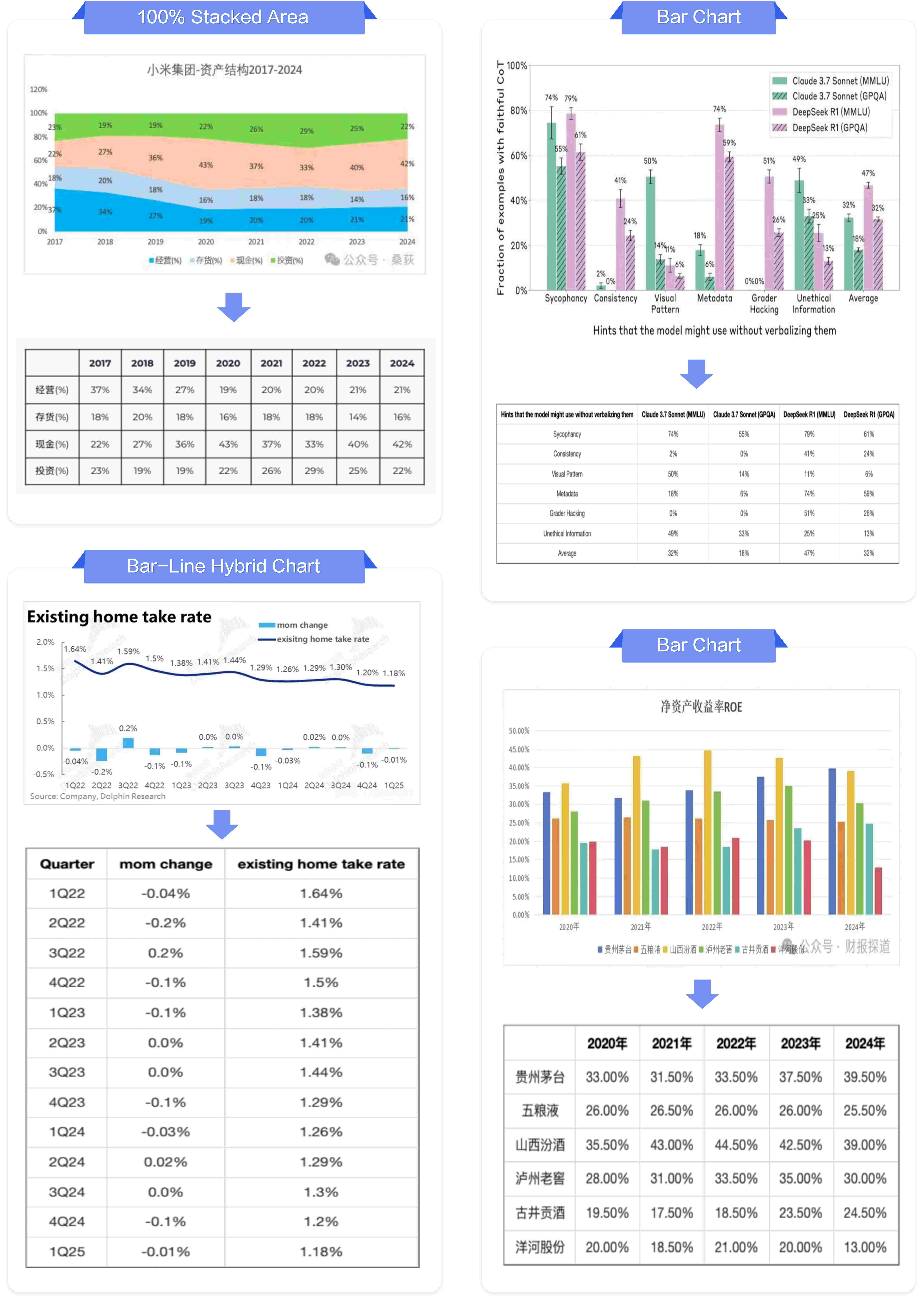} 

\caption{
    \centering
    The markdown output for various types of Charts.
}
\label{fig:chart_01}
\end{figure}

\begin{figure*}[t]
\centering
\includegraphics[width=0.88\linewidth]{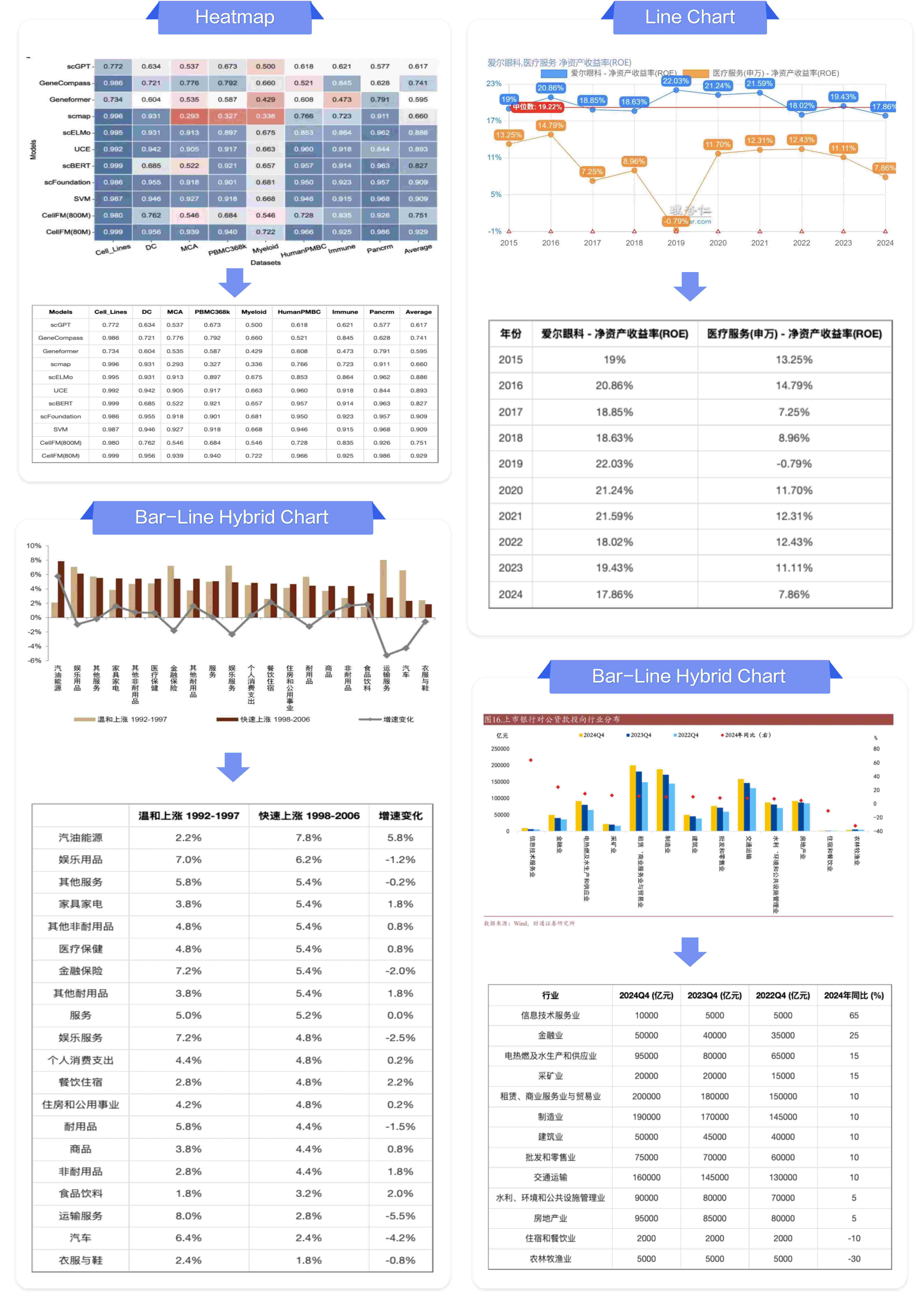} 

\caption{
    \centering
    The markdown output for various types of Charts.
}
\label{fig:chart_02}
\end{figure*}

\begin{figure}[h]
\centering
\includegraphics[width=0.88\linewidth]{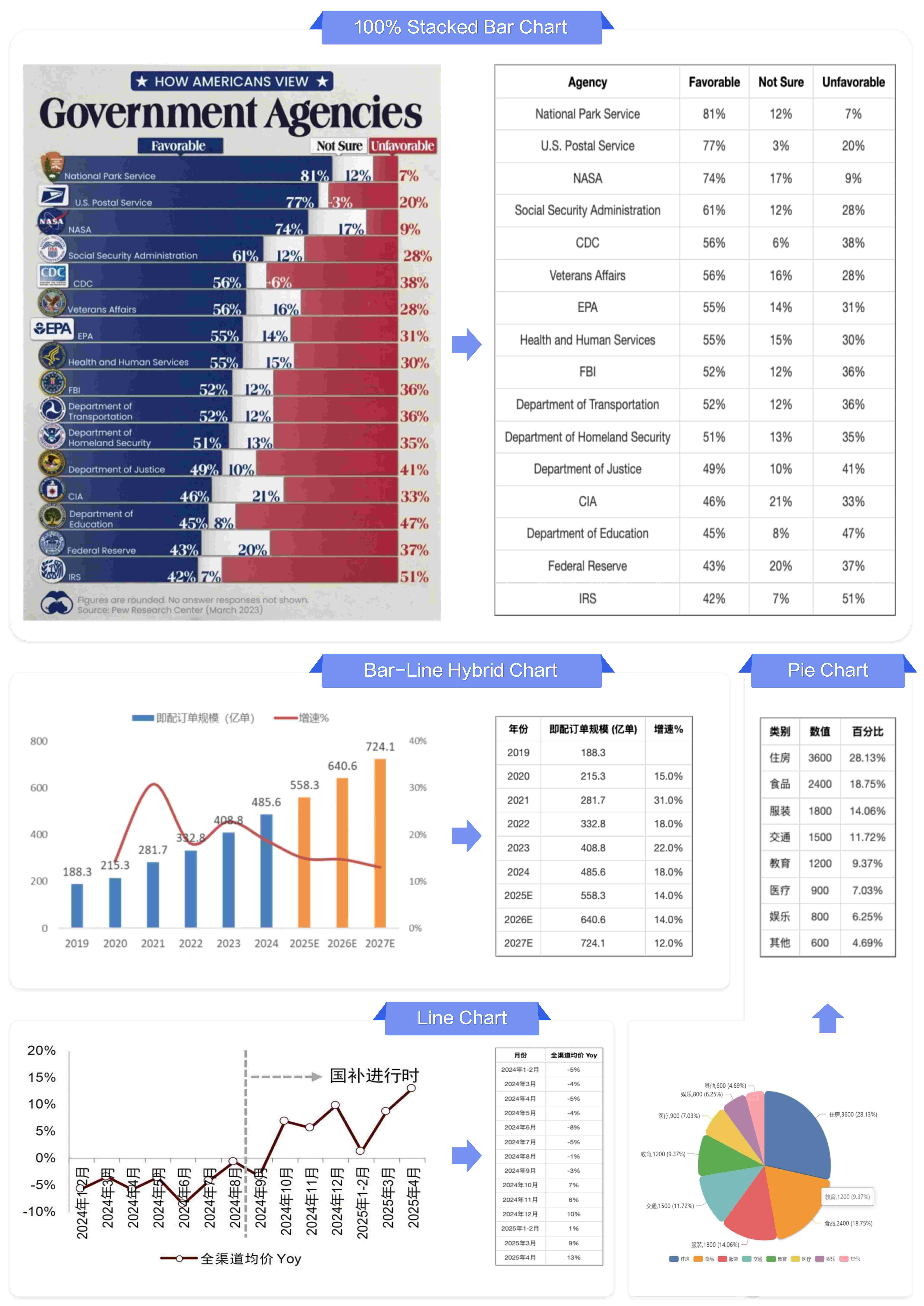} 

\caption{
    \centering
    The markdown output for various types of Charts.
}
\label{fig:chart_03}
\end{figure}

In section \ref{subsec:Chart Recognition}, PaddleOCR-VL demonstrates impressive chart recognition capabilities, a feature currently lacking in many expert OCR VLMs like MinerU2.5 \cite{MinerU2}, dots.ocr \cite{dotsocr} or MonkeyOCR \cite{li2025monkeyocr}. Figures \ref{fig:chart_01}-\ref{fig:chart_03} showcase our ability to parse various chart types, including pie charts, bar charts, line charts, bar-line hybrid charts and heatmap.

\subsection{Compare with Others}
\label{subsec:Compare with Others}

\begin{figure}[h]
\centering
\includegraphics[width=0.83\linewidth]{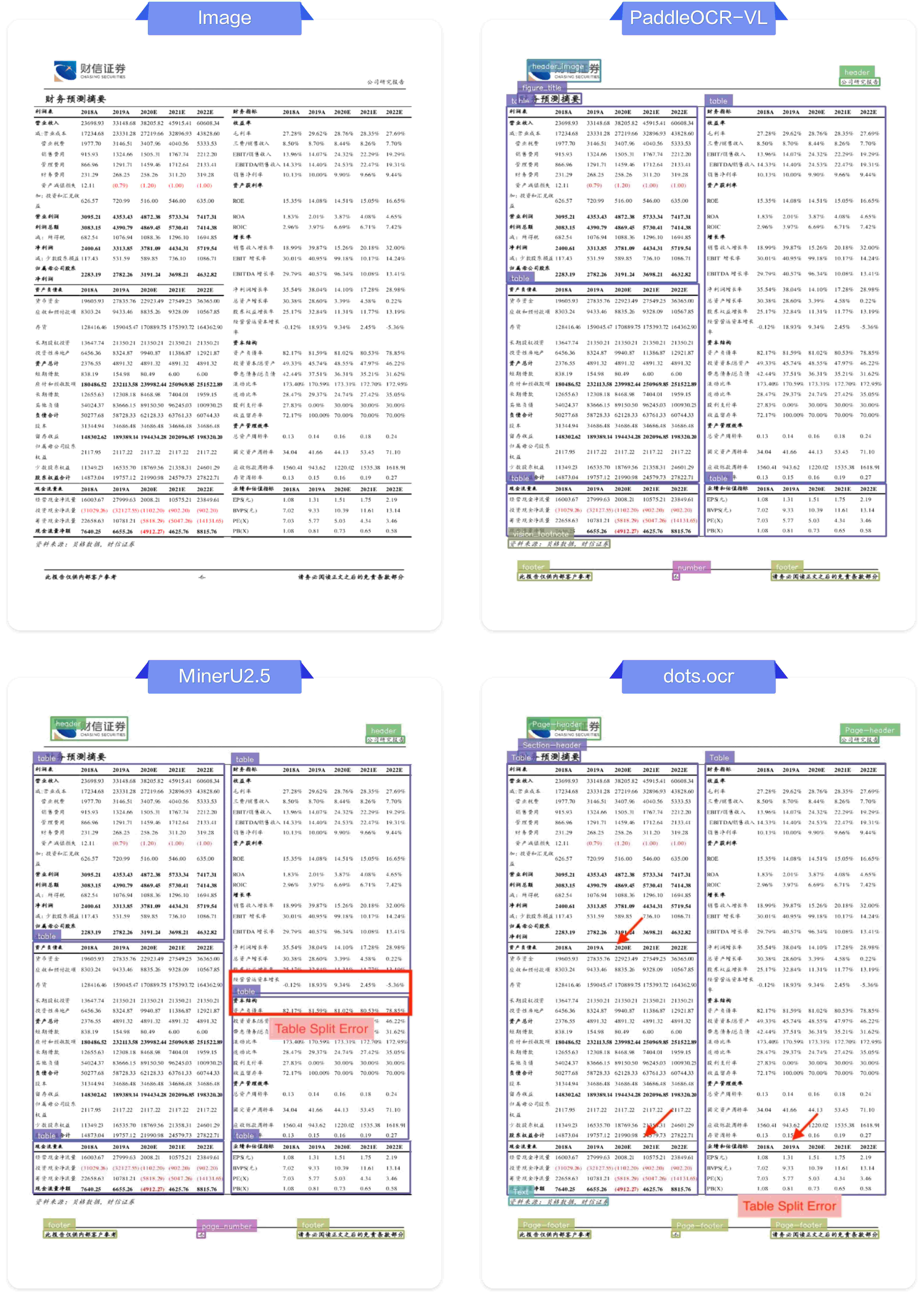} 

\caption{
    \centering
    Compare with others in Layout Detection.
}
\label{fig:cmp_layout_01}
\end{figure}

\begin{figure}[h]
\centering
\includegraphics[width=0.88\linewidth]{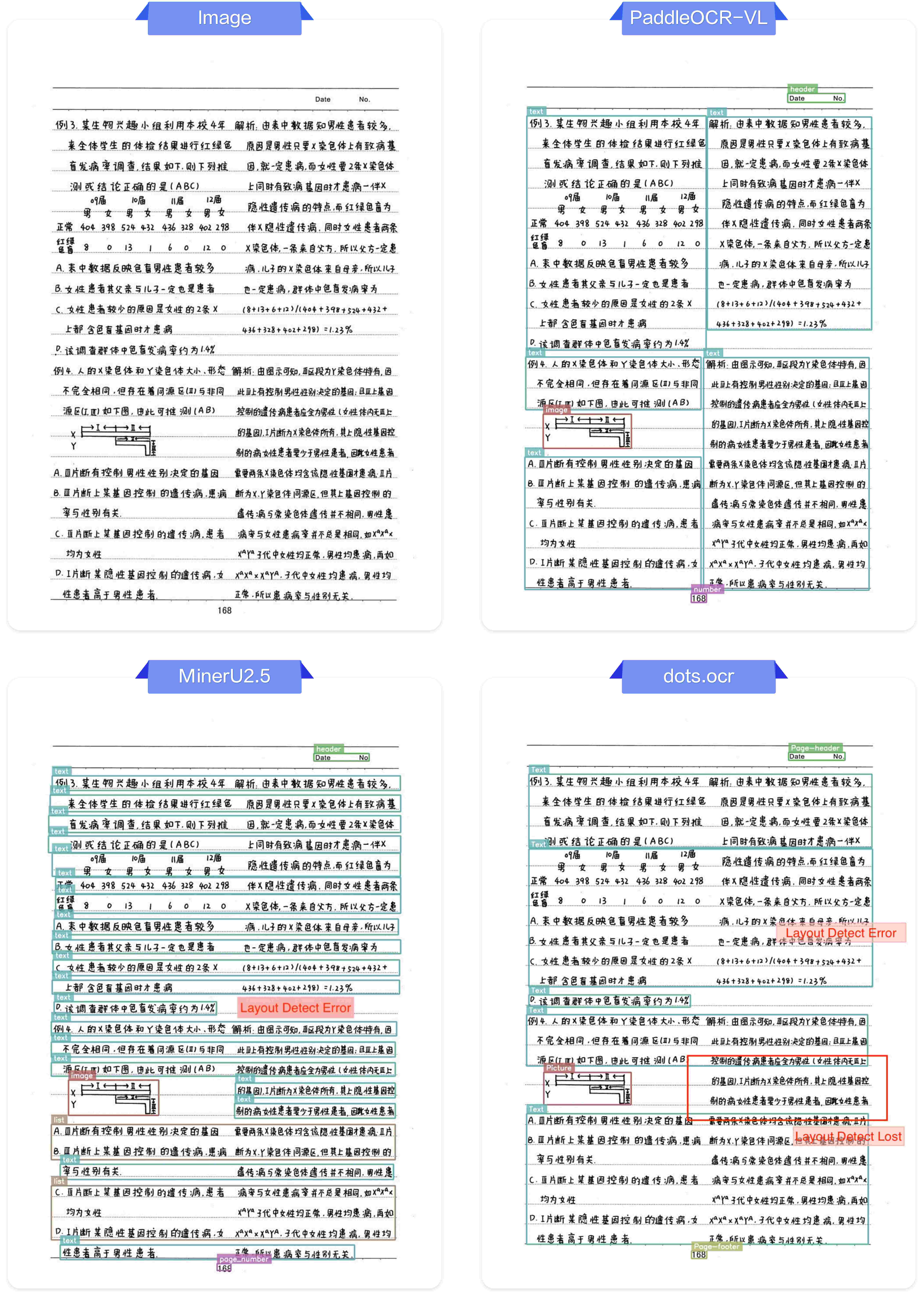} 

\caption{
    \centering
    Compare with others in Layout Detection.
}
\label{fig:cmp_layout_02}
\end{figure}

PaddleOCR-VL showcases superior performance in scenarios involving PDF pages with complex layout, consistently outperforming existing state-of-the-art (SOTA) models. This is evident from Figures \ref{fig:cmp_layout_01} and \ref{fig:cmp_layout_02}, which highlight its exceptional capability in handling pages with intricate layouts and unique elements, surpassing other solutions.

\begin{figure}[h]
\centering
\includegraphics[width=0.80\linewidth]{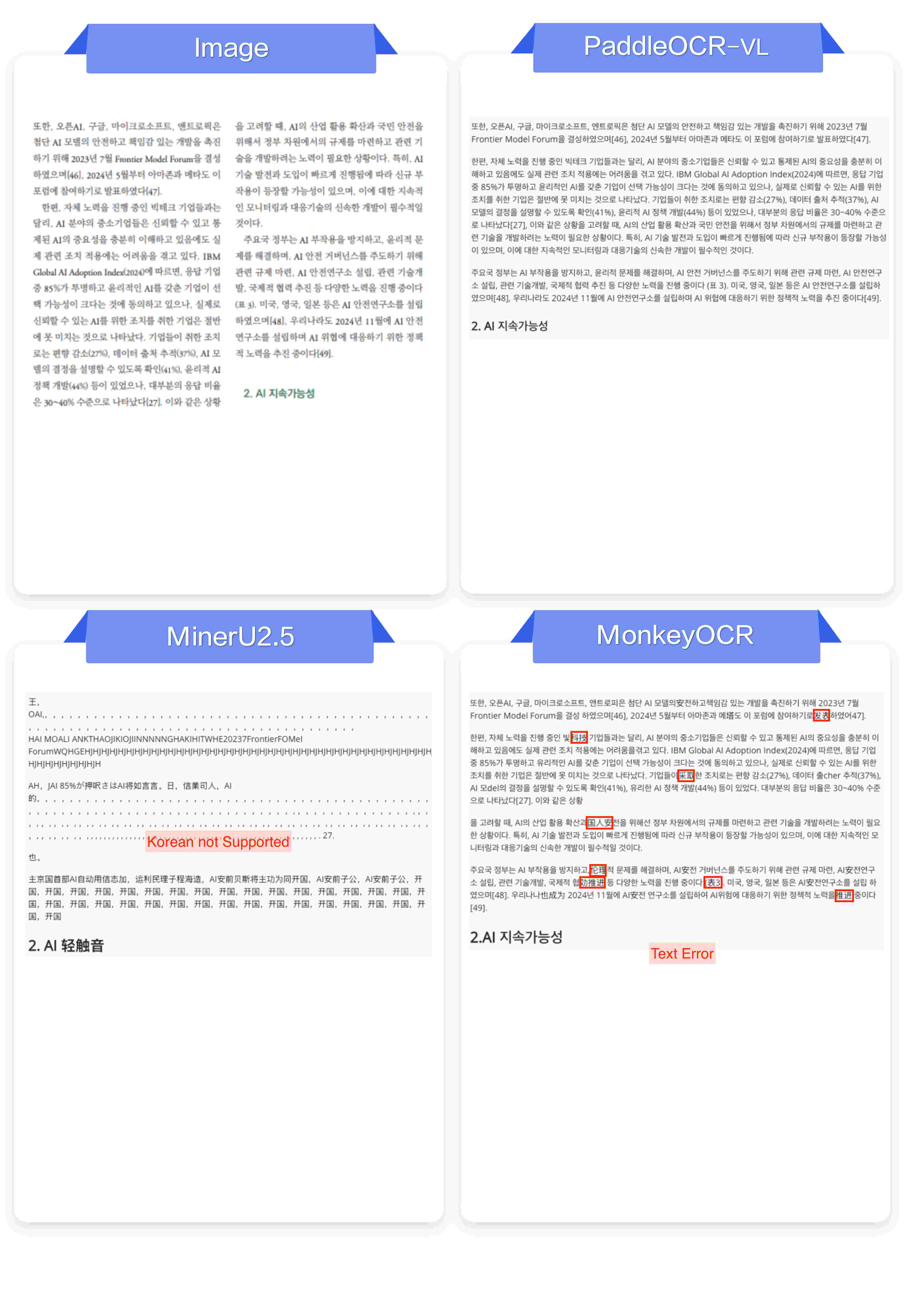} 

\caption{
    \centering
    Compare with others in Multilingual Text Recognition.
}
\label{fig:cmp_text_recognition_multilingual_01}
\end{figure}

\begin{figure}[H]
\centering
\includegraphics[width=0.88\linewidth]{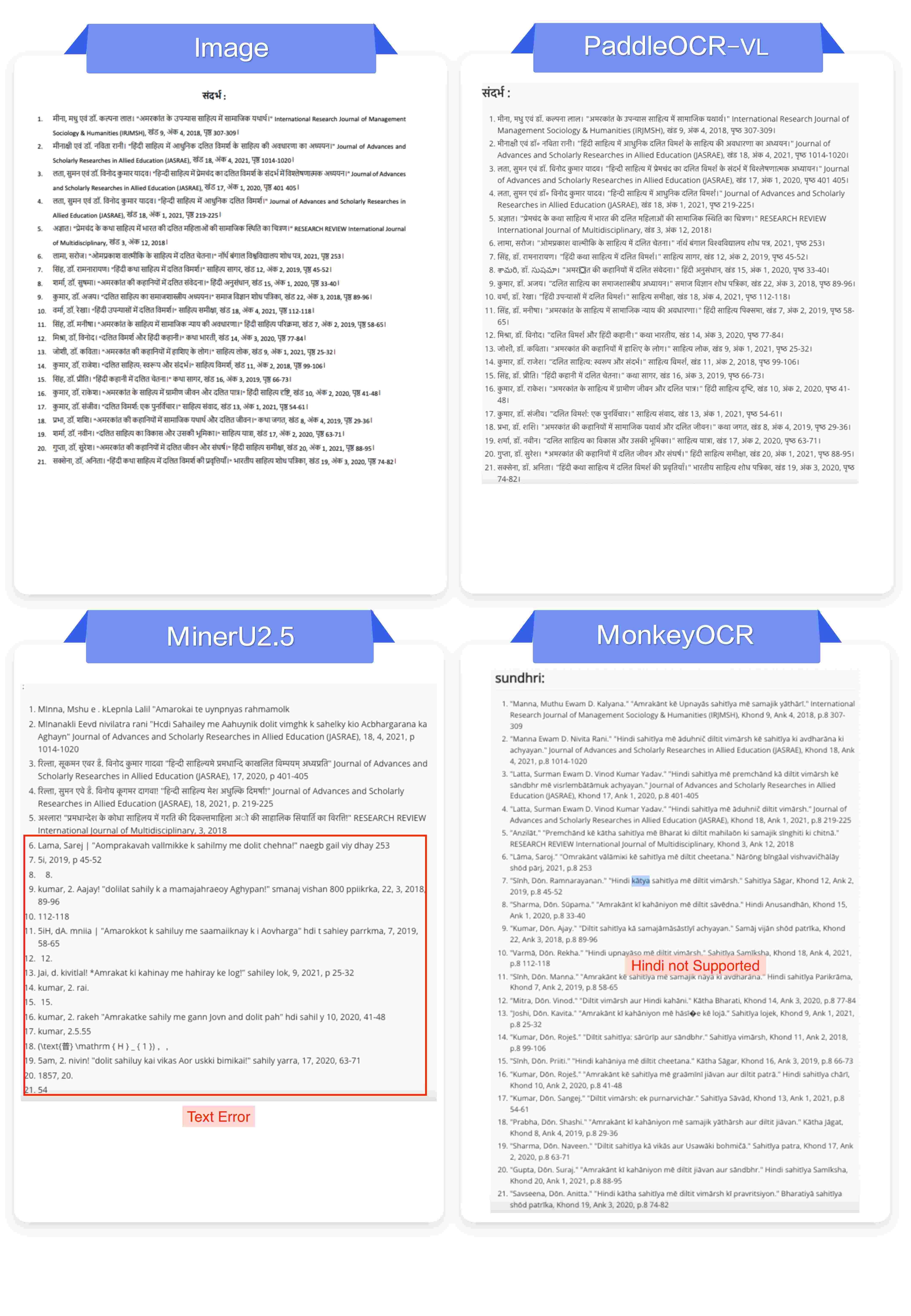} 

\caption{
    \centering
    Compare with others in Multilingual Text Recognition.
}
\label{fig:cmp_text_recognition_multilingual_02}
\end{figure}

\begin{figure}[H]
\centering
\includegraphics[width=0.88\linewidth]{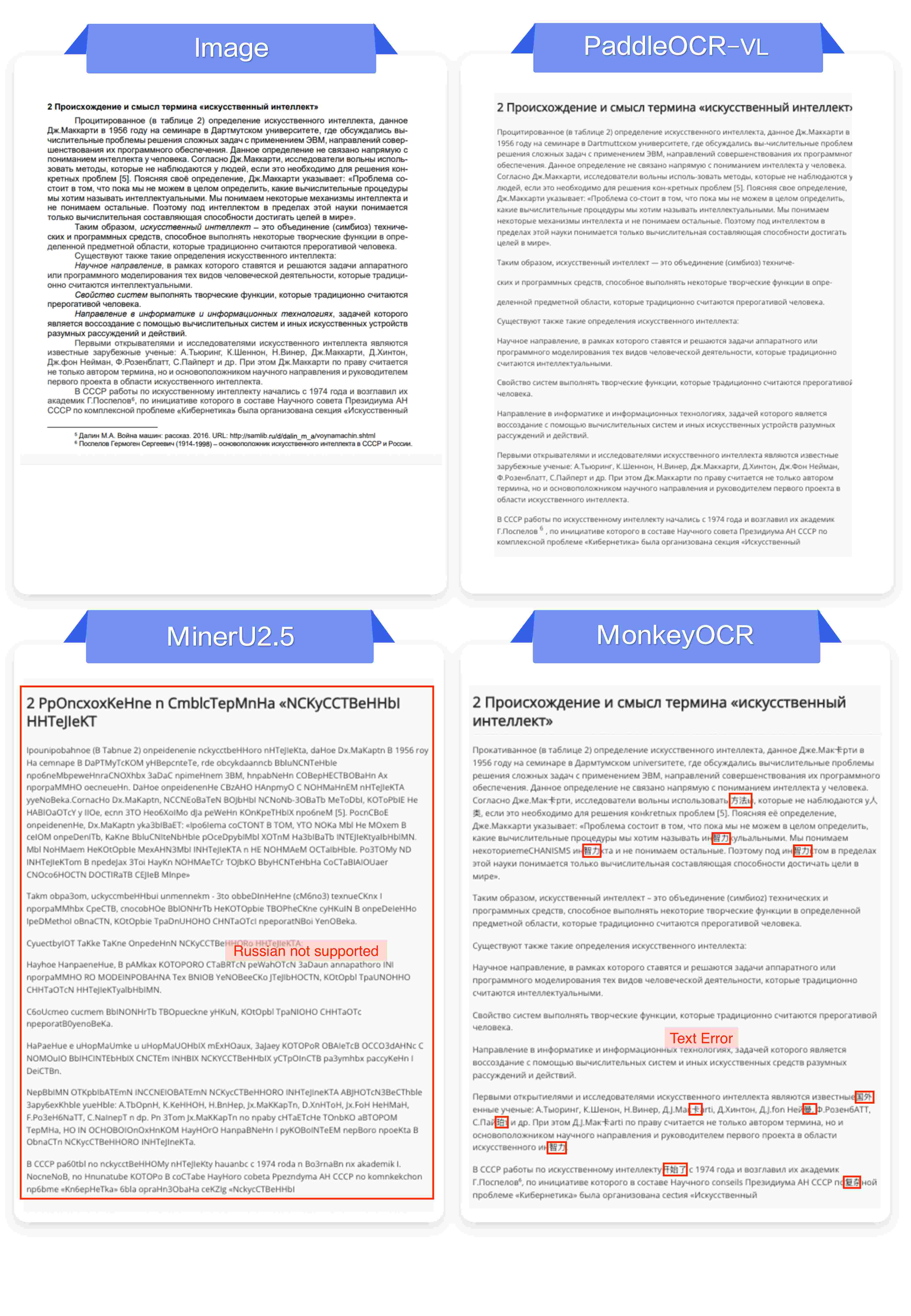} 

\caption{
    \centering
    Compare with others in Multilingual Text Recognition.
}
\label{fig:cmp_text_recognition_multilingual_03}
\end{figure}

\begin{figure}[h]
\centering
\includegraphics[width=0.80\linewidth]{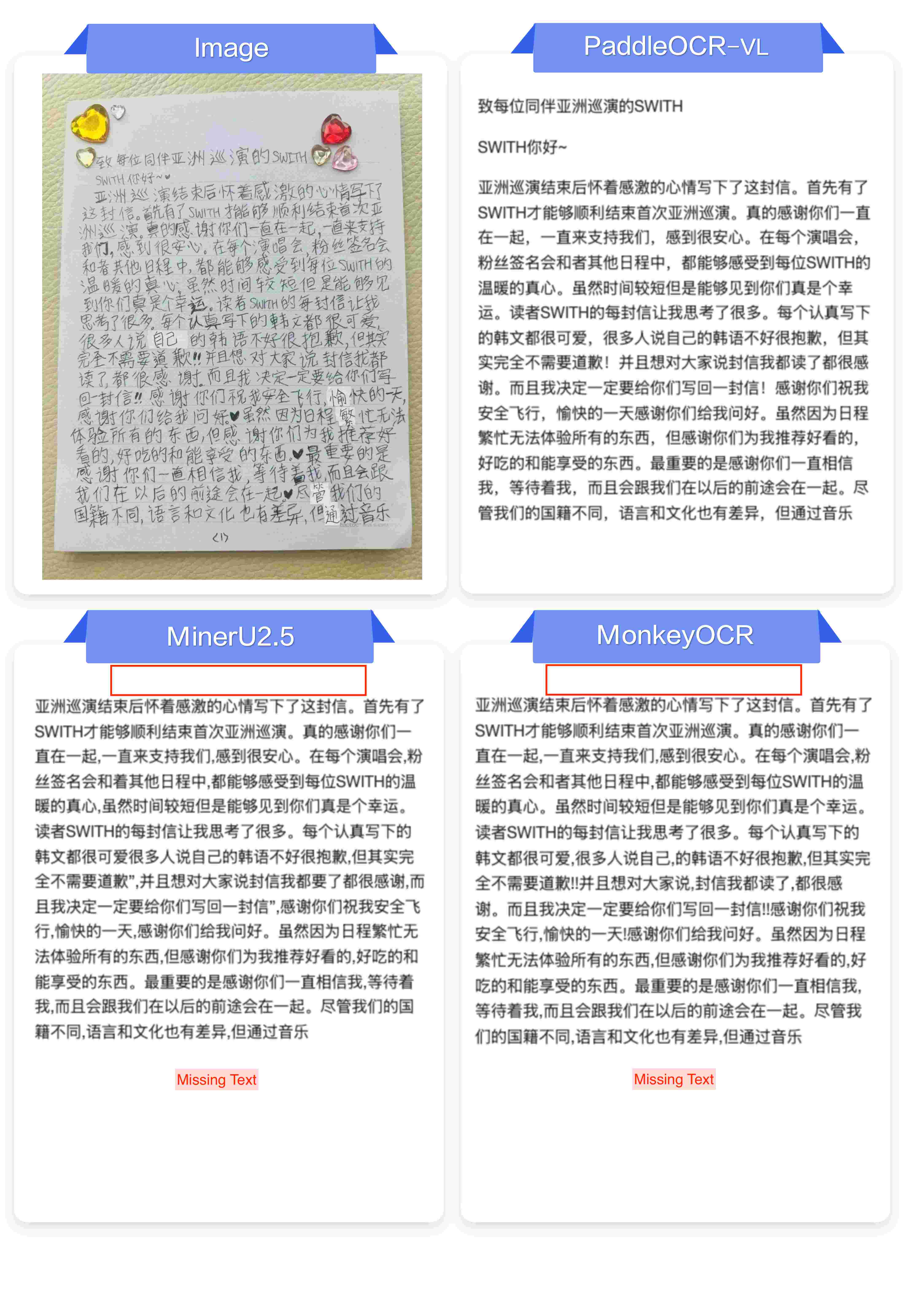} 

\caption{
    \centering
    Compare with others in Handwriting Text Recognition.
}
\label{fig:cmp_text_recognition_handwrite_01}
\end{figure}

\begin{figure}[h]
\centering
\includegraphics[width=0.86\linewidth]{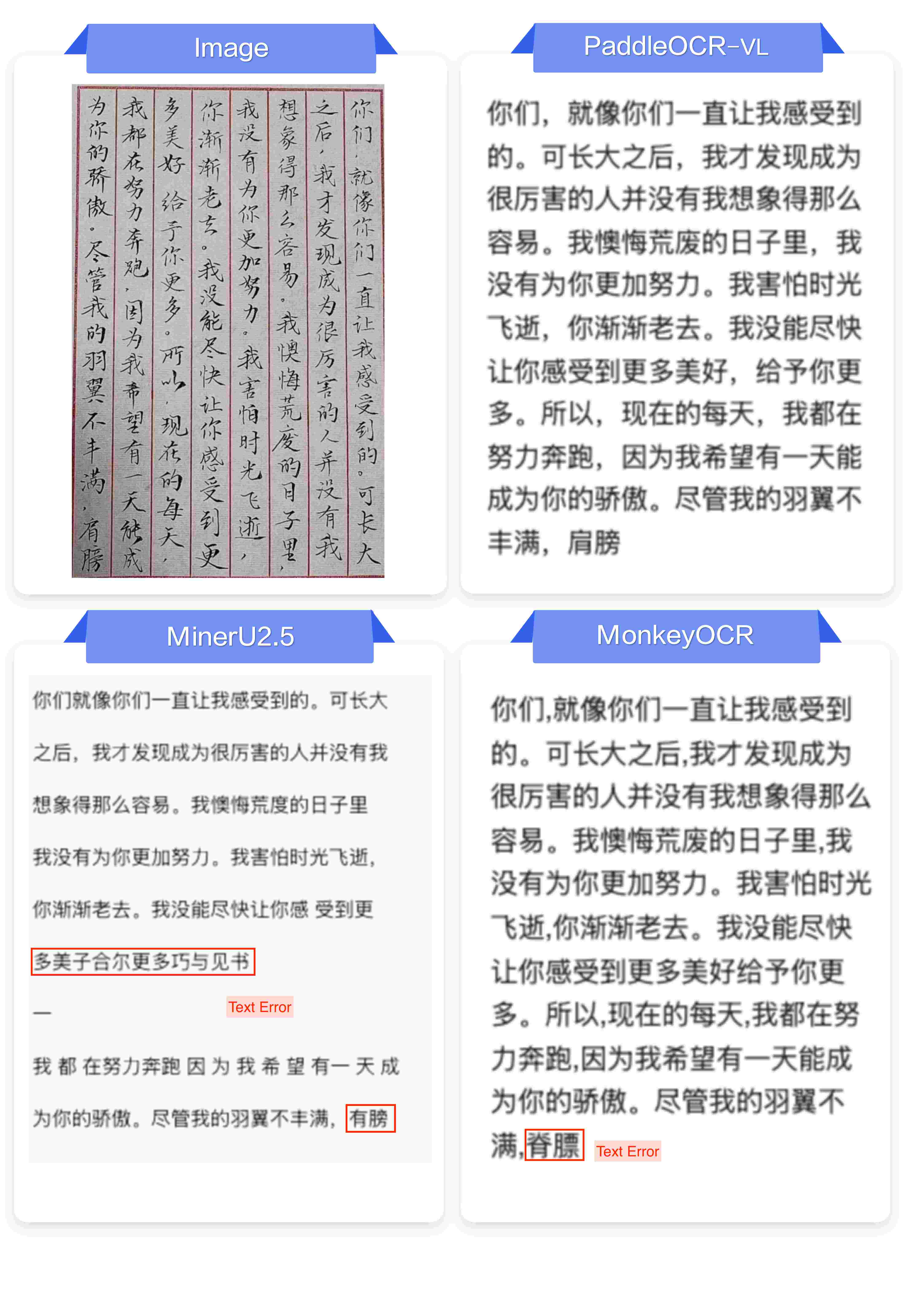} 

\caption{
    \centering
    Compare with others in Handwriting Text Recognition.
}
\label{fig:cmp_text_recognition_handwrite_02}
\end{figure}

\begin{figure}[h]
\centering
\includegraphics[width=0.83\linewidth]{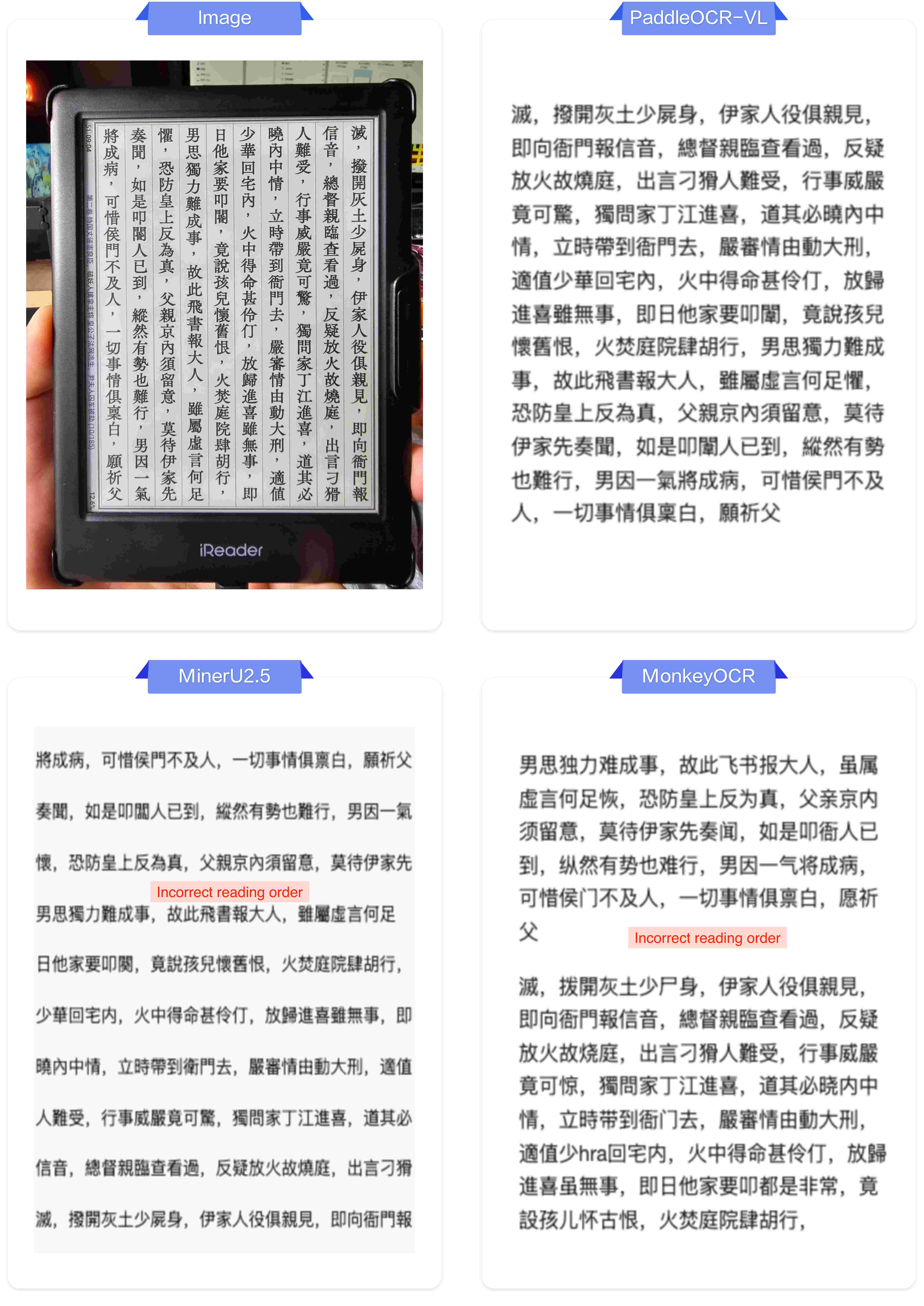} 

\caption{
    \centering
    Compare with others in Vertical Text Recognition.
}
\label{fig:cmp_text_recognition_vertical}
\end{figure}

Moreover, the model demonstrates exceptionally high recognition accuracy in several domains, including Multilingual Text Recognition, Handwriting Text Recognition, and Vertical Text Recognition. Figures \ref{fig:cmp_text_recognition_multilingual_01}- \ref{fig:cmp_text_recognition_vertical} illustrate how PaddleOCR-VL outperforms competitors such as MinerU2.5~\cite{niu2025mineru2} and MonkeyOCR~\cite{li2025monkeyocr}, which tend to misidentify languages like Russian and Hindi as English, overlook some handwritten characters, and struggle with vertical text recognition.

\begin{figure}[h]
\centering
\includegraphics[width=0.80\linewidth]{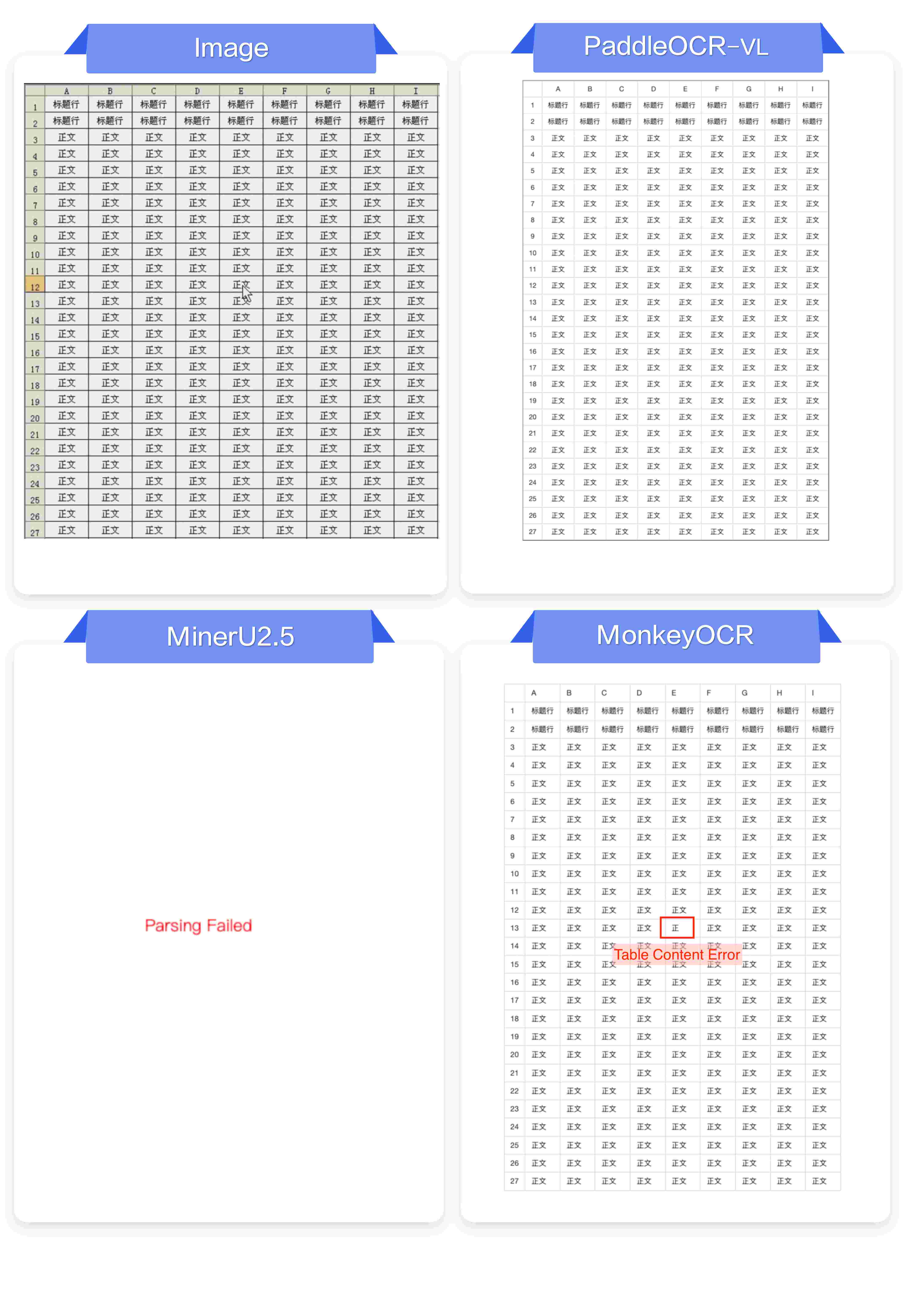} 

\caption{
    \centering
    Compare with others in Table Recognition.
}
\label{fig:cmp_table_01}
\end{figure}

\begin{figure}[h]
\centering
\includegraphics[width=0.85\linewidth]{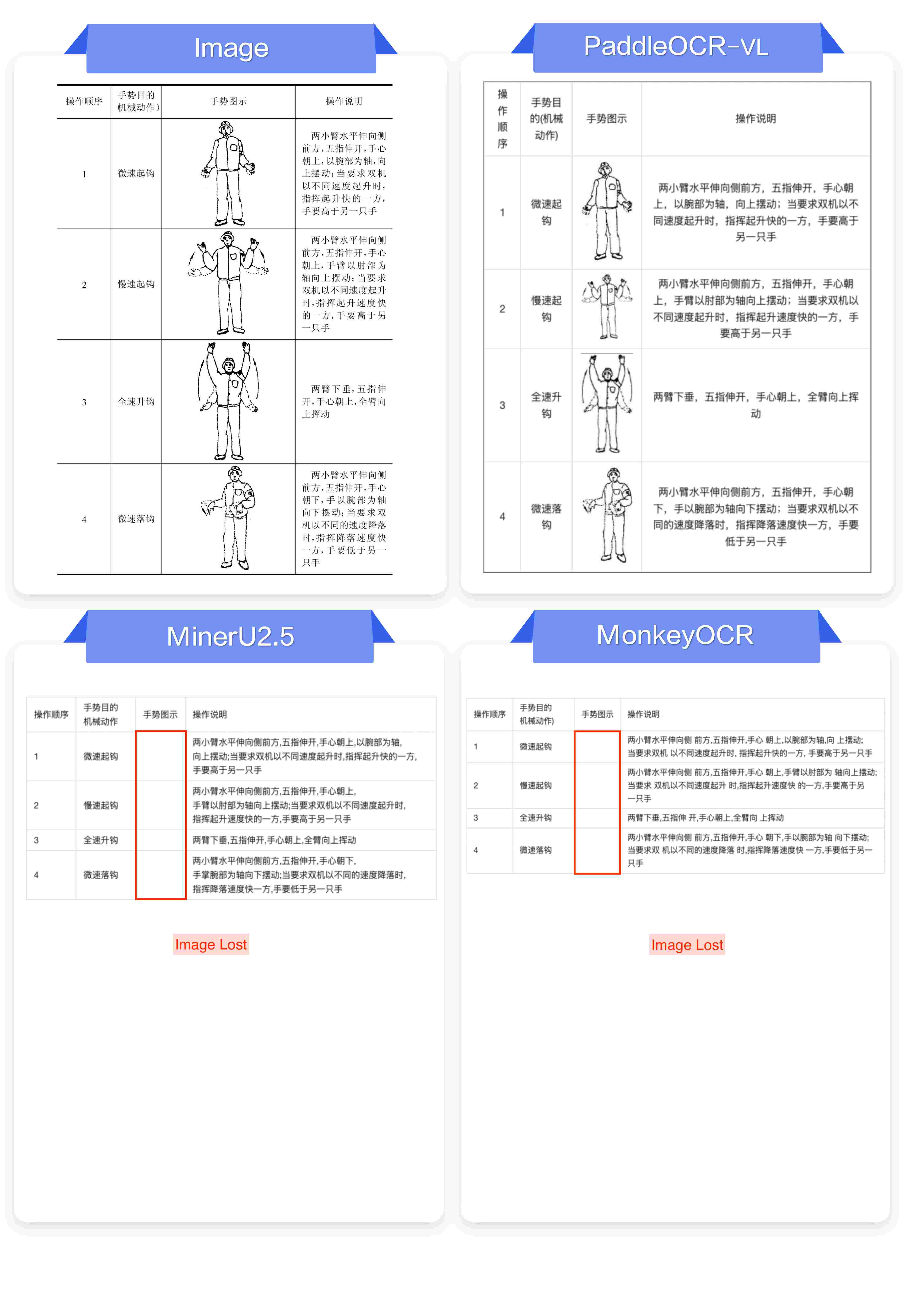} 

\caption{
    \centering
    Compare with others in Table Recognition.
}
\label{fig:cmp_table_02}
\end{figure}

In dealing with complex tables, PaddleOCR-VL's parsing accuracy stands out, as evidenced by Figures \ref{fig:cmp_table_01} and \ref{fig:cmp_table_02}. This is a domain where other models frequently encounter difficulties.

\begin{figure}[h]
\centering
\includegraphics[width=0.83\linewidth]{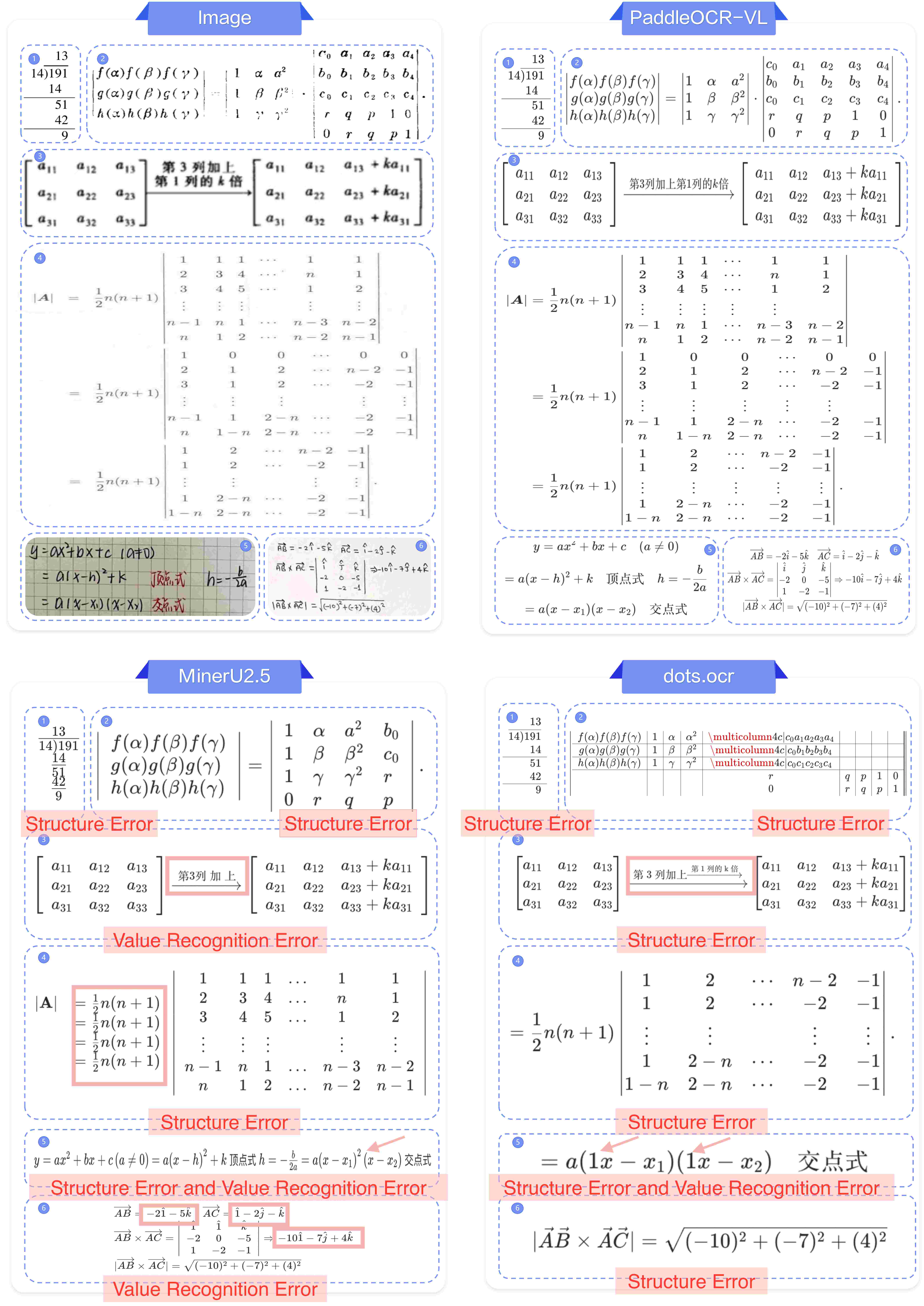} 

\caption{
    \centering
    Compare with others in Formula Recognition.
}
\label{fig:cmp_formula}
\end{figure}

Additionally, Figure \ref{fig:cmp_formula} demonstrates PaddleOCR-VL's proficiency in accurately parsing complex formulas. In contrast, other SOTA models often produce incorrect or flawed outputs when faced with challenging mathematical notations.

\begin{figure}[h]
\centering
\includegraphics[width=0.80\linewidth]{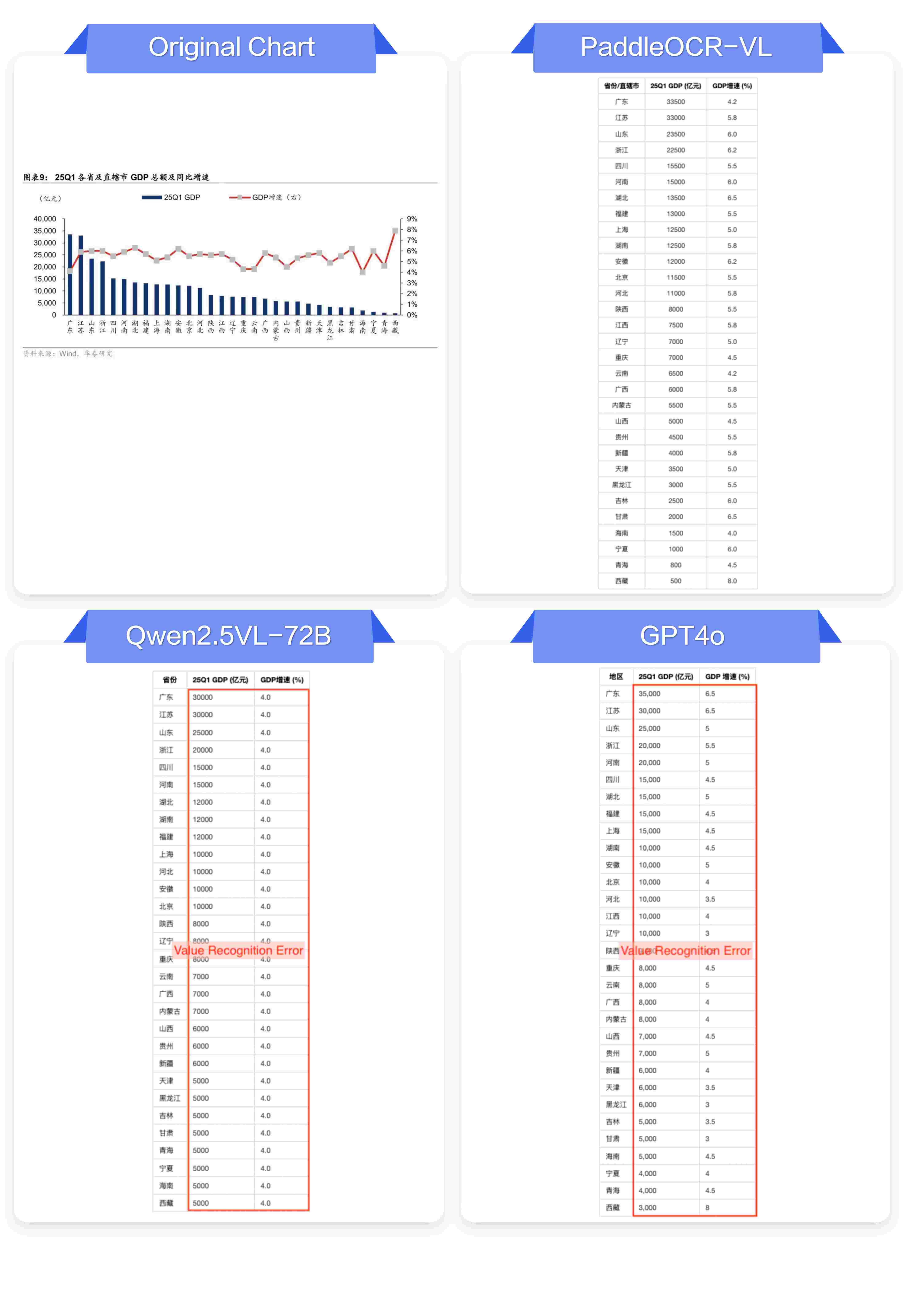} 

\caption{
    \centering
    Compare with others in Chart Recognition.
}
\label{fig:cmp_chart_01}
\end{figure}

\begin{figure}[h]
\centering
\includegraphics[width=0.85\linewidth]{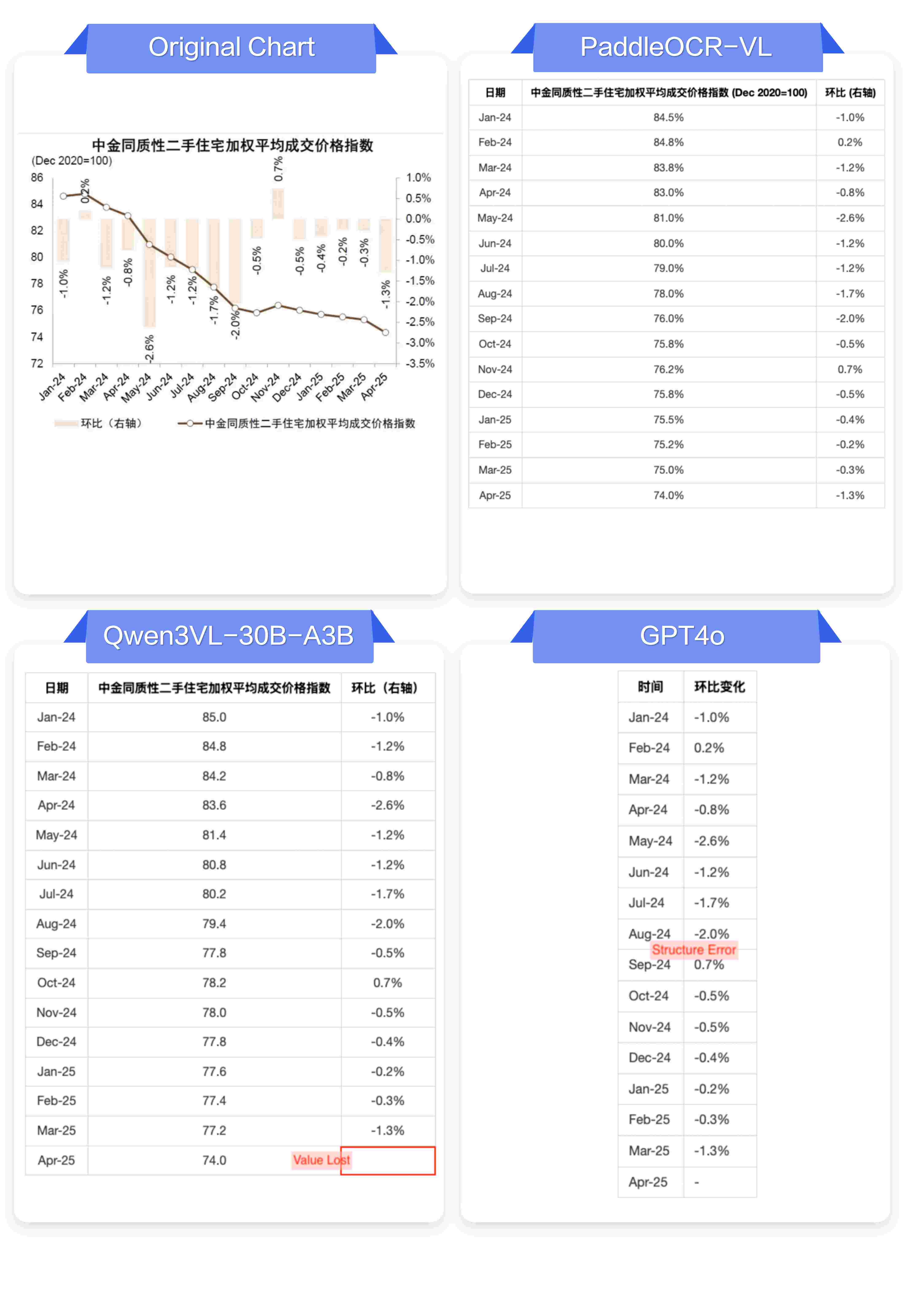} 

\caption{
    \centering
    Compare with others in Chart Recognition.
}
\label{fig:cmp_chart_02}
\end{figure}

Finally, as depicted in Figures \ref{fig:cmp_chart_01} and \ref{fig:cmp_chart_02}, PaddleOCR-VL also excels in Chart Recognition. It outperforms multi-modal large language models like Qwen2.5VL-72B~\cite{bai2025qwen2} and GPT-4o by accurately reconstructing the structure and content of charts.

\clearpage 
\newpage
\onecolumn

\subsection{Comprehensive Document Parsing}
\label{subsec:Comprehensive Document Parsing}

\clearpage 
\newpage

\clearpage 
\newpage

\clearpage 
\newpage

\clearpage 
\newpage
\onecolumn
\subsection{Layout Detection}
\label{subsec:Layout Detection}

\clearpage 
\newpage
\onecolumn
\subsection{Reading Order}
\label{subsec:Reading Order}

\clearpage 
\newpage
\onecolumn
\subsection{Text Recognition }
\label{subsec:Text Recognition}

\subsubsection{Multilingual Text Recognition}

\clearpage 
\newpage
\onecolumn

\subsubsection{Handwriting Text Recognition}

\clearpage 
\newpage
\onecolumn
\subsubsection{Vertical Text Recognition}

\clearpage 
\newpage
\onecolumn

\subsection{Table Recognition}
\label{subsec:Table Recognition}

\begin{figure*}[h]
\centering
\includegraphics[width=0.83\linewidth]{images/appendix/table_01.jpg} 

\caption{
    \centering
    The markdown output for various types of Tables.
}
\label{fig:table_01}
\end{figure*}

\clearpage 
\newpage

\subsection{Formula Recognition}
\label{subsec:Formula Recognition}

\clearpage  \newpage
\subsection{Chart Recognition}
\label{subsec:Chart Recognition}

\clearpage  
\newpage

\subsection{Layout Detection}
\label{subsubsec:layout detection}
\clearpage 
\newpage

\clearpage  \newpage
\subsection{Text Recognition}

\label{subsubsec:text_recognition}
\subsubsection{Multilingual Text Recognition}
\label{subsubsec:text_recognition_multilingual}

\clearpage  
\newpage

\subsubsection{Handwriting Text Recognition}
\label{subsubsec:text_recognition_handwriting}

\clearpage  
\newpage

\subsubsection{Vertical Text Recognition}
\label{subsubsec:text_recognition_vertical}

\clearpage  
\newpage
\subsection{Table Recognition}
\label{subsubsec:table_recognition}

\clearpage  
\newpage
\subsection{Formula Recognition}
\label{subsubsec:formula_recognition}

\clearpage  
\newpage
\subsection{Chart Recognition}
\label{subsubsec:chart_recognition}

\clearpage

\end{document}